\documentclass[12pt]{article}

\usepackage{newtxtext,newtxmath}

\usepackage{graphicx}

\usepackage[letterpaper,margin=1in]{geometry}

\renewenvironment{abstract}
	{\quotation}
	{\endquotation}

\date{}

\makeatletter
\renewcommand{\fnum@figure}{\textbf{Figure \thefigure}}
\renewcommand{\fnum@table}{\textbf{Table \thetable}}
\makeatother

\usepackage{scicite}

\usepackage{xurl}

\usepackage{array}
\usepackage{subfig}  
\usepackage{enumitem} 
\usepackage{tikz}
\usetikzlibrary{trees}
\usepackage{colortbl}
\usepackage{databar}  
\usepackage{tikz}     
\usepackage{pgfplots}
\usepackage{booktabs} 
\usepackage[most]{tcolorbox}
\usepackage{pifont}
\usepackage{pdfpages}
\usepackage{appendix}

\usepackage{xcolor}
\usepackage{multirow}

\newcommand{\barwithsd}[2]{%
    \raisebox{.6\height}{%
    \begin{tikzpicture}[baseline=(current bounding box.center)]
    \fill[teal] (0,0) rectangle (#1*3,0.3);
    \draw[black] (0,0) rectangle (3,0.3);
    \draw[black, very thick] ({#1*3-#2*3},0.15) -- ({#1*3+#2*3},0.15);
    \end{tikzpicture}%
    }
}

\usepackage[breaklinks=true]{hyperref}

\newcommand{\mbd}{\textsc{Media Bias Detector}}
\newcommand\red[1]{{\color{red}#1}}

\def\scititle{
	The Media Bias Detector: A Framework for Annotating and Analyzing the News at Scale
}
\title{\bfseries \boldmath \scititle}

\author{
	Samar Haider$^{1\dagger\ast}$,
	Amir Tohidi$^{1\dagger}$,
        Jenny S. Wang$^{2\dagger}$,
        Timothy D\"orr$^{3}$, \and
        David M. Rothschild$^{4}$,
        Chris Callison-Burch$^{1}$,
        Duncan J. Watts$^{1, 3, 5}$ \and
	\small$^{1}$Department of Computer and Information Science, University of Pennsylvania, Philadelphia, USA. \\
	\small$^{2}$Harvard Business School, Boston, USA. \\
        \small$^{3}$Microsoft Research, New York, USA. \\
        \small$^{4}$Annenberg School for Communication, University of Pennsylvania, Philadelphia, USA. \\
        \small$^{5}$Department of Operations, Information, and Decisions, University of Pennsylvania, Philadelphia, USA. \\
        \small$^\dagger$These authors contributed equally to this work.\\
	\small$^\ast$Corresponding author. E-mail: samarh@seas.upenn.edu
}

\begin{document} 

\maketitle

\begin{abstract} \bfseries \boldmath

Mainstream news organizations shape public perception not only directly through the articles they publish but also through the choices they make about which topics to cover (or ignore) and how to frame the issues they do decide to cover. However, measuring these subtle forms of media bias at scale remains a challenge. Here, we introduce a large, ongoing (from January 1, 2024 to present), near real-time dataset and computational framework developed to enable systematic study of selection and framing bias in news coverage. Our pipeline integrates large language models (LLMs) with scalable, near-real-time news scraping to extract structured annotations---including political lean, tone, topics, article type, and major events---across hundreds of articles per day. We quantify these dimensions of coverage at multiple levels---the sentence level, the article level, and the publisher level---expanding the ways in which researchers can analyze media bias in the modern news landscape. In addition to a curated dataset, we also release an interactive web platform for convenient exploration of these data. Together, these contributions establish a reusable methodology for studying media bias at scale, providing empirical resources for future research. Leveraging the breadth of the corpus over time and across publishers, we also present some examples (focused on the \emph{150,000+ articles} examined in 2024) that illustrate how this novel data set can reveal insightful patterns in news coverage and bias, supporting academic research and real-world efforts to improve media accountability.
\end{abstract}

\section{Introduction}
\label{sec:intro}

The importance of the media in information dissemination and shaping public discourse is undeniable. As the primary vehicle for delivering news and framing societal dialogues, the media influences what people think about and how they think about it \cite{agenda_setting_1972, entman1993framing}. Even in an era where a majority of US adults say they get their news from social media and mobile apps\cite{liedke2023news}, media organizations remain important: news websites or apps are preferred by 23\% of adults, compared to the 18\% who favor social media, and content that is consumed on social media often originates from mainstream media sources \cite{wu2011says}. 
Given its reach and influence, the reliability of the information that is produced by news organizations continues to be an important concern. In contrast with social media, mainstream media organizations typically adhere to professional journalistic standards of fact checking and accuracy, and hence tend not to be the primary sources of what is often referred to as ``misinformation" or ``fake news." Nonetheless, the information they communicate to the public can still be biased in at least two ways that have long been studied by communications and media scholars\cite{stroud2008media, baum2008new, chong2007framing}.

First, ``selection bias" holds that media organizations actively choose which topics, events, issues, and perspectives to cover as well as how prominently or extensively to cover them \cite{bourgeois2018selection}. For example, although Hillary Clinton's use of a private email server while US Secretary of State was arguably a matter of public interest that ought to have been covered by the media in the lead up to the 2016 presidential election, the appearance of ten front page articles in the New York Times in the eight days leading up to the election represented an editorial choice to focus on this one issue over all others~\cite{watts2017don}. 
More broadly, the attention that media organizations pay to topics such as crime or health risks has been shown to be unrelated to, or even inversely related to, the corresponding rates expressed by objective administrative data\cite{sheley1981crime, Rothschild2023,isch2025media}, suggesting again that the degree of coverage of these issues represents a choice by news organizations themselves rather than a reflection of external reality. In both cases, the information contained in the stories that are published might be entirely factually accurate and may even be presented in a neutral, objective manner. Nonetheless, as theories of ``agenda setting'' have long held, selection bias can still distort public opinion by driving attention to some issues at the cost others that might be equally or more important\cite{agenda_setting_1972, mccombs2005agenda, dearing1996agenda}. Moreover, because the counterfactual issues are, by definition, not being paid attention to, the distorting influence of selection bias can be hard to detect.  

Second, ``framing bias" contends that, conditional on having selected a topic to cover, journalists and editors can exploit additional degrees of freedom in how they present the constituent facts and issues, again without necessarily saying anything that would fail a fact check \cite{entman2007framing}. Framing bias can involve choices about the inclusion or omission of specific details, perspectives or narratives, the tone or language used, the context provided, and the inclusion or omission of background information \cite{entman1993framing, gentzkow2006media, Rothschild2023}. Such decisions can  lead a reader to reach a desired conclusion that, while not necessarily false, is only one of potentially many equally true--and potentially qualitatively different---conclusions. For example, news organizations are often accused of covering elections through the frame of sporting contests (aka ``horse race coverage''), which emphasize the popularity of candidates, rather than through the frame of policy differences and other issues that are likely to impact voters post-election \cite{watts2017don, farnsworth2007media}. As with selection bias, framing bias can be subtle and hard to detect from the available data. For example, ``false equivalency" or ``bothsidesism" is a type of framing bias in which manifestly unequal views are presented in a balanced and evenhanded manner \cite{haas2007false, shapiro2016special}. In such cases, the bias can be difficult to identify because it is actively masquerading as neutrality. 

As the media landscape continues to expand and simultaneously fragment, these potential sources of media bias and their attendant consequences for public understanding of controversial issues make the development of scalable tools to measure bias more critical than ever. However, the methods that have traditionally been used to study bias were developed when news cycles were slower and media consumption was more centralized, allowing researchers to manually analyze coverage patterns across a small set of publications. For example, traditional content analysis relies on manual annotation \cite{budak2016fair, lim2020annotating, mitchell2017covering}, which cannot scale to match the volume and speed of contemporary news production. Meanwhile, computational techniques such as keyword frequency analysis and topic modeling\cite{gentzkow2010drives, d2000media, park2009newscube} do scale but struggle to capture the contextual nuances necessary for identifying complex forms of bias \cite{kapoor2024qualitative}. Finally, commercial bias-rating tools such as AllSides, Ad Fontes, and Media Bias/FactCheck measure bias by assigning static, publisher-level ratings along a one-dimensional left-right spectrum \cite{allsides, biasly, mediabiasfactcheck, adfontesmedia}. Although useful, these classifications necessarily overlook within-publisher variation on different topics or events or how these patterns evolve over time, all of which are questions of potential interest to researchers.

Recent advancements in large language models (LLMs) present new opportunities to move beyond these limitations \cite{vaswani2017attention, radford2019language}. Unlike traditional natural language processing methods, LLMs can analyze an article's context to capture nuanced semantic relationships within the text \cite{brown2020language,achiam2023gpt}. By leveraging vast training corpora that encompasses a wide variety of published content, these models can identify subtle variations in tone, emphasis, and narrative structure that shape media bias \cite{pena2023leveraging}. Although the resulting ratings remain imperfect and require careful validation from humans, especially as news cycles shift \cite{feng-etal-2023-pretraining, amirizaniani2024developing}, carefully administered LLMs can now generate human-quality ratings for even very complex and subtle judgments \cite{gilardi2023chatgpt}. Equally important, although the cost of using commercial LLMs still presents some scaling limitations, it is feasible to rate hundreds to thousands of articles per day, allowing for near-real-time, article-level classification across multiple publishers, where we expect costs to continue to fall.   

\begin{figure}[ht!]
    \centering
    \includegraphics[width=0.8\linewidth]{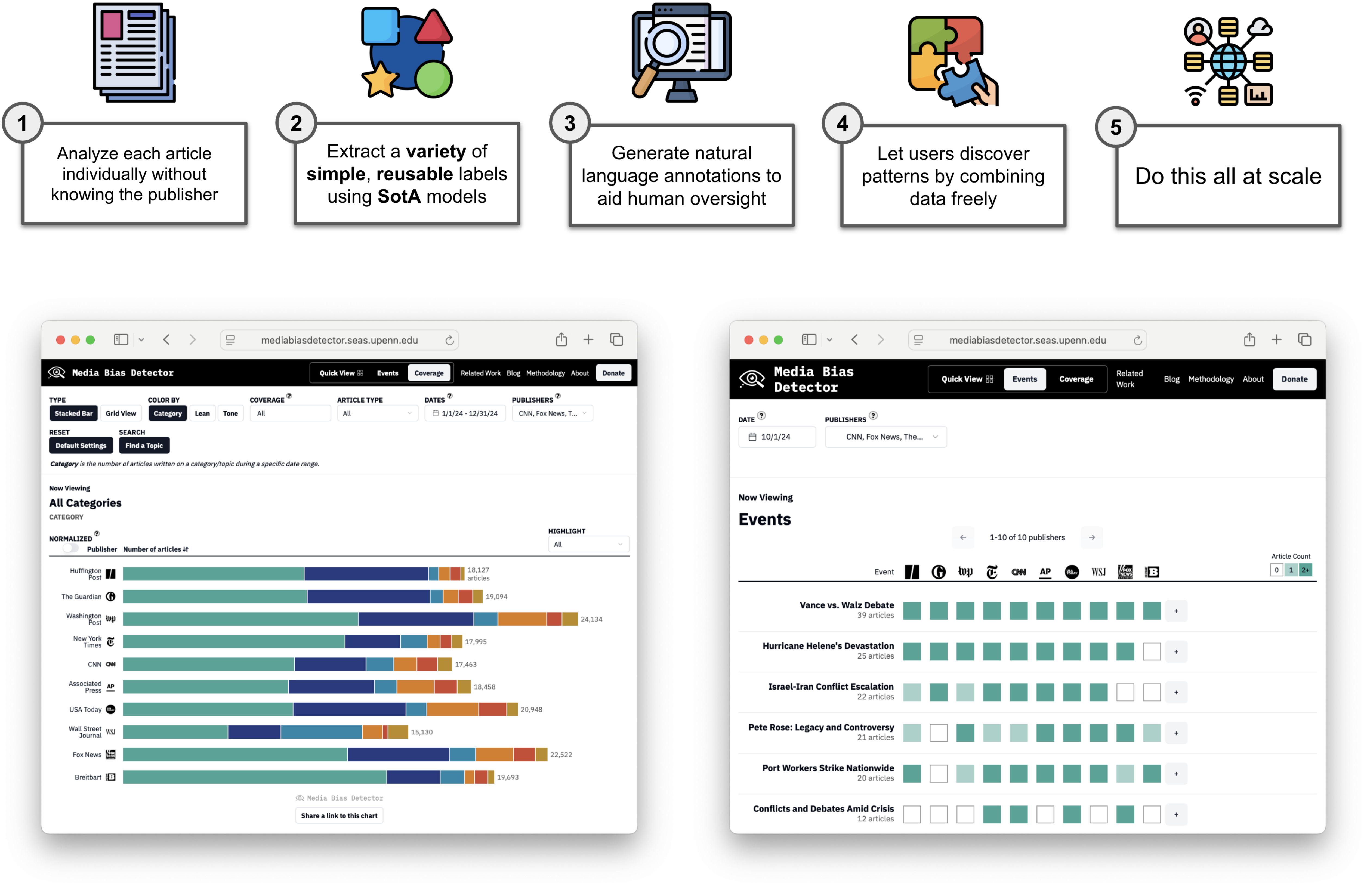}
    \caption{\textbf{Top:} The core guiding principles behind the design of our framework. We analyze each article individually to allow direct, data-driven comparisons between publishers regarding their selection and framing of news topics. By extracting simple labels using state-of-the-art LLMs, we ensure a high level of trust in our dataset. As we have a variety of data points for each article, our framework allows users to freely combine them to gain new insights into the media. \textbf{Bottom:} The two primary views of our dashboard: coverage (left) and events (right). The coverage page gives users a long-term view of what topics the media chooses to cover. Users can set filters to narrow the time frame and click through on each news topic to compare the subtopic-level distribution. They can also choose to color the stacked bars by the tone or political lean reflected in the coverage. The event view, on the other hand, offers a more fast-paced view of the daily news cycle and shows the top news events of the past day along with the amount of attention given to them by each publisher. Users can click on an event to view the top facts within it as well as trace their origin to the original articles themselves.}
    \label{fig:dashboard_screenshots}
\end{figure}

In this paper, we introduce the \mbd, a scalable, near-real-time system that measures elements of framing and selection bias at multiple levels from individual articles to publisher-wide trends. The \mbd\  continuously collects clean and comprehensive data from a collection of major news publishers across the political spectrum. We also describe a unique dataset derived from the \mbd\  and illustrate how this data can be used to analyze media coverage during a critical election year. In this paper we describe data collected during 2024 for the ten publishers with which the \mbd\  initially launched; however, we note that our collection of publishers has grown (currently to 21) as has the time period of our data, where we anticipate both will continue to grow over time. Finally, we make the \mbd\ methodology and data publicly available to support academic research on media bias, and also provide an interactive online tool\footnote{\url{https://mediabiasdetector.seas.upenn.edu}} for users to explore and analyze coverage patterns dynamically \cite{wang2025media}. Figure 
\ref{fig:dashboard_screenshots} shows the core principles that guided the design of our framework and screenshots of our dashboard.

At the core of our methodology is a multi-stage pipeline that systematically tracks and analyzes news articles. While television and cable news also play significant roles in shaping public discourse, we currently focus on text-based digital news where the unit of analysis is each individual news article. We capture homepage snapshots from major publishers at regular intervals, recording article content, as well as metadata on homepage placement and prominence to measure selection bias. Using LLMs, we classify articles by topic, political lean, tone and news type. Additionally, we break analysis down to the sentence level, capturing nuanced aspects of framing bias such as sentence type and tone. We also perform event clustering to group articles together based on their content on a daily basis and track which events received more coverage than others, and by which publishers. 

While existing media bias ratings organizations judge media outlets as a whole, placing them on a discreet left-right spectrum based on the entirety of their reporting as well as any political alignment their pundits or owners may have expressed, our approach evaluates each article independently and without knowledge of the publisher or their history. To analyze articles, we use state-of-the-art LLMs like GPT-4o to extract a large set of simple labels. By focusing on labels that are tractable for current models (such as the topic or tone of a news article) and by validating them with human raters (see below), we ensure a high degree of confidence in the outputs of our framework and the resulting dataset. In addition to extracting numerical labels, we ask the models to generate natural language explanations for the reasoning behind their decisions, which both aids human oversight and also makes our framework more interpretable by helping us understand which aspects of these articles the model focuses on when it makes certain judgments. Finally, we allow users to freely combine our data in various ways so they can use it to answer a variety of research questions about the media. This flexibility is facilitated by the variety of labels we extract for each news article and the number of news articles we have collected in our dataset.

To ensure accuracy and transparency, we incorporate systematic human oversight at multiple stages. Trained annotators regularly validate a random sample of the model outputs every day, correcting biases and ensuring consistency in the LLM's classification.  This human-in-the-loop approach balances scalability with quality control, addressing key limitations of both fully manual and fully automated methods. We also conducted an in-depth data validation exercise to benchmark the effectiveness of LLMs at complex tasks like identifying political lean and tone from news articles. Furthermore, our methodology is transparent and publicly documented, with detailed information about our LLM prompting strategies and validation procedures available to researchers who use our system.

We argue that the \mbd\ enables new avenues of research into media bias that were previously difficult or impossible to study at scale. As we describe in Section 3 of the paper, our data reveal interesting patterns in contemporary news coverage and offer large-scale, quantitative evidence to earlier qualitative claims of bias. Additionally, the scale and granularity of our data enables deeper investigation of these research questions, revealing  nuanced patterns and generating novel insights that extend beyond prior research findings. 
For example, in coverage of political topics we find a systematic tendency for media outlets to focus more on criticizing their ideological opposition than supporting aligned positions, extending previous research suggesting that partisan media effects are primarily driven by negative coverage of opposition candidates \cite{smith2014let}. 
In another example, we find that headlines often present issues in a manner that leans more Republican than the associated article text, complementing recent research showing that biased headlines can significantly impact readers' ability to make inferences and recall facts from news stories \cite{ecker2014effects}. 
In a third example, our data replicates and extends prior claims of negativity bias in the media, which can in turn shape how the public perceives the reality of the world. 

The remainder of the paper is structured as follows: First, we provide a detailed explanation of our system for real-time analysis of media bias, which combines the processing power of LLMs with systematic human oversight. In this section, we also present evidence validating our methodology through extensive testing, demonstrating high accuracy in topic classification, event clustering, and labeling article lean and tone using LLMs. Second, we use our curated dataset to explore a range of questions related to media practices and biases, showcasing the potential use cases of our dataset for gaining deeper insights into media. Finally, we discuss the practical implications of our work and outline how other researchers can build upon our transparent methodology to advance the study of media bias.

\section{Methodology}

\subsection{System Architecture}
\label{sec:system_architecture}

As already noted, our multi-stage data pipeline is built upon large language models, which we utilize for both their capacity to produce deep contextual representations of text documents and their instruction-tuned ability to generate labels without requiring extensive training data. 
Prior work has found that LLMs perform well in generative tasks such as summarization \cite{liu-etal-2024-learning, ravaut-etal-2024-context}, as well as discriminative tasks such as sentiment analysis and topic classification \cite{pena2023leveraging, zhang2023sentiment}, where importantly these tasks can be performed in a zero shot setting\cite{wang_humanllm_2024}. Furthermore, the latest generation of these models exhibit increasingly complex reasoning and problem-solving capabilities \cite{bubeck2023sparks}. These findings suggest that LLMs can be especially useful for the complex task of media bias detection, where understanding context, tone, and subtle language is crucial. 
Critically, LLMs also scale well to large volumes of content that would be unmanageable with manual annotation alone \cite{ziems2024can}. 
As a result, LLMs can competently and efficiently extract detailed information like sentence-level codings and classifications by topic, subtopic, article type, tone, and political lean at the scale of many publishers in close to real time \cite{kiss_mobius}.

\begin{figure}
    \centering
    \includegraphics[width=0.9\linewidth]{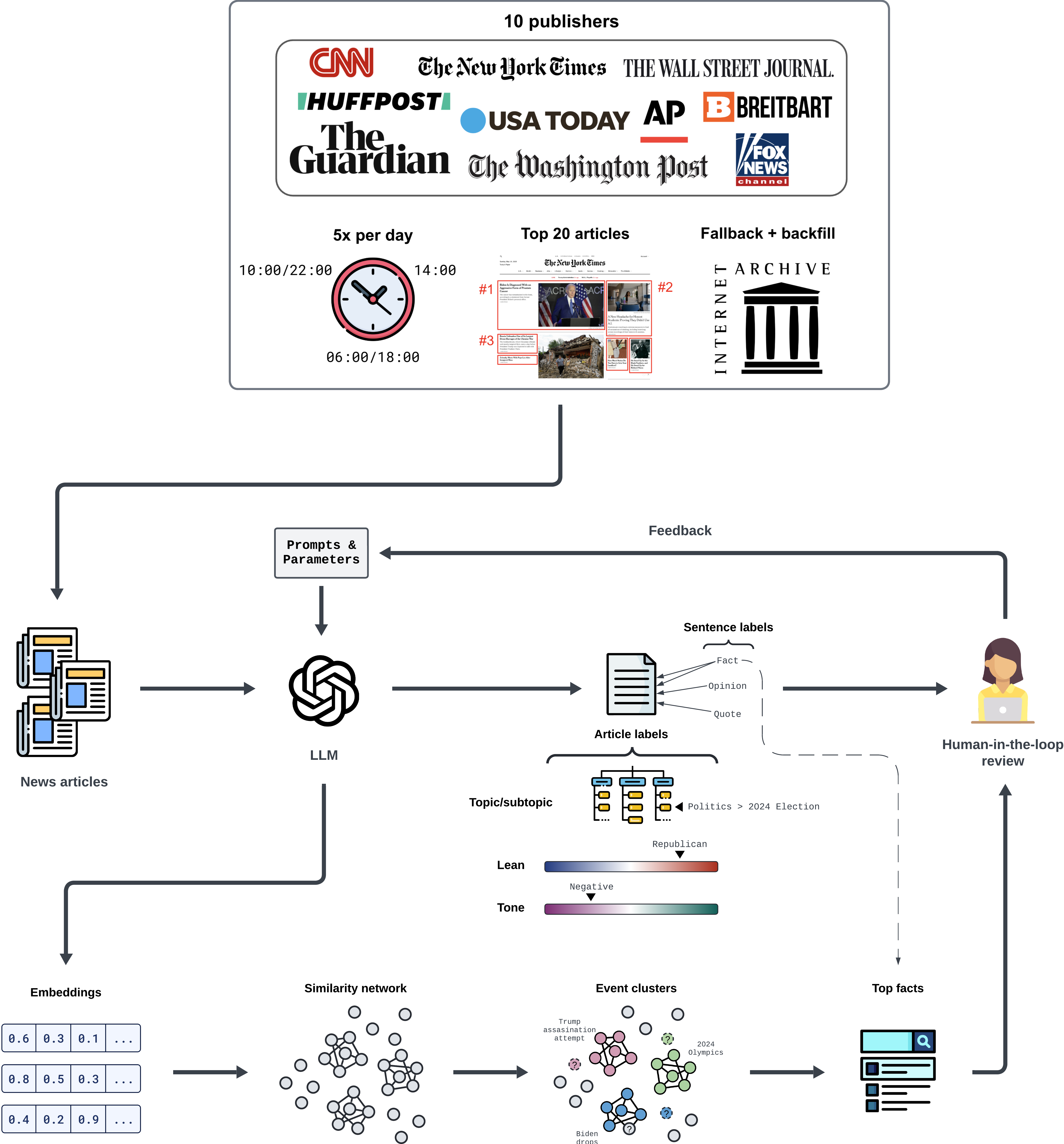}
    \caption{\textbf{The \mbd\ framework.} For each of the ten publishers, we take a snapshot of their homepage five times per day and scrape the top ranking news articles. We then use LLMs to extract multiple levels of structured labels from them at the article-level (topic, subtopic, lean, tone, type) and the sentence-level (type, tone, focus). In addition to this, we use OpenAI's text embedding model to obtain document embeddings for each article, which are used for event clustering. We then generate embeddings for every sentence in a news event and cluster them as well to extract the top facts about it. Our regular human-in-the-loop process adds oversight to this framework and ensures that the generated labels are accurate.}
    \label{fig:pipeline}
\end{figure}

The \mbd\ uses LLMs in two ways: first, for labeling news articles and sentences using zero-shot prompting; and second, for generating contextual embeddings to cluster articles about major events and highlight important facts within them. Figure \ref{fig:pipeline} provides a high level overview of our pipeline, which comprises three primary components: data collection, data labeling, and data validation. In the following sections, we discuss each of these in detail.

\subsection{Data Collection}
\label{sec:data_collection}

Our dashboard currently covers 21 major news publishers chosen to capture a mix of audience size (see Figure \ref{fig:pub_count}), agenda-setting influence, and ideological diversity: Associated Press,  Breitbart News, CNN, Fox News, HuffPost, New York Times, The Guardian, USA Today, Wall Street Journal, the Washington Post, LA Times, Chicago Tribune, Star Tribune, BBC, Financial Times, Reuters, Bloomberg, CNCB, New York Post, Newsmax, and Daily Wire. Of these, the first ten were included in the initial launch in June 2024 and the remaining eleven were added in May 2025. As indicated earlier, we expect the collection of publishers to continue to increase over time, including other publication modes such as television, podcasts, and social media. 

The justification for starting with a relatively small number of prominent publishers and growing the collection over time is twofold. First, our objective of characterizing bias requires us to capture all relevant text while minimizing extraneous content such as ad copy or boilerplate text that could generate spurious results. This goal is complicated by the highly variable formatting of news websites, in particular dynamic mixed-mode features such as ``live" articles that increasingly serve as top stories on several publications. Focusing on just a few publishers allows us to customize our approach to individual websites and hence obtain a level of precision that would be infeasible at the scale of hundreds or thousands of publishers. Second, media consumption, especially by elites, is highly skewed such that the handful of prominent actors that we do cover arguably plays a larger agenda setting role than even thousands of publishers with tiny audiences. 

Every day, we take a snapshot of each publisher's homepage five times, at 6 AM, 10 AM, 2 PM, 6 PM, and 10 PM Eastern Standard Time (EST). Then, we use headline placement on these homepage snapshots to identify the top 30 most prominent articles displayed to readers. We assign this position on the page as a combination of distance from the top, font size of the title text, and the presence and size of images accompanying the article. Currently, we disregard content that is not a news article, such as videos, podcasts, and photo galleries.
We then scrape the text, title, and date of publication for every article in these top 30 (while we collect the data for the top 30 articles, for cost purposes the labeled data that we release on the dashboard and review in this paper is currently restricted to the top 20 articles). We then pre-process the text to remove superfluous and boilerplate content and remove advertisements, direct mentions of the publisher outside the context of the story, and repetitive phrases that are irrelevant to the article (e.g. ``Listen 5 minutes", ``Click Here for More Information", ``Enter your email address"). We continue to grow this dictionary of repetitive phrases as we collect more data.

Our data collection started on Jan 1, 2024 for an initial ten publishers (Associated Press, Breitbart News, CNN, Fox News, Huffington Post, New York Times, The Guardian, USA Today, Wall Street Journal, the Washington Post) and an additional eleven publishers (LA Times, Chicago Tribune, Star Tribune, BBC, Financial Times, Reuters, Bloomberg, CNCB, New York Post, Newsmax, and Daily Wire) were added in May 2025. In planned future work, we will backdate our coverage of the later additions to Jan 2024 by using screenshots drawn from the Internet Archive \footnote{\url{https://archive.org}} to replace the ``live'' screenshots in the first step of the data acquisition process described above (we will also use this procedure to backdate all 21 publishers prior to Jan 2024). 
We note, however, that the illustrative results presented in this paper are based only on the initial ten publishers for the period January 1 to December 31, 2024, during which time we collected over 150,000 unique news articles. Figure \ref{fig:pub_count} shows the total number of articles per publisher, along with their most recent monthly traffic statistics \cite{pressgazette_topnewswebsites}. As all publications are being scraped at the same times, for the same number of articles, the differences in the number of unique articles reflects the speed in which they rotate their content.

\subsection{Data Labeling}
\label{sec:data_labeling}

\begin{figure}[ht]
    \centering
    \includegraphics[width=0.8\linewidth]{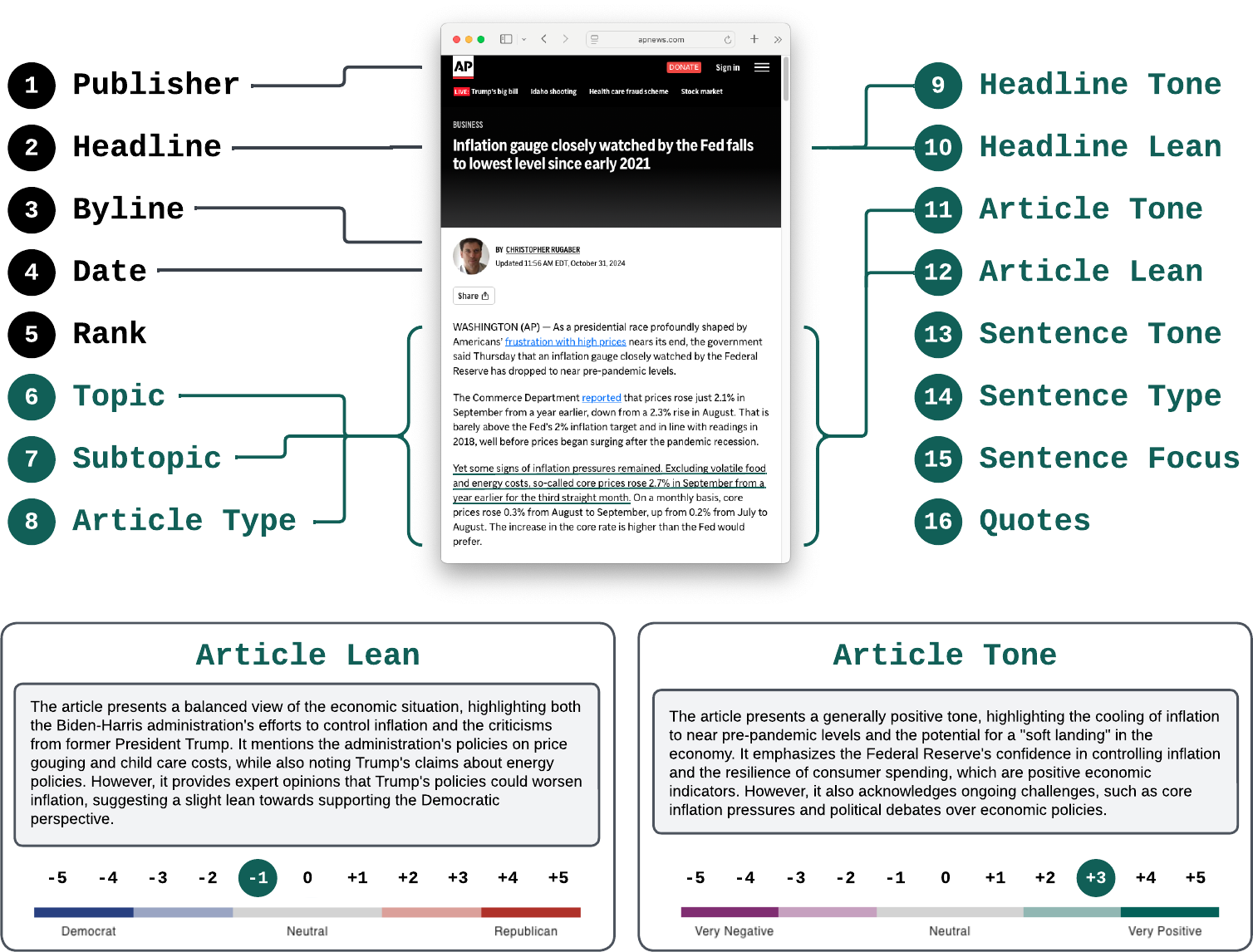}
    \caption{\textbf{Top:} A list of the data points we extract from each article. The data in black (1-5) are obtained automatically during the scraping process, and the ones in color (6-16) are extracted using LLMs. All of these data points can be combined together in various ways to answer a multitude of research questions about the media. \textbf{Bottom: } An example of GPT-4o's article lean and  tone analysis and labels for the news article. The analyses generated by GPT-4o showcases its deep understanding of the subject matter of the article, how it connects to the political landscape, and how it's framing can be read as support for one party or the other.}
    \label{fig:data_points}
\end{figure}

Measuring media bias is an inherently complex task even for humans, in part due to the lack of a universally accepted set of standard metrics \cite{spinde2023media}. Unlike outright falsehoods, which can be fact-checked against verifiable sources\cite{lazar_2018, soroush_2018, pennycook2021psychology}, bias is inherently subjective and difficult to quantify \cite{park2021presence}. Readers, that is, often perceive articles as ``biased" on the grounds that they disagree with them, but in so doing they are (usually implicitly) treating their own subjective opinions as the ``truth" \cite{pronin2004objectivity}. In many cases of interest, such as contentious social and political issues for which subjective views vary widely, there is no \emph{objective} or even \emph{inter-subjective} ground truth for measuring bias in the news. As noted earlier, for example, even ostensibly neutral articles can be perceived as biased on the grounds that they ought not to have been neutral. Unfortunately, however, in such cases the appropriate lack of neutrality is highly subjective. A second challenge is that the distinction between selection and framing bias is often blurry in practice. Consider, for example, a journalist who--intentionally or unintentionally--writes a negative story about the state of the economy by emphasizing facts about high inflation while omitting facts about high employment and wage increases. From one perspective, this example looks like framing bias: having selected the topic of the economy, the journalist is framing the story around inflation rather than employment. From another perspective, however, it resembles selection bias, just operating at the level of facts rather than topics.

In response to these challenges, our approach is guided by two general principles. First, we avoid defining bias in terms of a deviation from an objective truth or reality, instead, measuring more easily quantifiable properties such as tone and lean, and looking for patterns in these properties across different publishers and over time. By observing differences between publishers at one period in time, or by tracking shifts within a single publisher over time, we can detect and surface the signatures of potential bias, but we refrain from assigning explicit ``bias scores" to individual stories or publishers. Second, we also avoid making a categorical distinction between selection and framing bias, instead labeling our data at both the sentence and article level thereby building in the flexibility to aggregate at either of these levels as well as higher levels such as events, topics, publishers, and even groups of publishers. Although our approach is in some respects less satisfying that definitive seeming bias scores and cannot detect certain instances of bias (e.g. false equivalency at the article level) we believe that it is more robust and scalable than methods that assume a notion of objective truth. Moreover, in instances where objective data is available (e.g. published economic indicators), our data can potentially combined with such data to yield more definitive claims of bias for those specific cases. Figure \ref{fig:data_points} shows the various data points we extract from each news article, a sample of which can be found in Text S1, with more available for viewing online\footnote{\url{https://tinyurl.com/media-bias-detector}}.

\subsubsection{Topic labeling}
\label{sec:topic_labeling}

\begin{figure}[ht]
    \centering
    \subfloat{
        \includegraphics[width=0.5\linewidth]{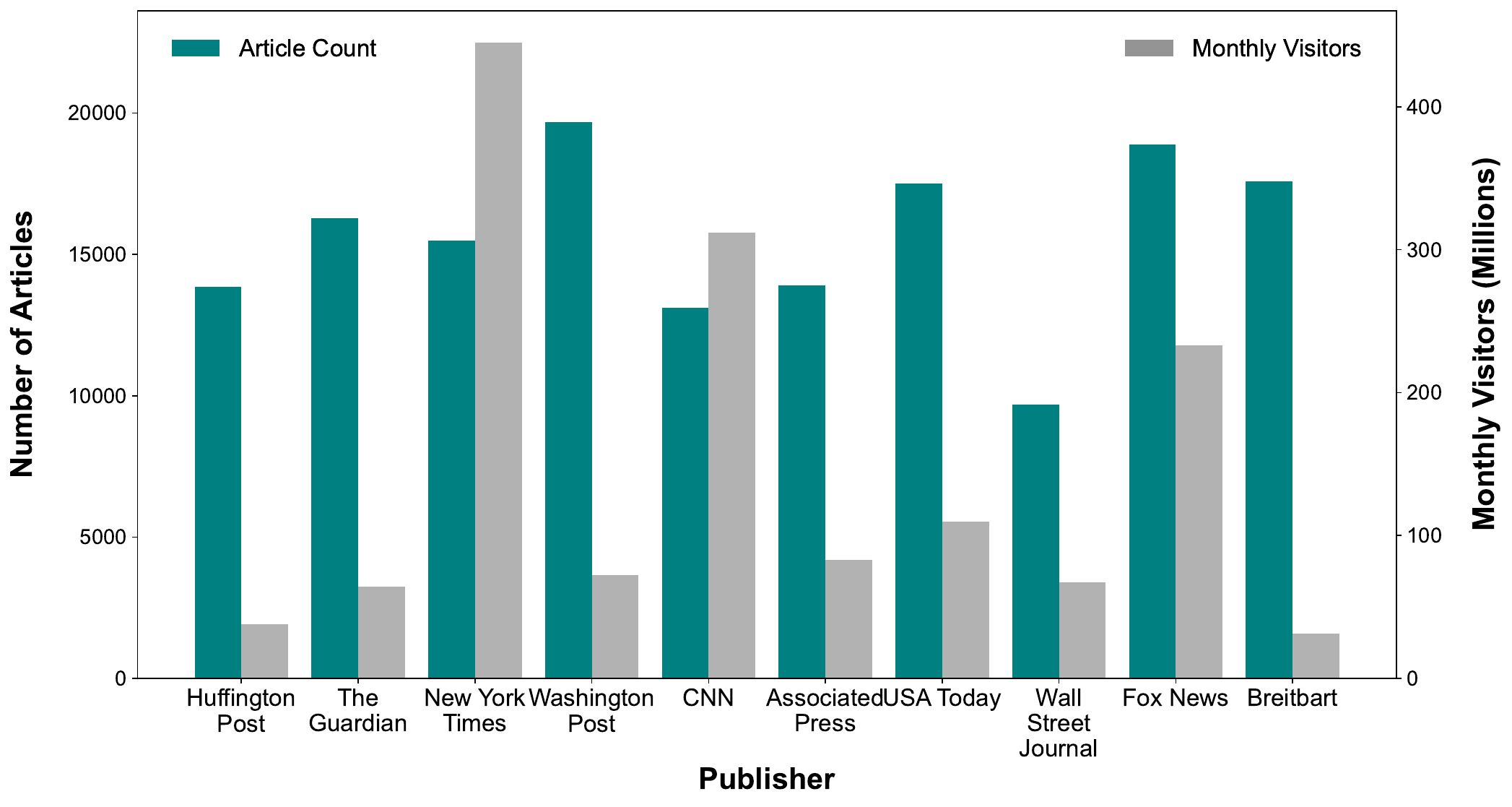}}
    \hfill
    \subfloat{
        \includegraphics[width=0.47\linewidth]{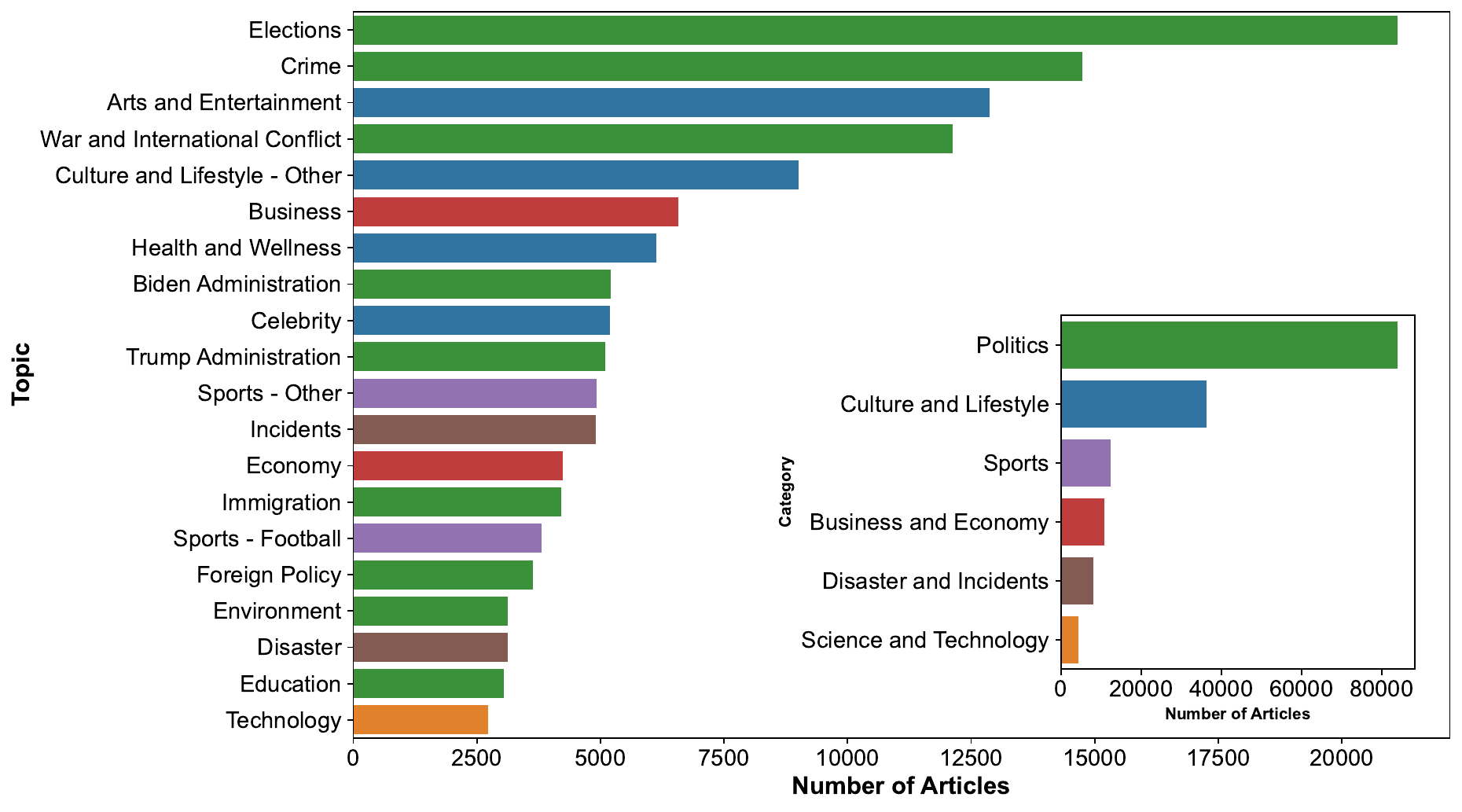}}
    \caption{\textbf{Left:} Total number of unique articles in our dataset from each publisher in 2024, compared to their current monthly traffic for May 2025 \protect\cite{pressgazette_topnewswebsites}. \textbf{Right:} The most popular news topics across this dataset, colored by category (shown in the inset). As expected in an election year, most of the news was dominated by politics, which comprises more than half of all articles in our dataset.}    
    \label{fig:pub_count}
\end{figure}

Each day we used the LLM to classify the content of every news story at three levels: Category (Politics, Culture and Lifestyle, Business, Health, Disaster, Economy, Sports, and Science and Technology); Topic (e.g. within politics: crime, elections, etc.); and subtopic (within crime: violent crime, political corruption, etc.). Given the large number of articles per day and the many possible ways to classify their content, we created a top-down category-topic-subtopic hierarchy via an iterative human-in-the-loop process. First, a selection of co-authors hand-coded the topic and subtopic of all of the front-page articles in the New York Times and Washington Post from September 1 to the date of the midterm elections on November 8, 2022 \cite{Rothschild2023}. Second, we asked the LLM to freely generate these labels for one month of news. Then, with a mix of humans and LLMs, checked to see which topics and subtopics clearly fell under our initial set from the 2022 work, and what new topics and subtopics needed to be generated. Third, we followed an iterative process of asking the LLM to pick its labels from this list instead or suggest others if the list did not work. After several iterations we landed on an initial list of topics and subtopics. Currently, the list contains 36 topics and 194 subtopics across the eight categories, where we add new topics and subtopics when human editors see evidence of articles not cleanly falling within our current list. 
Each article is assigned to exactly one topic and subtopic, and each predefined topic is mapped to one of these eight categories, defaulting to Politics when an article may fall in several categories. 
Clearly many articles could plausibly be classified into more than one topic, and many other topics and subtopics could have been included beyond the ones we chose. However, our preference for single assignment to a relatively short list of topics has the dual benefit of improving classification accuracy and simplifying aggregation. 

Table \ref{fig:topic_hierarchy} shows the full list of topics and subtopics developed in this process, and Figures \ref{fig:topic_prompt} and \ref{fig:subtopic_prompt} show the topic and subtopic labeling prompts that ask the LLM to choose from within this hierarchy, respectively. The most popular news topics in 2024 by number of articles can be seen in Figure \ref{fig:pub_count}. Most of the coverage was dominated by politics, in particular the U.S. Presidential Election.

\subsubsection{Article Lean, Tone and Type}
\label{sec:article_labeling}
 
Aside from topic, the two article-level properties that we measure are political lean and tone. Political lean, to which we also refer as lean, is defined as ``the extent to which the article either explicitly or implicitly aligns with the viewpoints, policies, or concerns of Republicans vs. Democrats." Lean can manifest itself as overt partisan bias, such as when an article directly criticizes a political party or its representatives, but can also be more subtle, such as when certain topics or perspectives are inherently more aligned with one party's preferences or policies. For example, GPT-4o tends to code articles supporting social justice movements as Democrat-leaning, even when the article neither endorses nor critiques either party, on the grounds that such issues are more explicitly associated with Democrats than Republicans. 

Lean, in other words, is itself a complex and multidimensional concept that is also somewhat subjective. To improve consistency and transparency, we ask the LLM for a numerical score on an 11-point Likert scale in the range of [-5, +5] and also ask it to generate a natural language analysis of why it thinks this score is appropriate (see Figure \ref{fig:article_lean_prompt} for the full prompt). By leveraging what is known as ``chain-of-thought'' prompting \cite{wei2022chain} this approach improves the model's reasoning (because it pays attention to its own analysis before assigning a score) and provides a human-readable explanation for its decision which makes its output more interpretable. As we will describe in Section \ref{sec:lean_tone_validation}, these explanations generally make sense to human raters who also overwhelmingly agree with GPT-4o's lean ratings. We caution that we cannot rule out that both GPT-4o and human ratings are themselves biased in the sense that a right-leaning rater might code lean systematically differently from a left-leaning rater. Nonetheless, we believe that lean ratings are more reliable than asking directly about bias in the sense that it is easier (both for LLMs and humans) to determine that a particular article is pro-Republican or pro-Democrat than what its ``true'' orientation ought to be, and hence how it differs from the truth. 

In addition to lean, we also measure tone, defined as ``the extent to which the article overall has a positive vs. negative tone. As with lean, the tone of an article--or of a body of news coverage---also represents an editorial choice that can shape consumers' opinions. For example, talking about the economy in a pessimistic tone can be interpreted as a publisher taking a negative stance with respect to the incumbent government, and by extension, the ruling political party. Although tone tends to be simpler and more easily interpretable than lean, it too is open to a certain amount of subjectivity. As with tone, therefore, we also exploit chain-of-thought reasoning, asking the LLM both for a numerical rating in the range [-5, +5] and a human-interpretable explanation for the rating (see Figure \ref{fig:article_tone_prompt} for the full prompt). Figure \ref{fig:data_points} shows an example of the output from both prompts illustrating the surprising sophistication and depth of the explanations. In Section \ref{sec:lean_tone_validation}, we provide results on the validation of these labels.

Finally, we repeat the above labeling process separately for the articles' headlines in order to study how they may deviate from the actual article text in terms of lean and tone, which is itself a form of media bias. We use the same wording for both prompts with the addition of the article's topic to give the model more context to interpret the headline. The prompts for headline lean and tone labeling can be found in Figures \ref{fig:headline_lean_prompt} and \ref{fig:headline_tone_prompt}, respectively.

\subsubsection{Sentence Type, Tone, and Focus}
\label{sec:sentence_labeling}

In addition to article-level labels, we also label every sentence in an article with features: type, tone, and focus. The type of a sentence can be one of ``fact", ``opinion", ``borderline", ``quote", or ``other," where we note that we do not ask the LLM to evaluate the truth or falsity of any claims, and hence by ``fact'' we mean only that it has the form of a factual claim or statement, not that it is a true fact \cite{mitchell2018distinguishing}. The tone of a sentence is defined in a similar way to tone at the article level but is assigned only one of three values: ``positive", ``negative", or ``neutral," where the coarser rating reflects the more limited data available at the sentence level. Finally, focus asks if the sentence is referring to one political party or another, offering the options: ``Democrat", ``Republican," ``neither," or ``both." We note that focus differs from lean in that it only asks what is being referred to, not whether the reference is ``pro" or ``con," and hence is simpler to answer reliably (we do not code individual sentences for lean, as these ratings do not appear reliable). Although these labels are extracted at the sentence-level, we still require the context of the article to make these classifications since a sentence by itself may not have enough content to accurately understand its tone and who it is referring to. Therefore, we pass the entire article to the model in the form of an enumerated list of sentences, which we obtain by using the Stanza library \cite{qi-etal-2020-stanza}. The full sentence labeling prompt can be found in Figure \ref{fig:sentence_prompt}. 

\subsubsection{Quote Extraction}
\label{sec:quote_extraction}

Quotations serve as critical elements of news articles, providing first-hand perspectives to readers while simultaneously enabling journalists to frame their stories by choosing whom to quote and how. This allocation of voice---determining whose perspectives are amplified through direct quotation---represents a critical yet understudied dimension of media influence. To address this problem, we extract structured information for all quotes in a news article using an LLM. We perform this task independently because, unlike fact or opinion sentences, quotes are accompanied by important contextual information that may be found in a separate sentence, often in a different part of the article altogether. This metadata, such as the name, occupation, and affiliation of the person being quoted, can prove valuable for studying who gets quoted in the news. For each quote in the article, we extract the name of the person being quoted as well as their occupation and affiliation if that information is available anywhere in the article. We also ask the LLM to classify the person's domain (e.g. Politics, Sports, etc.) and the capacity in which they are being quoted (e.g. an observer, an expert, etc.) based on the context of the article. The full quote extraction prompt can be found in Figure \ref{fig:quote_prompt}.

\subsubsection{Event Clustering}
\label{sec:event_clustering}

To identify articles that are about the same event, we develop a novel clustering framework that yields very precise news clusters and generates easily interpretable labels for them. The clustering pipeline works at the day-level and is displayed schematically at the bottom of Figure \ref{fig:pipeline}. We  use OpenAI's text-embedding-3-large model to obtain vector embeddings for news articles. This model accepts articles that are up to 8191 tokens (about 6000 English words) long and produces 3072-dimensional contextual embeddings which capture nuanced semantic features about them. We then compute cosine similarity between all pairs of articles in a day and represent them as nodes in a network where the edge weight between them is equal to their similarity. By dropping edges below an empirically-determined threshold of 0.8, we are left with densely connected components which represent cohesive article clusters around major news events. Most critically, this method will not constrain the quantity or size of any cluster, if there are many closely related articles one day the top cluster can be huge, if there is a dispersed set of events one day there can be many more clusters.

We refine these clusters in a second step by first using an LLM to assign them thematic titles based on the headlines of their constituent articles (see Figure \ref{fig:event_title_prompt} for the prompt). This list of themes covers the major events that happened on that particular day. Then, for each article that has not been assigned to a cluster already, we ask the LLM to read its headline and assign it to one of the event themes for that day if it is relevant (see Figure \ref{fig:cluster_recall_prompt} for the prompt). This step helps improve the coverage of the event clusters to ensure that no related articles are left out if their similarity was slightly below the threshold used to generate the clusters. Similarly, to improve the precision of articles already assigned to clusters, we ask the LLM to compare their headlines with the other articles within their cluster and judge whether they should remain a part of this set (see Figure \ref{fig:cluster_precision_prompt} for the prompt). The cluster assignments after this refinement step yield our final event clusters for that day.

In addition to identifying events, we also extract the top facts around each event by tokenizing all articles within an event into sentences and keeping only those that have been labeled as factual statements in our sentence labeling step. We then repeat the above embedding-based graph clustering process at the fact level with a threshold of 0.85. Frequently repeated facts appear as large clusters of highly similar statements, and we summarize each of these into a synthetic fact using GPT-4o-mini for use in the events dashboard (see Figure \ref{fig:summary_fact_prompt} for the  prompt). These summary facts capture the most important pieces of information about a particular event and are used to study which facts publishers select to publish and which ones they leave out. Figure \ref{fig:event_clustering} shows a sample of the output from our event clustering pipeline. 

\subsection{Data Validation}
\label{sec:data_validation}

To evaluate the quality of data generated by our system, we built a custom data validation platform with dedicated interfaces for each dimension of the data: article labels, sentence labels, event and clusters. Given the inherent subjectivity in news analysis---where multiple valid interpretations can coexist without a single ``correct" answer---we adopted a quality-focused validation approach rather than traditional inter-annotator agreement studies. While having content independently labeled by humans and comparing it with the LLM's output can offer valuable insights, it's important to acknowledge that even individuals with similar backgrounds may often disagree with each other. This is especially true for complex tasks---such as reading a full article and assessing its political lean and tone, or extracting factual content from it---where achieving high inter-annotator agreement is inherently difficult \cite{mitchell2018distinguishing}. Our methodology therefore emphasizes human validation of GPT's labels, where annotators are shown the model outputs and asked to rate their level of agreement rather than independently labeling the content. Although showing raters the model outputs can result in anchoring bias, our approach more directly captures the experience of a user viewing a dashboard and deciding if the labels make sense or not. Because in many cases more than one label is plausible for any piece of content, it is more relevant to the user's experience that the model pick an answer that the user regards as reasonable than whatever answer the user would have picked in the (unobserved) counterfactual in which the dashboard is absent. 
This approach is also supported by recent research that demonstrates LLMs excel at a variety of text labeling tasks that require contextual understanding and subjective judgment \cite{ziems2024can}, and in some cases can even outperform crowdworkers on quality and consistency \cite{gilardi2023chatgpt}. Our own inspection of the labels and explanatory analyses generated by GPT-4o showed that the latest generation of LLMs also posses a very thorough understanding of nuances in news articles as well as their connection to current political events, figures, ideologies, and talking points. 

We hired a team of in-house annotators with backgrounds in communication, psychology, and computer science to validate GPT's outputs for the different dimensions of the data. For validating article lean and tone labels, which are the most challenging aspects for both LLMs and humans to measure properly, we hired three PhD students in communications who had a deep understanding of the media and the current U.S. political landscape. For the other tasks, the annotators were a mix of university students and staff with bachelor's or master's degrees. For all data points, we asked the annotators to first read the article itself and then read the LLM's output and rate its quality or correctness. In cases where they disagree, they can select a more appropriate label. Topic and subtopic labels are assessed regularly so we can update our master list accordingly. For article lean and tone, sentence type, tone, and focus, and event clusters, we conducted one-time in-depth data validation exercises to benchmark GPT-4o's performance at these tasks. In the following sections, we discuss results on each of these.

\subsubsection{Topics and Subtopics}
\label{sec:topic_validation}

Our annotators review topics and subtopics daily to keep up with the ever-changing news cycle. At the end of each week, we re-evaluate the model's accuracy across all features and implement the necessary adjustments. If the accuracy falls below acceptable levels, we conduct a deep-dive into that category and adjust our topic master list or prompting strategies. On average, we verify 150 articles' topic and subtopic classification every day. Each annotator is presented with the article and the topic/subtopic generated by GPT-4o-mini for each article. They are then asked to verify if the generated topic accurately captures the article's content. If it does not, they provide an alternative topic/subtopic from our master list or suggest a new one themselves, which we then review continuously. See Figure \ref{fig:topic_val} for the full task interface.

To validate our classification pipeline, we evaluated the models' error rate based on the annotators' feedback during June 2025, covering over 2,500 articles.  We found a topic-level accuracy of 86.5\% and a subtopic-level accuracy of 83.1\%. The full topic and subtopic classification confusion matrices are shown in Figures \ref{fig:topic_confusion} and \ref{fig:subtopic_confusion}, respectively. In most cases, we found that misclassifications were not outright wrong, but that the article existed at the intersection of multiple topics and could reasonably have been classified as either. For example, an article on the involvement of the US in a foreign conflict might be classified as about ``Foreign Policy" but also as about ``War and International Conflict." We have continued to evolve our news topic and subtopic hierarchy in order to accommodate shifts in the news cycle as well as performing back-updating to make it more robust to future changes.

\subsubsection{Article Lean and Tone}
\label{sec:lean_tone_validation}

We validated article lean and tone labels with ratings from three PhD students in communication and media studies. We divided our 11-point Likert scale for lean and tone into five buckets each ([-5, -4], [-3, -2], [-1, 0, +1], [+2, +3], [+4, +5]) and sampled 2 articles from each of the 25 possible lean-tone bucket combinations while ensuring a balanced distribution over the underlying 11-point values as well, yielding 50 articles that our annotators read. 
For each article, the annotators were asked how strongly they agreed or disagreed with GPT's natural language analysis of the article's lean. They were then shown GPT's numerical score and asked to input what score they would assign the article if they disagree. We repeated the same process for the tone labels, where they first evaluate GPT's analysis and then the assigned score. Finally, we showed them GPT's analysis and choice for the news type label and ask them to select their preference if they disagreed. The full interface for both of these tasks can be seen in Figure \ref{fig:lean_tone_val}, and the codebook provided to the annotators can be found in Text S2.

As shown in Table \ref{tab:validation_results}, we found a high level of agreement for both lean and tone. For both dimensions, our annotators agreed with GPT's understanding of the article more than 70\% of the time. Ambivalent responses (neither agree or disagree) and disagreement were rare, totaling roughly 6\% for tone and 14\% for lean. More detailed confusion matrices showing the comparison between the human and GPT ratings can be seen in Figure \ref{fig:lean_tone_confusion}.
We also asked the annotators for their feedback on GPT's performance after the task and received positive reviews: \textit{``It definitely seems like GPT is more than capable of doing this labeling. Maybe one time that it fundamentally seemed to not understand something. In most where I disagreed with the label, it was more so that the labels didn’t seem to apply, e.g. some things that had nothing to do with politics getting a non-zero lean."} and \textit{``\dots overall, I think it performed really well. Any differences in the scale that were only off by one point don’t feel significant to me."} In addition to this annotation exercise, to understand which themes in the articles the model picks up on when assigning a certain lean rating, we asked GPT-4o to summarize its own explanations for 50 articles in each of the above-mentioned five buckets. The results of this experiment are shown in Text S3, where we can see that it in fact picks up on nuanced thematic and stylistic patterns that align remarkably well with the political lean spectrum.

\begin{table}[htbp]
\centering
\renewcommand{\arraystretch}{0.7}
\begin{tabular}{>{\arraybackslash}p{2.5cm}>{\arraybackslash}p{5cm}>{\centering\arraybackslash}p{1cm}>{\centering\arraybackslash}p{1cm}>{\centering\arraybackslash}p{4cm}}
\multicolumn{2}{l}{\textbf{Articles}} & \textbf{Mean} & \textbf{SD} &  \\
 
\midrule

\multirow{5}{=}{{Tone}} & \textit{Agree} & 0.780 & 0.072 & \barwithsd{0.780}{0.072} \\
& \textit{Somewhat Agree} & 0.153 & 0.103 & \barwithsd{0.153}{0.103} \\
& \textit{Neither Agree nor Disagree} & 0.020 & 0.020 & \barwithsd{0.020}{0.020} \\
& \textit{Somewhat Disagree} & 0.047 & 0.050 & \barwithsd{0.047}{0.050} \\
& \textit{Disagree} & 0.000 & 0.000 & \barwithsd{0.000}{0.000} \\
\midrule

\multirow{5}{=}{{Lean}} & \textit{Agree} & 0.733 & 0.012 & \barwithsd{0.733}{0.012} \\
& \textit{Somewhat Agree} & 0.127 & 0.076 & \barwithsd{0.127}{0.076} \\
& \textit{Neither Agree nor Disagree} & 0.033 & 0.031 & \barwithsd{0.033}{0.031} \\
& \textit{Somewhat Disagree} & 0.060 & 0.072 & \barwithsd{0.060}{0.072} \\
& \textit{Disagree} & 0.047 & 0.046 & \barwithsd{0.047}{0.046} \\
\midrule

\multirow{2}{=}{{Type}} & \textit{Agree} & 0.907 & 0.090 & \barwithsd{0.907}{0.090} \\
& \textit{Disagree} & 0.093 & 0.090 & \barwithsd{0.093}{0.090} \\
\midrule

\\[1pt]
\multicolumn{5}{l}{\textbf{Sentences}} \\
\midrule

Type & & 0.887 & 0.061 & \barwithsd{0.887}{0.061} \\
Tone & & 0.970 & 0.026 & \barwithsd{0.970}{0.026} \\
Focus &  & 0.903 & 0.032 & \barwithsd{0.903}{0.032} \\
\midrule

\\[1pt]
\multicolumn{5}{l}{\textbf{Event Clusters}} \\
\midrule

Precision & & 0.981 & 0.021 & \barwithsd{0.981}{0.021} \\
Recall & & 0.956 & 0.028 & \barwithsd{0.956}{0.028} \\
F1 & & 0.966 & 0.024 & \barwithsd{0.966}{0.024} \\
\bottomrule
\end{tabular}

\caption{\textbf{Data Validation Results.} We observed a high level of agreement with the LLM's outputs across all three annotators. For article tone and lean, our annotators expressed clear agreement with GPT-4o's analysis in more than 70\% of cases. Disagreement and ambivalent responses were much rarer, totalling only 14\% for lean and 6\% for tone (which is less open to interpretation and personal opinion than partisan lean). For sentences, the annotator agreement with GPT-4o is higher still, with the LLM performing best at identifying the tone (97\%). Finally, our event clustering pipeline achieves both very high precision and recall, which results in highly cohesive clusters.}
\label{tab:validation_results}
\end{table}

\subsubsection{Sentence Type, Tone, and Focus}
\label{sec:sentence_validation}

To validate sentence labels, we used the same 50 articles from the task before and asked a separate set of annotators to go sentence-by-sentence, read the LLM's assigned label and select their preferred choice if they disagree with it. These 50 articles resulted in a total of 2190 sentences, for an average of 43.8 sentences per article. The full task interface is shown in Figure \ref{fig:sentence_val} and the codebook provided to the annotators can be found in Text S4. The results of this exercise are shown in Table \ref{tab:validation_results}. Most sentences in news articles are neutral facts that do not mention any particular political party, although this varies for different article types and topics. Over this distribution, we see that the LLM's accuracy is highest for sentence tone at 97\%, followed by focus at 90\% and type at 88\%. 
More detailed confusion matrices showing the comparison between the human and LLM labels can be seen in Figure \ref{fig:sentence_confusion}. We also compared the independently generated sentence- and article-level labels and found a high level of agreement between them, as can be seen in Figure \ref{fig:sent_art_comparison}. The distribution of fact and opinion sentences varies accordingly with the type of article, with opinion pieces having more opinion sentences than news reports and vice versa for facts. Similarly, the average tone of sentences in an article exhibits a positive, monotonically increasing relation with the overall tone of the article itself.

\subsubsection{Event Clusters}
\label{sec:event_validation}

To evaluate the accuracy of our event event clustering pipeline, we asked the three annotators to read the headline of each article from a cluster and determine whether it has been assigned to the appropriate event or not. If it has been assigned incorrectly, they select the event to which they think it is most relevant. Figure \ref{fig:event_val} shows the full interface for this task. We picked October 1, 2024, which was the date of the US Vice-Presidential Debate and thus represented a mix of major and minor events being covered. We had a total of 791 articles distributed across 22 events on that day, ranging from new escalations in the Israel-Iran conflict (75 articles) and the VP debate (65 articles) to Julian Assange's plea and a Verizon outage (3 articles each). The full list of events can be found in Table \ref{tab:event_stats}.

We used precision and recall to evaluate the quality of the clusters, where higher precision means that there are no unrelated articles in the cluster, and higher recall means that the cluster covers all the article about that event. Table \ref{tab:validation_results} shows the results of this exercise. Due to the high cosine similarity threshold used in our framework, we ensure that only highly related articles are clustered together, which results in a precision of 0.98. For recall, the automated prompt-based cluster refinement essentially mimics a human review phase and ensures a high recall of 0.96 by expanding the event coverage to include any articles that might be related.

\section{Illustrative findings}
\label{sec:findings}

In this section, we apply our curated dataset from 2024 to a handful of questions about media practices and biases. Due to length constraints, we do not attempt to answer these questions comprehensively, but rather simply use these exercises to illustrate the richness of the dataset and demonstrate its potential serve as a resource for future research. It's important to emphasize that the analyses presented are just samples of the overall data. The comprehensive nature of our dataset allows us to answer much more focused questions by examining different subsets, enabling targeted investigations into specific aspects of media coverage, topical biases, or temporal trends.

\subsection{The imbalance between policy and election horse race coverage}
\label{sec:selection_bias_finding}

\begin{figure}[h]
    \centering
    \includegraphics[width=\linewidth]{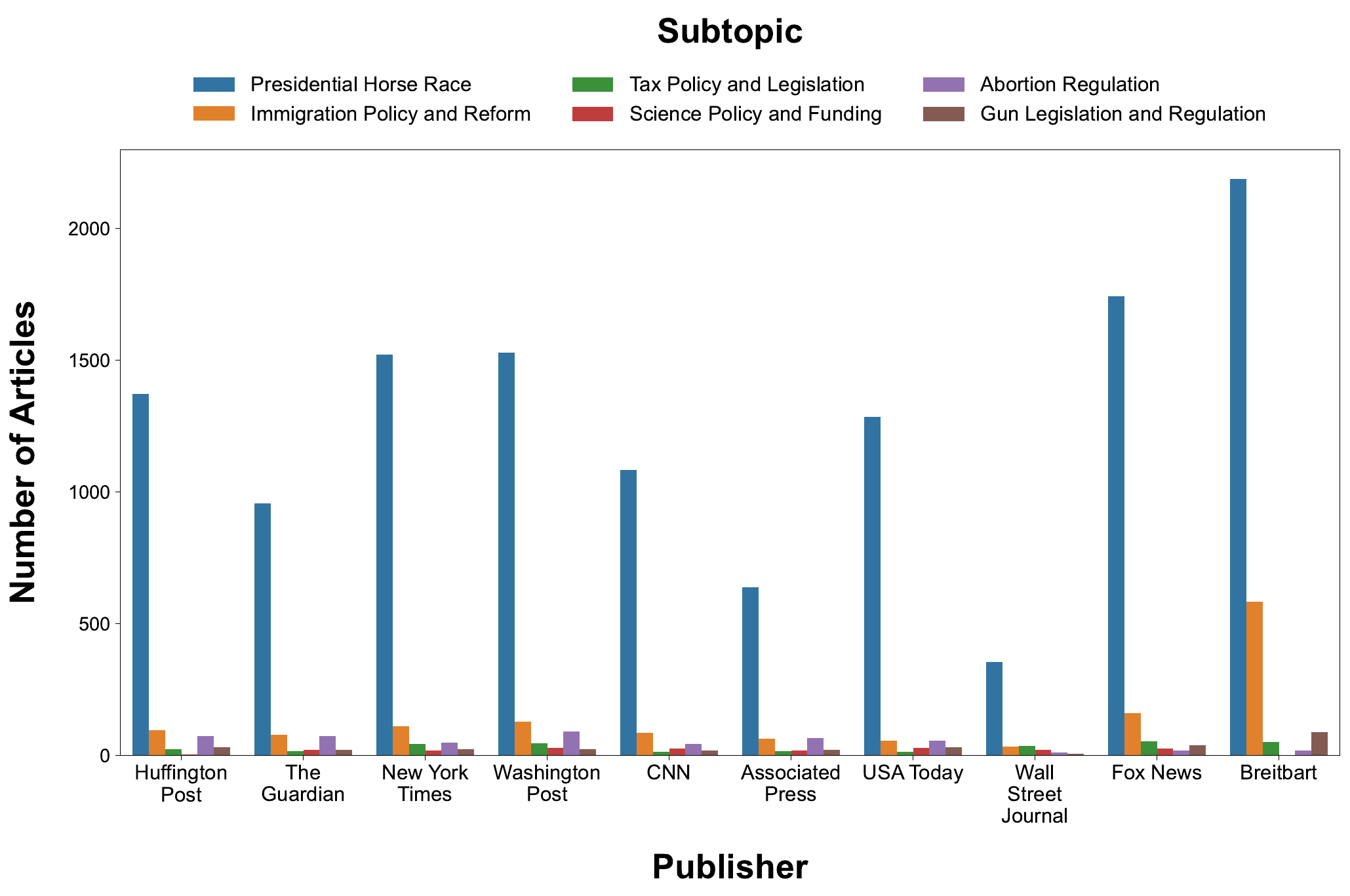}
    \caption{\textbf{The imbalance between policy and election horse race coverage.} The media primarily covered the election as a horse race instead of focusing on more substantive policy discussions to inform their readers of each candidate's and their party's agenda. Over 8,000 articles were published about the horse race across all publishers in 2024, compared to only 3,000 about all the policy topics put together (a significant fraction of which come from Breitbart's focus on immigration).}
    \label{fig:selection}
\end{figure}

News organizations operate under constraints of limited time and space, meaning they need to make careful choices about which stories to cover. This gatekeeping power has been extensively studied in the social sciences, particularly under the framework of agenda setting, which posits that while news media may not directly dictate what people think, they significantly influence what people think about \cite{agenda_setting_1972, mccombs2005agenda}. In making editorial decisions, news organizations assess the newsworthiness of events based on various factors such as relevance, impact, timeliness, prominence, and proximity \cite{harcup2017news}. This assessment of newsworthiness is particularly crucial given that news organizations are often private entities driven by profit or political motives \cite{amos2023media}. Furthermore, these editorial choices have a major impact on the readers, whose understanding of the world around them is shaped by the news they consume. Ideally, news organizations should help voters make informed choices by informing them on policy discussions. However, existing literature has highlighted that media outlets tend to focus predominantly on the "horse race" elements of elections—emphasizing candidates' polling numbers, strategic positioning, and projected wins— which can overshadow crucial policy issues \cite{iyengar2004consumer, vliegenthart2010covering}.

Our data offers a more comprehensive quantitative analysis than has previously been possible of the coverage of horse race dynamics compared with other critical policy issues in the lead-up to the 2024 US presidential election. Specifically, using our topic and subtopic classification system, we carried out a detailed comparison of the amount of coverage dedicated to the presidential horse race versus  policy discussions. Figure \ref{fig:selection} illustrates this disparity, showing that the coverage of the presidential race overwhelmingly dominated all news outlets, even receiving more attention than all policy issues combined. It also reveals how specific news values adapt by catering to different target audiences \cite{uscinski2014people, welbers2016news}. For example, outlets like Breitbart, and to a lesser extent Fox News, allocate substantially more coverage to Immigration Policy and Reform in comparison to left-leaning media such as CNN or The Guardian.

\subsection{Political lean varies across news topics}
\label{sec:political_lean_finding}

As discussed earlier, news producers can also influence public perception through their framing of issues\cite{chong2007framing} defined broadly as choices in language and context as well as the inclusion or omission of certain details and perspectives \cite{gentzkow2006media, chong2007framing, entman1993framing}. Here we study lean---the extent to which a news article supports Democratic or Republican viewpoints and policies---which, as described above, can indicate framing bias. Current methods typically reduce news publishers to this single metric by assigning them a political lean rating that ranges from ``Left" to ``Right" \cite{allsides, mediabiasfactcheck, groundnews}. While these static classifications are useful, they often fail to capture the nuances of bias within a single publisher, across various topics, or even from one article to another. Understanding how partisan alignment shifts depending on the subject matter can significantly enhance our comprehension of media dynamics and discourse.

\begin{figure}[t]
    \centering
    \includegraphics[width=\linewidth]{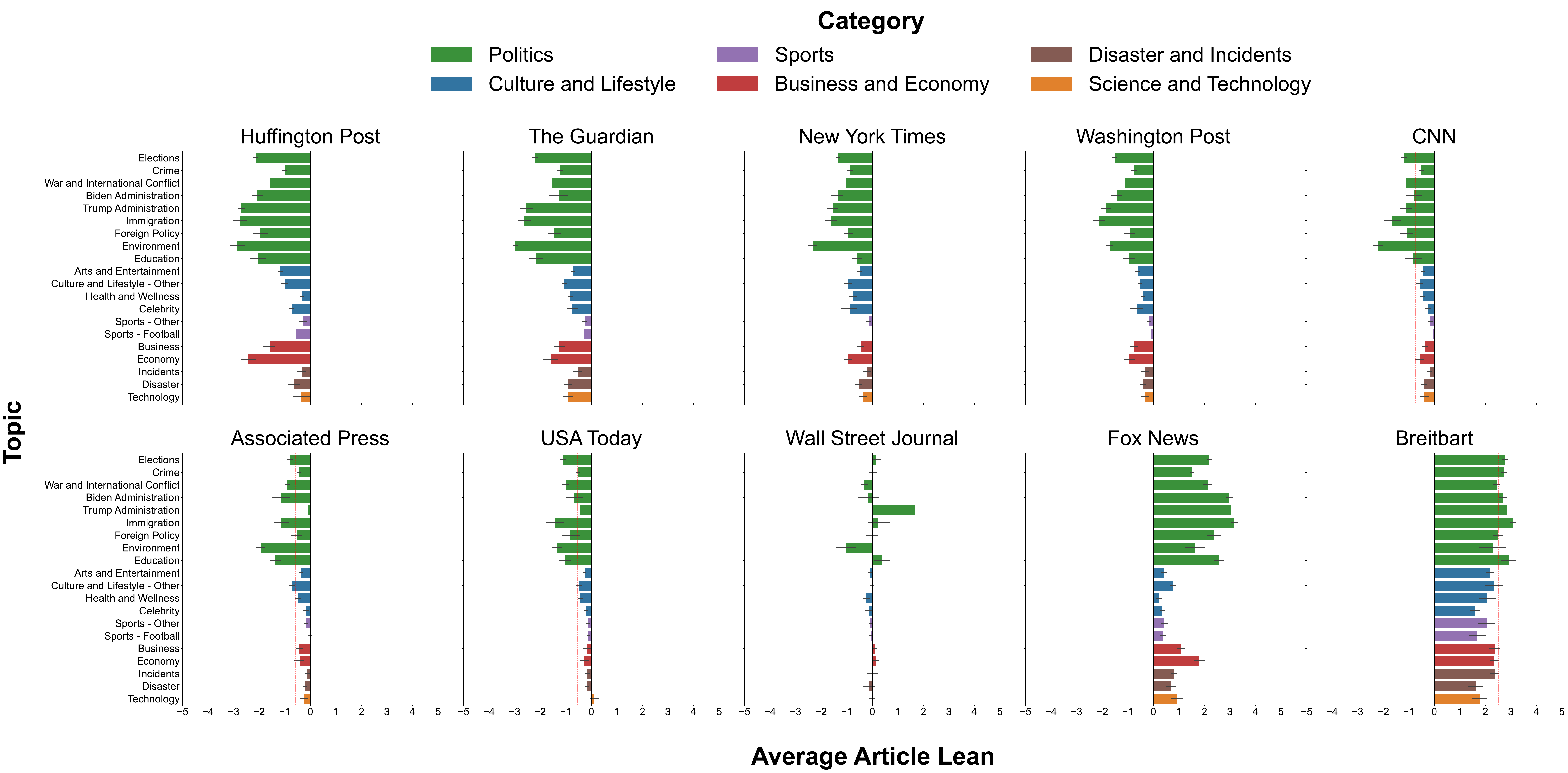}
    \caption{\textbf{Political lean varies across news topics.} This figure illustrates the average (signified by the vertical red line) partisan bias of selected news outlets, categorized by topic. Negative values indicate a Democratic lean, whereas positive values suggest a Republican lean. The red line shows the average lean for a publisher across all topics, and the diversity in lean across different topics highlights the nuanced and topic-dependent nature of media bias. Variations in the degree of political alignment provide insights into each outlet’s editorial stance and the complexity of bias representation in media coverage.}
    \label{fig:bias_breakdown}
\end{figure}

As described in Section \ref{sec:article_labeling}, \mbd\  provides lean labels at the article level (focusing on the substance of the articles, these labels ignore headlines, which we address in Section \ref{sec:headling_finding}), allowing for a more nuanced analysis of media outlets' partisan biases. Figure \ref{fig:bias_breakdown} presents the average political lean across publishers and various topics, with negative values indicating a Democratic lean and positive values indicating a Republican lean. Although the overall pattern aligns with commonly held perceptions of these publishers' partisan biases, it also reveals intriguing insights into the subtleties of media partisanship. This granular approach uncovers variations in bias not only across different outlets but also within publishers on specific topics, offering a more detailed understanding of how partisan framing can shift.

As depicted in Figure \ref{fig:bias_breakdown}, it is generally true that an outlet’s stance on one issue tends to correlate with its stance on others; however, the extent of this alignment can vary considerably depending on the topic. For instance, the New York Times exhibits a consistently left-leaning position on topics such as the environment, whereas it maintains a relatively neutral stance on technology. In contrast, Fox News presents a clear right-leaning stance on both topics.  

Our data also reveal that The Wall Street Journal is the most neutral outlet, maintaining a relatively balanced stance on most topics. Notably, it is the only publisher with a left-leaning stance on certain political topics, such as the environment, while exhibiting a right lean on others, like the Trump administration. Additionally, less political topics such as sports, travel, technology, and health and wellness are generally treated neutrally across most publishers. Nevertheless, Breitbart maintains its right-leaning stance even on these inherently less partisan topics.

\subsection{Most news is negative, especially Politics}
\label{sec:negativity_finding}

As discussed earlier, the tone in which stories are presented can be another indicator of framing bias. Prior work has found that coverage tonality---how positively or negatively an attitude object is evaluated---provides readers with templates for understanding politics \cite{eberl2017one} and influences the salience of the negative and positive elements within the public debate \cite{vreese2003valenced}. For example, coverage tonality around political actors in Europe affects party preferences \cite{eberl2017one}, while in economic reporting, news sentiment predicts macroeconomic variables like consumption and output \cite{shapiro2022measuring}.

\begin{figure}[t]
    \centering
    \includegraphics[width=\linewidth]{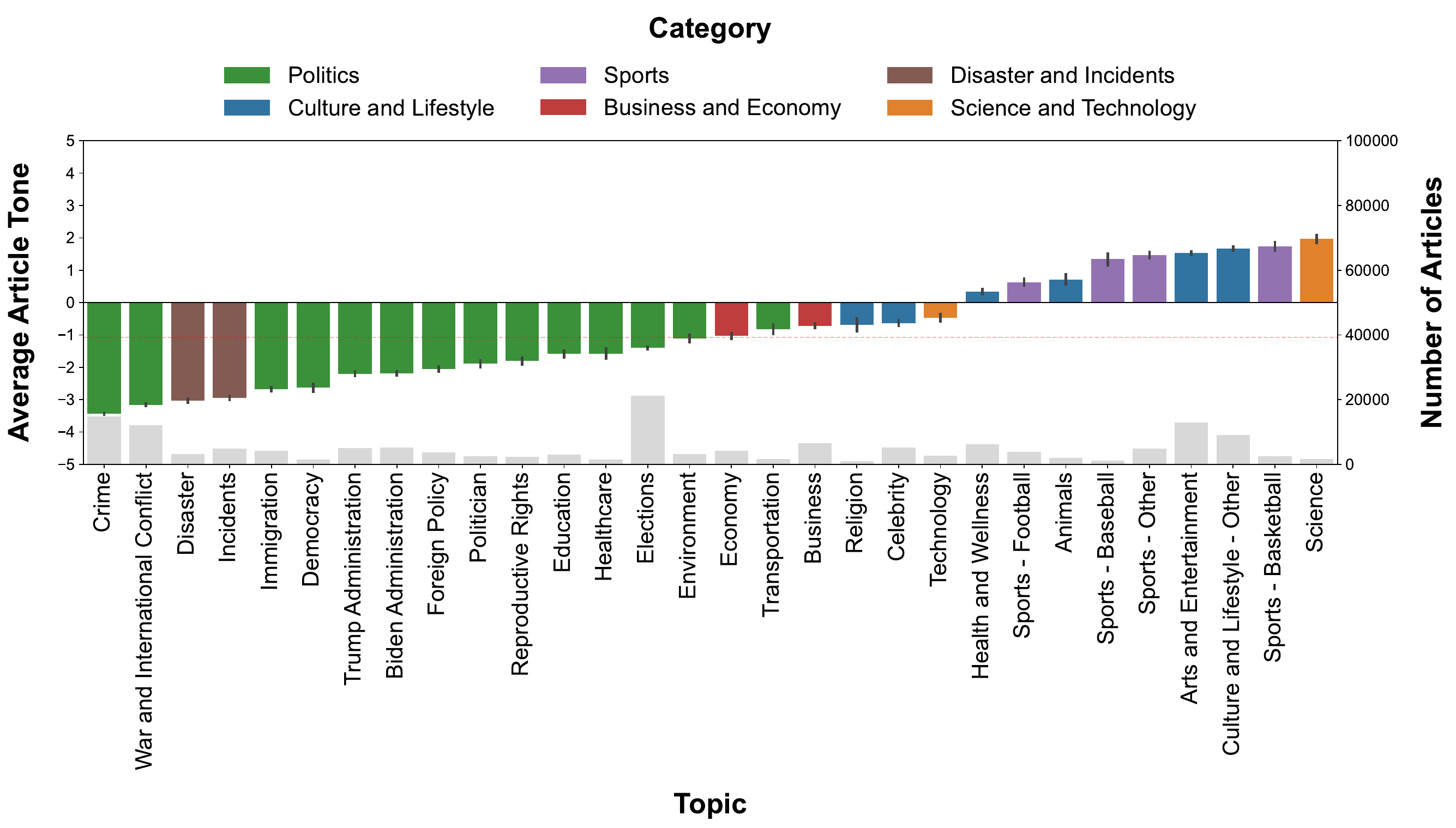}
    \caption{\textbf{Most news is negative, especially Politics.} The colored bars represent the average tone of articles that fall in the top 30 topics by number of articles (shown in gray), sorted from most negative to most positive. Negative tone is prevalent, both by number of topics and by the total volume of articles. Except for news about weather and natural disasters, all of the remaining topics that are more negative than the average (red line) are related to politics, highlighting the pervasive nature of negative elements in media coverage.}
    \label{fig:tone}
\end{figure}

Here we study an important special case of tone potentially playing the role of framing bias in news---namely negativity bias, which has been extensively documented in both the production and reception of news \cite{Lengauer_Esser_Berganza_2012}. 
This focus on the negative could arise from both the selection and framing of news. On the selection side, it often involves prioritizing stories inherently perceived as negative, such as those involving conflict, crime, and disasters \cite{Van_Der_Meer_Hameleers_2022}. Additionally, the framing of certain stories can enhance negativity through the deliberate use of specific words and phrases. The tone classifications in our dataset capture both of these aspects, providing a comprehensive view of how negativity is embedded in news content.

Prior research has primarily focused on studying news negativity within specific domains, such as politics \cite{esser2016negativity} or the economy \cite{svensson2017good}. Our data allows us to quantify the tone of news coverage across a wider range of topics and publishers than has previously been possible. 
Figure \ref{fig:tone} depicts the average tone (left y-axis) of articles across topics as well as the number of articles on that topic (right y-axis). We find that the majority of the news content analyzed contained negative elements, aligning with trends that are well-documented in the literature \cite{Lengauer_Esser_Berganza_2012, Van_Der_Meer_Hameleers_2022}. However, we also extend this finding by showing how negative tonality is particularly pronounced in political coverage, which also receives the highest volume of coverage. This combination of high negativity and high visibility in political news has important implications for democratic discourse, as it may contribute to political cynicism and disengagement \cite{esser2016negativity}. Moreover, our cross-topic analysis demonstrates where this negativity bias extends beyond politics (e.g. religion and economy), suggesting a systemic feature of contemporary news production that affects how the public perceives various aspects of social life. 
 
\subsection{Headlines are not always faithful to their articles}
\label{sec:headling_finding}

A headline is arguably the most important part of any news story, setting the stage for the rest of the article. Indeed, for many casual readers, the headline might be their only take away about an event, an effect that is exacerbated by the news feeds common to social media platforms that focus almost exclusively on headlines \cite{searles2023scrollability}. Consequently, any misalignment in lean or tone between news headlines and the accompanying articles raises concerns \cite{chesney2017incongruent} for at least two reasons. First, many individuals decide to share information based solely on the headline, without reading the full news article. A recent study on social media sharing demonstrated that more than 75\% of the news articles shared on Facebook did not click before sharing\cite{sundar2024sharing}. Second, even when people do read the entire article, the headline continues to exert a significant influence on opinion formation. If the information presented in the headline is incongruent with the article's content, individuals often struggle to adjust their initial impressions \cite{carcioppolo2021exaggerated, ecker2014effects}. Furthermore, this bias can affect not only the reader's ability to make accurate inferences from a news story, but also their capacity to recall the facts presented within it \cite{ecker2014effects}.

\begin{figure}[t]
    \centering
    \subfloat{
        \includegraphics[width=0.47\linewidth]{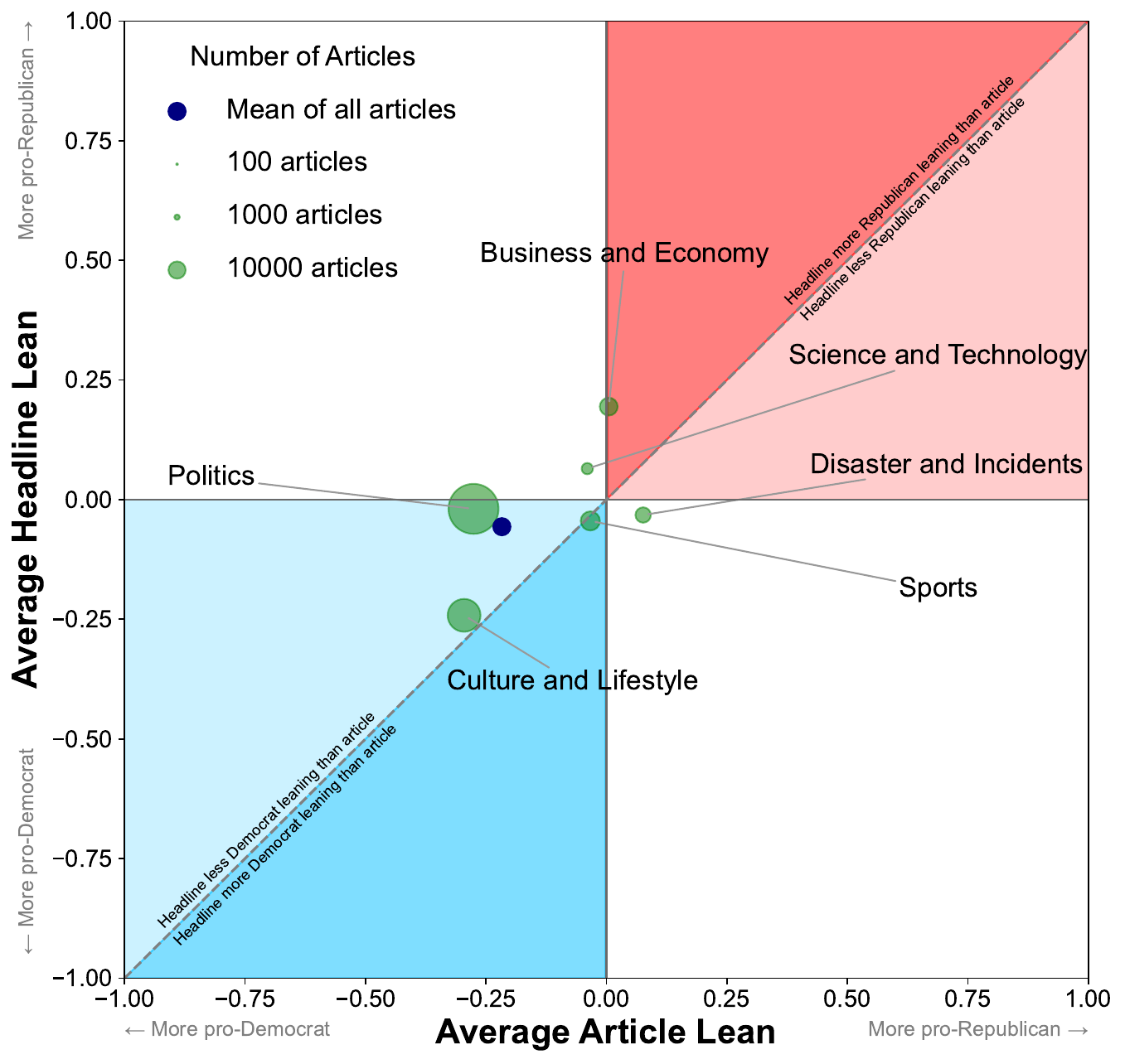}}
    \hfill
    \subfloat{
        \includegraphics[width=0.45\linewidth]{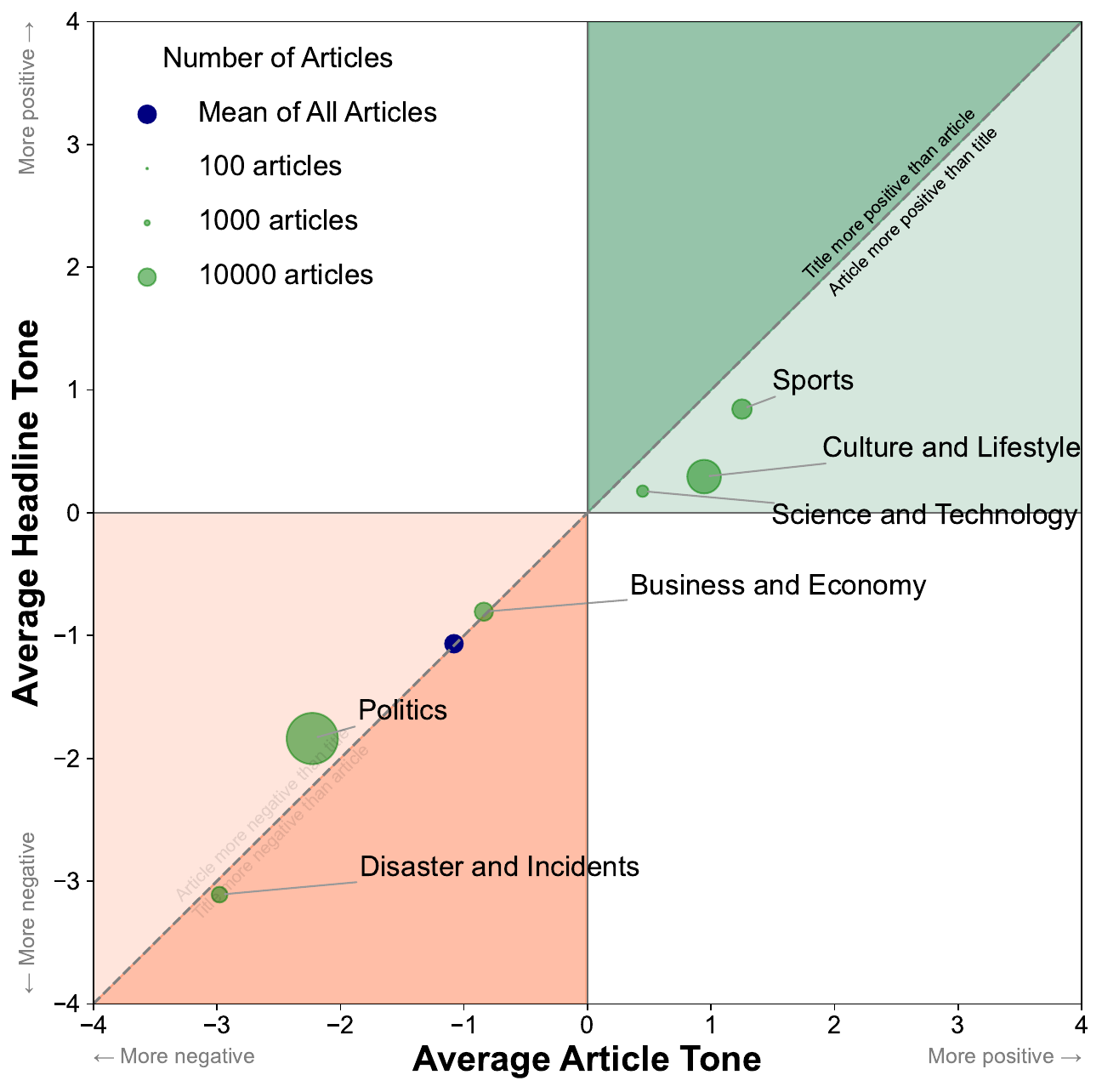}}
    \caption{\textbf{Headlines are not always faithful to their articles.} Each data point represents the average headline lean/tone (on the y-axis) and the average article lean/tone (on the x-axis) where positive values of lean refer to more pro-Republican stances and negative values refer to more pro-Democrat stances. All articles are labeled in the range [-5, +5], and the axes are adjusted separately in these plots for clarity. While the average for all articles for tone lies perfectly on the the identity line, showing that headlines are as negative or positive as the articles on average, the case for lean is different: headlines are slightly more pro-Republican than their articles. This is most pronounced in the case of articles about business and economy, where headlines lean more strongly towards Republicans compared to the articles.}    
    \label{fig:lean-tone-comparison}
\end{figure}

The separate labeling of headlines and article texts in our dataset enables us to explore this phenomenon across both tone and political lean dimensions. Identical prompts were used for analyzing both headline and article text. Figure \ref{fig:lean-tone-comparison} presents the results, broken down by news category, with the left Figure \ref{fig:lean-tone-comparison}a illustrating political lean and the right Figure \ref{fig:lean-tone-comparison}b illustrating tone comparison. In Figure \ref{fig:lean-tone-comparison}a, the average headline lean is represented on the y-axis, and the average article lean on the x-axis, with positive values indicating an increasingly pro-Republican lean (red) and negative values indicating an increasingly pro-Democrat lean (blue). The diagonal line represents cases where the headline and article lean are equivalent, while darker shading indicates regions where headlines have a stronger lean than the articles, and lighter shading indicates the inverse. The dark blue marker shows the mean across all articles, and the size of each scatter point reflects the number of articles within that category.

Figure \ref{fig:lean-tone-comparison}a reveals that news headlines systematically lean more pro-Republican than the articles themselves, as evidenced by the position of the dark blue marker—the overall average—being above the diagonal. This pattern is most pronounced in the ``Business and Economy" category, where articles are fairly neutral overall but are often accompanied by headlines that lean more Republican. Similarly for the ``Politics" category, the articles lean slightly pro-Democrat but the headlines are fairly neutral. Figure \ref{fig:lean-tone-comparison}b illustrates the comparison for tone, with average headline tone depicted on the y-axis and average article tone on the x-axis. While the overall mean lies close to the diagonal, suggesting similar average tone between headlines and articles, this aggregate measure masks substantial heterogeneity across topics. That is, topics with inherently positive content (sports, culture and lifestyle, and health) tend to have headlines that are less positive than their corresponding articles. Conversely, for typically negative topics like politics and disasters, headline tone either aligns with or is slightly more positive than the article text. Comparing headline and article labels on both of these dimensions also shows that the observed pro-Republican lean is not due to differences in length or style of headlines compared to the full articles (since the same shift is not seen in tone), but possibly a conscious editorial decision to frame the news a certain way.

\subsection{The media attacks the opposition more than defending their side}
\label{sec:focus_lean_finding}

\begin{figure}[t]
    \centering    
    \includegraphics[width=0.8\linewidth]{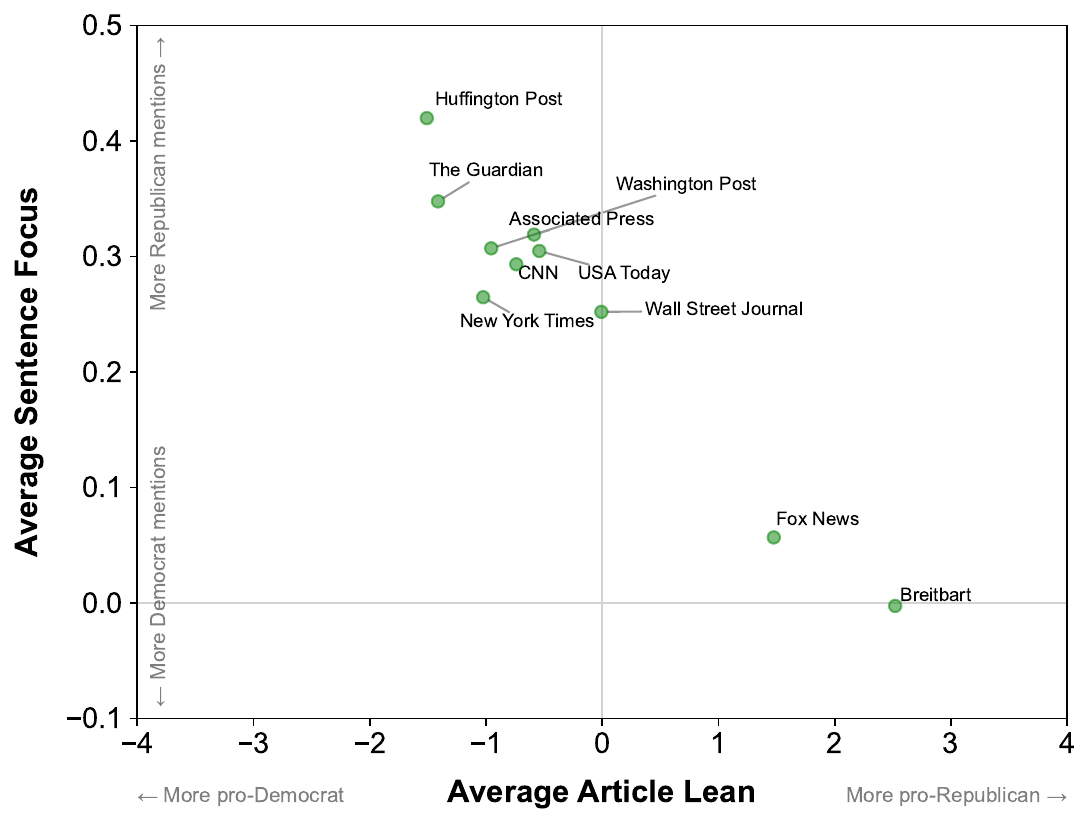}
    \caption{\textbf{The media attacks the opposition more than defending their side.} Association between publishers' political lean  and coverage of each political party. The x-axis shows the average political lean of all articles by a publisher (positive values represent more pro-Republican lean and negative values represent more pro-Democrat lean) and the y-axis shows the ratio of sentences that mention Republicans (positive) vs. Democrats (negative). The more a publisher talks about a particular party, the more they lean towards the opposition.}
    \label{fig:lean-focus}
\end{figure}

In today's politically charged media environment, partisan news outlets employ a variety of tactics to promote their ideological agenda. One intriguing example that has been studied previously is the tendency of partisan media to focus more on critical coverage of opposing political groups than on positive coverage of their ideological allies  \cite{smith2014let}. Additionally, it has been observed that politicians' denunciations of the news media can heighten perceived biases, irrespective of the politicians' party affiliations \cite{smith2010politicians}. This behavior raises important questions about the role of media in shaping public discourse, potentially fueling partisan polarization and influencing how audiences perceive political legitimacy and the opposition.
Our data can address this question by applying a combination of focus and lean measures at scale. 
Specifically, we compare the political lean of articles—which measures the extent to which each article supports the viewpoints and policies of one party over the other—with the focus of the corresponding article. As noted earlier, focus is measured on a sentence-by-sentence basis, indicating whether each sentence references the Democratic Party, the Republican Party, both, or neither. However, we can compute an article level focus score by averaging the individual sentence scores, with Republicans coded as +1, Democrats as -1, and neither or both as 0. 
Figure \ref{fig:lean-focus} illustrates the relationship between average lean and focus for the 10 publishers included in our dataset. The figure reveals a negative correlation between focus and lean, indicating that more Republican-leaning publishers tend to speak more about Democrats and vice versa. This pattern is consistent across outlets: the most left-leaning publishers, such as the Huffington Post and The Guardian, frequently focus on the Republican Party or its politicians, while right leaning outlets like Fox News and Breitbart exhibit a greater focus on Democrats. Interestingly, Figure \ref{fig:lean-focus} also reveals that the average focus across all outlets is skewed more towards Republicans.

\section{Discussion}
\label{sec:discussion}

In this paper, we have introduced The \mbd, which leverages LLMs to analyze news media at scale and in near-real time to support and accelerate the research on media bias. Our work builds on broader efforts to quantify the impact of news on democracy \cite{watts_democracy}, in particular by constructing robust, scalable data infrastructure for studying how news moves from production to consumption and absorption. In addition, \mbd  exemplifies the related goal of building web-based interactive visualizations to communicate insights to nonacademic stakeholders \cite{wang2025media, watts_democracy}. By making our data and dashboards widely accessible, we hope to enable data-driven transparency in journalism, allowing stakeholders---including policymakers, educators, and media professionals---to engage with the findings in meaningful ways.

The \mbd\ complements existing media datasets such as PeakMetrics, GDELT, and Media Cloud that already provide researchers with high-volume collections of online news articles, enabling large-scale studies of media coverage \cite{roberts2021media}. Although these datasets collectively constitute a valuable resource, like all such efforts they suffer from limitations. As noted earlier, for example, scraping thousands or tens of thousands of websites daily or faster over long periods inevitably introduces numerous problems with data missingness and quality. Articles that are easier to extract---such as those from publishers without paywalls---tend to be overrepresented, while publishers with sophisticated paywalls are more sparse. Furthermore, large-scale scraping efforts generally treat content across the entire website equally, regardless of whether it is featured prominently on the homepage or buried deep within the category pages. 
In contrast, the Media Bias Detector focuses on effectively capturing the \emph{top} stories from each publisher's homepage at regular intervals with high reliability. As such, our approach prioritizes capturing stories that receive significant exposure, creating opportunities to study how publishers shape public focus through their editorial choices.

Symmetrically, our approach also reveals which topics are absent from the headlines, allowing researchers to study patterns of omission---cases where major political actions, policy decisions, or social issues receive significant coverage from some outlets but are largely overlooked by others. By leveraging LLMs and measuring multiple quantities at the article and sentence level, our approach generates higher quality and more granular labels than has been possible until very recently. Finally, the continuous human-in-the-loop monitoring ensures that our system is able to adapt to the news cycle and avoid concept drift. In this way, we ensure that our data set is always up-to-date with the events and topics being covered in the media and classifies them accurately. While our current focus is on text-based news from mainstream media, our framework could also be adapted to social media by updating our prompts to more effectively label short-form text and by incorporating interaction data and network features such as likes and shares to measure the popularity and influence of various pieces of content.

Naturally, our approach also has important limitations. First, our data collection process does not provide exhaustive coverage of all online news. We limited ourselves to 21 major publishers (10 in the first stage of the project), but there are many more local and online news sources whose coverage may be of interest depending on the research question being studied. Further, by limiting ourselves to the top-ranking news articles from these publishers and only collecting them at regular intervals throughout the day, we may miss articles that either do not show up in the top 20 or do so and disappear before we are able to scrape them. The time frame of our collection is also limited to January 2024 onwards, which limits the depth of temporal analysis that can be done with this data (although efforts are currently underway to backfill beyond 2024). In some cases, we also face issues with paywalls or broken links which leads to  gaps in our coverage. 

Our data labeling and validation pipelines also have limitations. First, we use proprietary models which may become unavailable in the future, requiring us to switch to a different model, thus impacting the consistency of our data. While we used state-of-the-art models at the time of writing, future models may be more capable at these tasks and require re-labeling of past data. Beyond GPT-4o and GPT-4o-mini, we did not compare other models such as Claude, Gemini, LLaMA, DeepSeek, Qwen, or Grok, an exercise which would be helpful in understanding how models from different families and of different sizes understand and categorize news articles' political lean and tone. Our approach also used zero-shot prompts throughout the pipeline and did not attempt to fine-tune any models. For data validation, our approach focused primarily on checking for clearly incorrect outputs in our data, but a more rigorous approach could have been to conduct an independent human annotation exercise and then compare their agreement with the LLMs outputs afterwards.

A final limitation is one that we have already mentioned: that in the absence of clear, objective, ground truth, bias is inherently difficult to measure. In fact, some experts have concluded that the difficulty is so great as to render the detection of content that is misleading on account of bias rather than verifiable falsehoods unsuitable for scientific inquiry~\cite{williams2024misinformation}. Although we are sympathetic to this critique, we remain optimistic that something useful can be accomplished in spite of it. Specifically, our approach of measuring ``elemental'' quantities such as topic, tone, lean, and focus is explicitly designed to avoid making judgments about the ``true" or desired state of the world, and hence to result in more robust labels. For example, it is possible for raters with different biases to agree that an article is about the economy, has a positive tone, talks more about democrats than republicans, and hence is overall favorable to democrats while maintaining very different views about its accuracy, informativeness, and even whether it should have been published at all. 

As previously noted, a downside of this design principle is that we are unable to make direct assessments of article or even publisher bias---a limitation that media critics and political partisans alike may find frustrating. We maintain, however, that any attempt to impose a single notion of truth or objectivity against which bias can measured will itself inevitably be biased, undermining the integrity of the project. Moreover, by measuring many of these elemental quantities both at the sentence and article level, and by allowing analysts the flexibility to aggregate them to yet higher levels (e.g. publisher, groups of publishers, the entire collection, etc.), we believe our approach does generate valuable insights about media bias, albeit indirectly. For example, our demonstration that news publishers devote dramatically more coverage to horse race stories than to policy issues does not settle the question of how much coverage ought to have been devoted to each; but it does call into question public declarations by news organizations that their goal is to inform citizens about the most important issues in an upcoming election~\cite{watts2017don}. Likewise, our demonstration that the political lean of headlines is systematically different from the text of the corresponding stories does not settle which one, if either, is correct; but the difference does show that writers of articles and headlines alike exploit their flexibility to cast the same story in different lights. Finally, our demonstration that article lean is negatively correlated with partisan focus does not determine which of our publishers is more or less biased than the others; but it does identify a possible mechanism (attacking the opposition rather than supporting one's own side) by which all publishers express their bias.

Summarizing, by integrating systematic data collection, LLM-driven analysis, and human-in-the-loop validation, we have laid a scalable foundation for advancing research on media bias. Although no single study can fully encapsulate the complexities of media influence, and no single dataset can comprehensively define media bias, we believe this work represents a significant step forward. This study sets the stage for future research by providing a robust methodology and highlighting key dynamics in media bias. By continually refining our methodology and expanding our dataset to encompass past and future time periods, we aim to enhance both academic understanding and public awareness of media bias dynamics through an empirical, data-driven approach.


\newpage


\clearpage 

%
\bibliography{science_template} 
\bibliographystyle{sciencemag}


\section*{Acknowledgments}
We thank Yuxuan Zhang, Grace Jennings,  Elliot Pickens, and Yifei Duan for developing the data collection and labeling infrastructure, and Ajay Patel and Bryan Li for helpful discussions on the methodology. We also thank Anushkaa Gupta helping organize and coordinate the project. DJW and DR acknowledge their long collaboration with Markus Mobius, during which they worked on pre-LLM methods to cluster news articles and display them visually on a web-based dashboard as part of what was called “Project Ratio,” initially at Harmony Labs and subsequently at Microsoft Research (https://www.microsoft.com/en-us/research/project/project-ratio/). This research was developed with funding from Richard Jay Mack, Knight Foundation (Grant ID G-2023-67859), and the Defense Advanced Research Projects Agency’s (DARPA) SciFy program (Agreement No. HR00112520300). The views expressed are those of the authors and do not reflect the official policy or position of the Department of Defense or the U.S. Government.

\renewcommand{\thefigure}{S\arabic{figure}}
\renewcommand{\thetable}{S\arabic{table}}
\renewcommand{\theequation}{S\arabic{equation}}
\renewcommand{\thepage}{S\arabic{page}}
\setcounter{figure}{0}
\setcounter{table}{0}
\setcounter{equation}{0}
\setcounter{page}{1} 

\clearpage
\begin{center}
\section*{Supplementary Materials for\\ \scititle}


	Samar Haider$^{1\dagger\ast}$,
	Amir Tohidi$^{1\dagger}$,
        Jenny S. Wang$^{2\dagger}$,
        Timothy D\"orr$^{3}$ \\ \and
        David M. Rothschild$^{4}$,
        Chris Callison-Burch$^{1}$,
        Duncan J. Watts$^{1, 3, 5}$ \\ \and
	\small$^{1}$Department of Computer and Information Science, University of Pennsylvania, Philadelphia, USA. \\
	\small$^{2}$Harvard Business School, Boston, USA. \\
        \small$^{3}$Microsoft Research, New York, USA. \\
        \small$^{4}$Annenberg School for Communication, University of Pennsylvania, Philadelphia, USA. \\
        \small$^{5}$Department of Operations, Information, and Decisions, University of Pennsylvania, Philadelphia, USA. \\
    
        \small$^\dagger$These authors contributed equally to this work.\\
	\small$^\ast$Corresponding author. E-mail: samarh@seas.upenn.edu

\end{center}


\subsubsection*{This PDF file includes:}
Supplementary Texts S1 to S4\\
Figures S1 to S26\\
Table S1\\


\newpage
    
\appendix

%

\subsection*{Topic Hierarchy}
\label{sec:topics}

\begin{figure}[h]
    \centering
    \includegraphics[width=\textwidth, trim=0 100 0 100, clip]{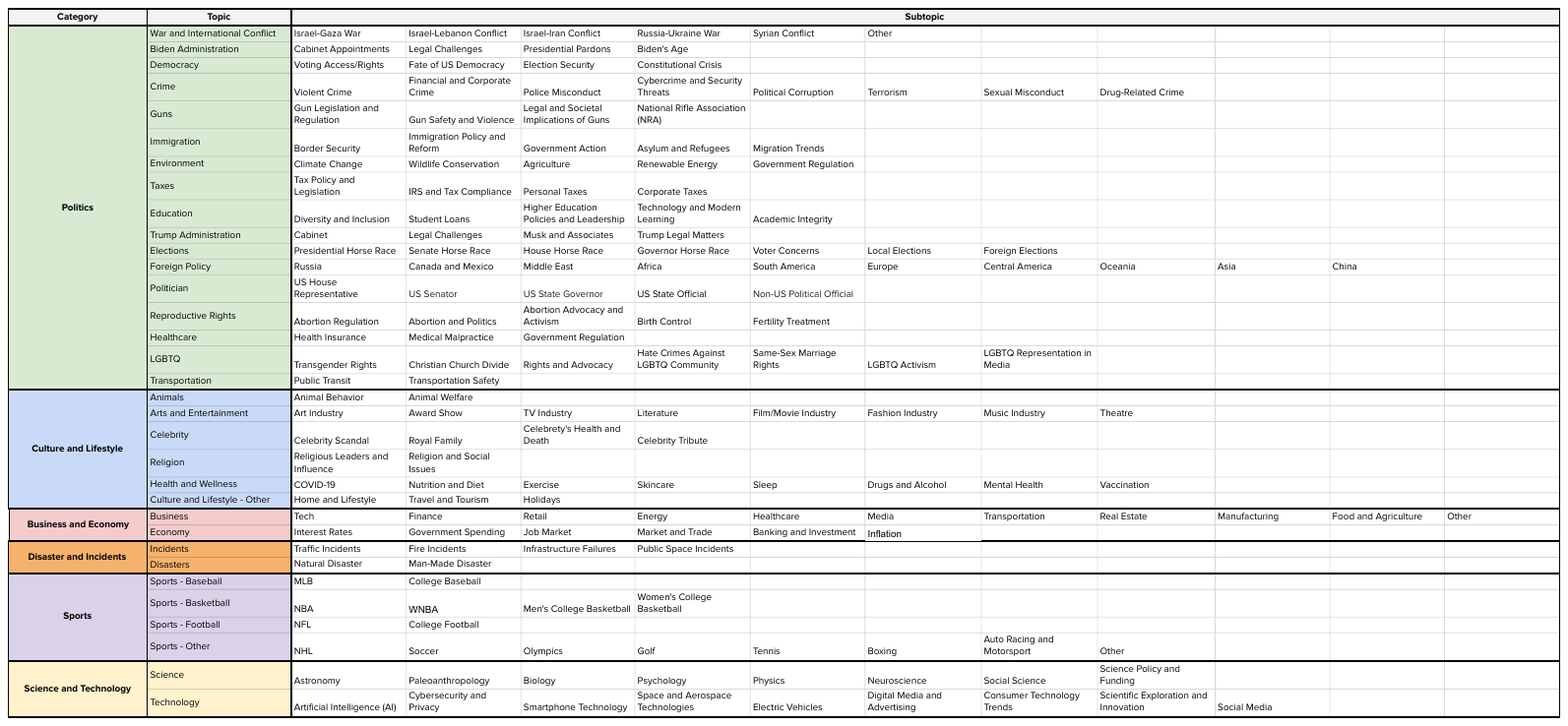}
    \caption{Our full topic hierarchy consisting of 6 broad news categories composed of 33 topics and 162 subtopics.}
    \label{fig:topic_hierarchy}
\end{figure}

\clearpage

\subsection*{Prompts}

\begin{figure*}[h]
    \centering
    \begin{tcolorbox}[enhanced, width=\linewidth, boxrule=0.8mm, colback=gray!5, colframe=gray!60, 
    title=Article Topic, fonttitle=\bfseries\ttfamily\small, coltitle=black]

    \small \tt
    The following is a news article. Read it and perform the task that follows. Respond with a JSON object of key-value pairs. \\

    \#\#\#\#\#\#\#\#\#\#\#\#\#\#\#\#\#\#\#\# \\

    \red{\{article\}} \\

    \#\#\#\#\#\#\#\#\#\#\#\#\#\#\#\#\#\#\#\# \\

    Task: Classify the article into one of the listed topics. \\

    Instruction: Try your best to bucket the article into one of these topics. DO NOT write anything that is not listed. \\

    Key: "topic"\\
    Value: One of: \red{\{topic\_list\}}. \\

    Do not return anything except the JSON object of key-value pairs as output.

    \end{tcolorbox}
    \caption{\textbf{Article topic labeling prompt.} The LLM is provided with a list of the 36 topics in our list and asked to choose the most appropriate on from them. Model: GPT-4o-mini.}
    \label{fig:topic_prompt}
\end{figure*}

\begin{figure*}[h]
    \centering
    \begin{tcolorbox}[enhanced, width=\linewidth, boxrule=0.8mm, colback=gray!5, colframe=gray!60, 
    title=Article Subtopic, fonttitle=\bfseries\ttfamily\small, coltitle=black]

    \small \tt
The following is a news article on the topic of \red{\{predicted\_topic\}}. Read it and perform the task that follows. Respond with a JSON object of key-value pairs. \\

\#\#\#\#\#\#\#\#\#\#\#\#\#\#\#\#\#\#\#\# \\

\red{\{article\}} \\

\#\#\#\#\#\#\#\#\#\#\#\#\#\#\#\#\#\#\#\# \\

Task: Classify the article into one of the listed subtopics under the predicted topic. \\

Instruction: Try your best to bucket the article into one of these subtopics. Label it as 'Other' if the article does not fit any possible subtopics. \\

Key: "subtopic" 

Value: One of: \red{\{subtopic\_list\_under\_the\_topic\}}. \\

Do not return anything except the JSON object of key-value pairs as output.

    \end{tcolorbox}
    \caption{\textbf{Article subtopic labeling prompt.} Once the article topic has been labeled, we ask the LLM to pick the most appropriate subtopic from those that fall under that topic in our hierarchy. Model: GPT-4o-mini.}
    \label{fig:subtopic_prompt}
\end{figure*}

\begin{figure*}[h]
    \centering
    \begin{tcolorbox}[enhanced, width=\linewidth, boxrule=0.8mm, colback=gray!5, colframe=gray!60, 
    title=Article Takeaways, fonttitle=\bfseries\ttfamily\small, coltitle=black]

 \tt
The following is a news article. Read it and perform the task that follows. Respond with a JSON object of key-value pairs. \\

\#\#\#\#\#\#\#\#\#\#\#\#\#\#\#\#\#\#\#\# \\

\red{\{article\}} \\

\#\#\#\#\#\#\#\#\#\#\#\#\#\#\#\#\#\#\#\# \\

Task: Summarize the main points of the news article. \\

Instruction: List short takeaway points that readers are likely to remember from the article. \\
Key: "takeaways" \\
Value: A 3-4 sentence summary. \\
    \end{tcolorbox}
    \caption{\textbf{Article takeaways prompt.} To summarize the article for use in later prompts and to aid manual review of the topic and subtopic labels, we use an LLM to generate a short summary of the main points of the article. Model: GPT-4o-mini.}
    \label{fig:prompt21}
\end{figure*}

\begin{figure*}[h]
    \centering
    \begin{tcolorbox}[enhanced, width=\linewidth, boxrule=0.8mm, colback=gray!5, colframe=gray!60, 
    title=Article Type, fonttitle=\bfseries\ttfamily\small, coltitle=black]

    \footnotesize \tt
The following is a news article. Read it and perform the task that follows. Respond with a JSON object of key-value pairs. \\

\#\#\#\#\#\#\#\#\#\#\#\#\#\#\#\#\#\#\#\# \\

\red{\{article\}} \\

\#\#\#\#\#\#\#\#\#\#\#\#\#\#\#\#\#\#\#\# \\

Task: Determine the news type of this news article. \\

1. Instruction: Classify the above news article into one of three categories: news report, news analysis, or opinion. Each category has distinct characteristics: \\
- News Report: Objective reporting on recent events, focusing on verified facts without the writer's personal views. \\
- News Analysis: In-depth examination and interpretation of news events, providing context and explaining significance, while maintaining a degree of objectivity. \\
- Opinion: Articles reflecting personal views, arguments, or beliefs on current issues, often persuasive and subjective. \\
Consider criteria such as language objectivity, focus on facts versus interpretation, author's intent, and article structure. 

Key: "news\_type" \\
Value: One of "news report" or "news analysis" or "opinion". \\

2. Instruction: Provide a short paragraph to justify your classification, citing specific elements from the text. \\
Key: "justification" \\
Value: A paragraph of text. \\

Do not return anything except the JSON object of key-value pairs as output.

    \end{tcolorbox}
    \caption{\textbf{Article type labeling prompt.} To classify the type of the article, we give the LLM descriptions of the three primary news types (report, analysis, opinion) and ask it to choose the most appropriate one and provide a justification for its choice. Model: GPT-4o.}
    \label{fig:article_type_prompt}
\end{figure*}

\begin{figure*}[h]
    \centering
        \begin{tcolorbox}[enhanced, width=\linewidth, boxrule=0.8mm, colback=gray!5, colframe=gray!60, 
        title=Article Lean, fonttitle=\bfseries\ttfamily\small, coltitle=black]

        \small \tt
The following is a news article. Read it and perform the task that follows. Respond with a JSON object of key-value pairs. 

\#\#\#\#\#\#\#\#\#\#\#\#\#\#\#\#\#\#\#\# \\

\red{\{article\}} \\

\#\#\#\#\#\#\#\#\#\#\#\#\#\#\#\#\#\#\#\# \\

Task: Determine the political lean of this article within the U.S. political context. Is it supporting the Democrat party or the Republican party? Supporting a party can mean supporting its viewpoints, politicians, or policies. Provide reasoning for your answer. \\

1. Instruction: Give a short paragraph summarizing in what ways the article supports the Democrat party or the Republican party. \\
Key: "reason" \\
Value: A paragraph of text. \\

2. Instruction: Give a number from -5 to 5, with -5 indicating strong support for Democrats and 5 indicating strong support for Republicans. A value of 0 indicates that the article has no clear political lean towards either side. \\
Key: "lean" \\
Value: An integer number from -5 to 5. \\

Do not return anything except the JSON object of key-value pairs as output.

        \end{tcolorbox}
        \caption{\textbf{Article lean labeling prompt.} For classifying the political lean of an article, we use the same approach as that for classifying tone: first asking for an analysis of the article's political lean in the context of the U.S. political landscape and then asking for an appropriate numerical score. However, since political lean is more nuanced than tone, we provide additional instructions which define support for a political party to mean support for its viewpoints, politicians, or policies. Model: GPT-4o.}
        \label{fig:article_lean_prompt}

\end{figure*}

\begin{figure*}
        \centering
        \begin{tcolorbox}[enhanced, width=\linewidth, boxrule=0.8mm, colback=gray!5, colframe=gray!60, 
        title=Article Tone, fonttitle=\bfseries\ttfamily\small, coltitle=black]

        \small \tt
The following is a news article. Read it and perform the task that follows. Respond with a JSON object of key-value pairs. 

\#\#\#\#\#\#\#\#\#\#\#\#\#\#\#\#\#\#\#\# \\

\red{\{article\}} \\

\#\#\#\#\#\#\#\#\#\#\#\#\#\#\#\#\#\#\#\# \\

Task: Determine the overall tone of the article. Is it negative, positive, or neutral? \\

1. Instruction: Provide a short paragraph summarizing in what ways the article has a negative or positive tone. \\
Key: "reason" \\
Value: A paragraph of text. \\

2. Instruction: Provide a number from -5 to 5, with -5 indicating a very negative tone and 5 indicating a very positive tone. A value of 0 indicates that the article has a neutral tone. \\
Key: "tone" \\
Value: An integer number from -5 to 5. \\

Do not return anything except the JSON object of key-value pairs as output.

\end{tcolorbox}
\caption{\textbf{Article tone labeling prompt.} To classify an article's tone, we instruct the LLM to first analyze the article and provide a summary of the main features which give it a certain tone. We then instruct the LLM to assign the article a numerical score on an 11-point Likert scale that accurately reflects the above analysis. Model: GPT-4o.}
\label{fig:article_tone_prompt}
\end{figure*}

\begin{figure*}[h]
    \centering
        \begin{tcolorbox}[enhanced, width=\linewidth, boxrule=0.8mm, colback=gray!5, colframe=gray!60, 
        title=Headline Lean, fonttitle=\bfseries\ttfamily\small, coltitle=black]

        \small \tt
The following is the title of a news article on the topic of \red{\{topic\} (\{subtopic\})} in the \{category\} news category. Read it and perform the task that follows. Respond with a JSON object of key-value pairs.

\#\#\#\#\#\#\#\#\#\#\#\#\#\#\#\#\#\#\#\# \\

\red{\{article\}} \\

\#\#\#\#\#\#\#\#\#\#\#\#\#\#\#\#\#\#\#\# \\

Task: Determine the political lean of this article title within the U.S. political context. Is it supporting the Democrat party or the Republican party? Supporting a party can mean supporting its viewpoints, politicians, or policies. Provide reasoning for your answer. \\

1. Instruction: Give a short paragraph summarizing in what ways the article title supports the Democrat party or the Republican party. \\
Key: "reason" \\
Value: A paragraph of text. \\

2. Instruction: Give a number from -5 to 5, with -5 indicating strong support for Democrats and 5 indicating strong support for Republicans. A value of 0 indicates that the article title has no clear political lean towards either side. \\
Key: "lean" \\
Value: An integer number from -5 to 5. \\

Do not return anything except the JSON object of key-value pairs as output.

        \end{tcolorbox}
        \caption{\textbf{Headline lean labeling prompt.} Similarly to tone, we use the same prompt to label headline lean as we do for the article text, except that we explicitly mention the topic of the article to give the model context that the headline may lack. Model: GPT-4o.}
        \label{fig:headline_lean_prompt}

\end{figure*}

\begin{figure*}
        \centering
        \begin{tcolorbox}[enhanced, width=\linewidth, boxrule=0.8mm, colback=gray!5, colframe=gray!60, 
        title=Headline Tone, fonttitle=\bfseries\ttfamily\small, coltitle=black]

        \small \tt
The following is the title of a news article on the topic of \red{\{topic\} (\{subtopic\})} in the \{category\} news category. Read it and perform the task that follows. Respond with a JSON object of key-value pairs.

\#\#\#\#\#\#\#\#\#\#\#\#\#\#\#\#\#\#\#\# \\

\red{\{article\}} \\

\#\#\#\#\#\#\#\#\#\#\#\#\#\#\#\#\#\#\#\# \\

Task: Determine the overall tone of the article title. Is it negative, positive, or neutral? \\

1. Instruction: Provide a short paragraph summarizing in what ways the article title has a negative or positive tone. \\
Key: "reason" \\
Value: A paragraph of text. \\

2. Instruction: Provide a number from -5 to 5, with -5 indicating a very negative tone and 5 indicating a very positive tone. A value of 0 indicates that the article has a neutral tone. \\
Key: "tone" \\
Value: An integer number from -5 to 5. \\

Do not return anything except the JSON object of key-value pairs as output.

\end{tcolorbox}
\caption{\textbf{Article headline tone labeling prompt.} For consistency, we use the same prompt to label headline tone as we do for the article text, except that we explicitly mention the topic of the article to give the model context that the headline may lack. Model: GPT-4o.}
\label{fig:headline_tone_prompt}
\end{figure*}

\begin{figure*}[h]
    \centering
    \begin{tcolorbox}[enhanced, width=\linewidth, boxrule=0.8mm, colback=gray!5, colframe=gray!60, 
    title=Sentence Labeling, fonttitle=\bfseries\ttfamily\small, coltitle=black]
 
    \tiny \tt
The following is a numbered list of sentences from a news article. Read them all and perform the task that follows: \\

\#\#\#\#\#\#\#\#\#\#\#\#\#\#\#\#\#\#\#\# \\

\red{\{sentences\}} \\

\#\#\#\#\#\#\#\#\#\#\#\#\#\#\#\#\#\#\#\# \\

Task: For each sentence above, perform the following analysis. Return the output as a list of JSON objects of key-value pairs, one for each sentence, where the keys are: \\

1. Instruction: Output the number of the current sentence being analyzed. \\
Key: "sentence" \\
Value: The sentence number. \\

2. Instruction: Determine whether the sentence is a "fact", "opinion", "borderline", "quote", or "other" type of sentence. A "fact" is something that's capable of being proved or disproved by objective evidence. An "opinion" reflects the beliefs and values of whoever expressed it, but should not be a quote. A "borderline" sentence is one that is not entirely a fact or opinion. A "quote" is a passage from another person or source that comes with quotation marks. If a quote is composed of multiple sentences, they should all be labeled as "quote". \\
Key: "type" \\
Value: One of "fact", "opinion", "borderline", "quote", "other". Select "other" if the sentence does not fit the possible values. \\

3. Instruction: Classify the tone of the sentence as "positive", "negative", or "neutral". \\
Key: "tone" \\
Value: One of "positive", "negative", "neutral". \\

4. Instruction: Classify the focus of the sentence as being "democrat", "republican", or "neither" within the U.S. political context. If the sentence is referring to the Democratic Party, including its policies or politicians, label it as "democrat". If it is referring to the Republican Party, including its policies or politicians, label it as "republican". If it is referring to both, label it as "both". Otherwise, label it as "neither". \\
Key: "focus" \\
Value: One of "democrat", "republican", "both", "neither". \\

Do not return anything except the list of JSON objects of key-value pairs as output. Do not skip analyzing any sentences. Do not return any key values except those specified.

    \end{tcolorbox}
    \caption{\textbf{Sentence labeling prompt.} To label every sentence in an article, we provide the entire article in the prompt since classifying a sentence requires knowing its neighboring context. We format the article as a list of enumerated sentences, one per line, and ask the LLM to classify the tone, type, and focus of each one. While tone is self-explanatory, we provide definitions for type and focus to serve as guidelines. Model: GPT-4o.}
    \label{fig:sentence_prompt}
\end{figure*}

\begin{figure*}[h]
    \centering
    \begin{tcolorbox}[enhanced, width=\linewidth, boxrule=0.8mm, colback=gray!5, colframe=gray!60, 
    title=Quote Extraction, fonttitle=\bfseries\ttfamily\small, coltitle=black]
 
    \tiny \tt

The following is a news article. Read it and perform the task that follows. Respond with a list of JSON objects of key-value pairs, one for each quote, where the keys are as follows: \\

\#\#\#\#\#\#\#\#\#\#\#\#\#\#\#\#\#\#\#\# \\

\red{\{article\}} \\

\#\#\#\#\#\#\#\#\#\#\#\#\#\#\#\#\#\#\#\# \\

Task: Extract all direct quotes from the article above and provide a list of JSON objects of key-value pairs, one for each quote, where the keys are: \\

1. Instruction: Extract the full text of the quote itself. If a quote is split over multiple fragments, combine them into one. \\
Key: "quote" \\
Value: The full quote text. \\

2. Instruction: Extract the full name of the person being quoted. \\
Key: "person\_name" \\
Value: The person's full name. Leave blank if unknown. \\

3. Instruction: Extract the full occupation of the person being quoted. \\
Key: "person\_occupation" \\
Value: The person's full occupation. Leave blank if unknown. \\

4. Instruction: Extract the affiliation of the person being quoted. \\
Key: "person\_affiliation" \\
Value: The entity or organization that the person is affiliated with. Leave blank if unknown. \\

5. Instruction: Classify the domain of the person being quoted based on their occupation and affiliation. \\
Key: "person\_domain" \\
Value: One of "Politics", "Corporate", "Academia", "Legal", "Media and Arts", "Science and Technology", "Sports", "Law enforcement", "Military", "Health and Medicine", "Religion", "Community and Social Services", "Other". Select "Other" if the domain does not fit the possible values. \\

6. Instruction: Classify the capacity in which the person is being quoted based on the context of the article. An "illustrative anecdote" is when a person is being quoted as an example or illustration of a larger trend, a "commentary" is when a person is providing their thoughts on a topic or event, an "expert" is when a person is providing specialized knowledge or analysis, a "subject" is when a person is the main focus of the article, an "observer" is when a person is providing a first-hand account of an event, a "participant" is when the person being quoted was involved in an event, a "spokesperson" is when a person is speaking on behalf of an organization or entity. \\
Key: "person\_capacity" \\
Value: One of "expert", "observer", "participant", "subject", "commentary", "illustrative anecdote", "spokesperson", "other". Select "other" if the capacity does not fit the possible values. \\

Do not return anything except the list of JSON objects of key-value pairs as output. Do not skip any quotes.
    \end{tcolorbox}
    \caption{\textbf{Quote extraction prompt.} In addition to extracting the text of the quote and the name of the person being quoted, we instruct the LLM to extract additional information like the person's occupation and affiliation as it is reported in the article. In addition to this, we also have the model classify the person's domain based on their occupation and affiliation, and their capacity based on the context in which they are being quoted in the article. Model: GPT-4o.}
    \label{fig:quote_prompt}
\end{figure*}

\begin{figure*}[h]
    \centering
    \begin{tcolorbox}[enhanced, width=\linewidth, boxrule=0.8mm, colback=gray!5, colframe=gray!60, 
    title=Event Title, fonttitle=\bfseries\ttfamily\small, coltitle=black]

    \small \tt
The following is a list of titles of a collection of news articles around a particular event. Read them and perform the task that follows. Respond with a JSON object of key-value pairs. \\

\#\#\#\#\#\#\#\#\#\#\#\#\#\#\#\#\#\#\#\# \\

\red{\{article\_titles\}} \\

\#\#\#\#\#\#\#\#\#\#\#\#\#\#\#\#\#\#\#\# \\

Task: Summarize the titles. \\

1. Instruction: Analyze the titles and produce a good thematic title that best summarizes the event. \\
Key: "theme" \\
Value: The thematic title. \\

2. Instruction: Analyze the titles and produce a shorter title that is not longer than 4 to 5 words. \\
Key: "theme\_short" \\
Value: The shorter title. \\

Do not return anything except the JSON object of key-value pairs as output.

    \end{tcolorbox}
    \caption{\textbf{Event title prompt.} To generate descriptive titles for each event, we provide the model a list of headlines of articles from a cluster and ask it to generate a suitable theme for the event based on them. We generate these themes at two levels: a short one for the high-level, multi-event view in the dashboard and a longer one for the detailed, single-event view. Model: GPT-4o-mini.}
    \label{fig:event_title_prompt}
\end{figure*}

\begin{figure*}[h]
    \centering
    \begin{tcolorbox}[enhanced, width=\linewidth, boxrule=0.8mm, colback=gray!5, colframe=gray!60, 
    title=Cluster Recall, fonttitle=\bfseries\ttfamily\small, coltitle=black]

    \tt
Your task is to assign a news article to one of the following event themes based on its headline and summary. If the article falls under any of these themes, return the corresponding number. If the article does not fall under any of these themes, return -1. Only return an integer value and nothing else.\\

Headline: \red{\{title\}}\\

Summary: \red{\{takeaways\}}\\

Themes:
\red{\{themes\}}

    \end{tcolorbox}
    \caption{\textbf{Cluster recall prompt.} To  improve  recall and expand event clusters to include any articles that may be related, we show the LLM every headline from that day and ask it to classify it into one of the events of that day if it fits. Model: GPT-4o.}
    \label{fig:cluster_recall_prompt}
\end{figure*}

\begin{figure*}[h]
    \centering
    \begin{tcolorbox}[enhanced, width=\linewidth, boxrule=0.8mm, colback=gray!5, colframe=gray!60, 
    title=Cluster Precision, fonttitle=\bfseries\ttfamily\small, coltitle=black]

    \tt
Your task is to go through a list of news headlines associated with an event theme and return the corresponding number of any that are unrelated to that theme. If multiple headlines are unrelated to the theme, return their respective numbers separated by commas. If all headlines are relevant to the theme, return -1. Only return integer values and nothing else. \\

Theme: \red{\{theme\}} \\

Headlines:
\red{\{titles\}}

    \end{tcolorbox}
    \caption{\textbf{Cluster precision prompt.} To improve precision and prune event clusters to exclude any articles that are unrelated to the theme, we give the LLM a list of headlines from that cluster and ask it to flag any that should be removed. Model: GPT-4o.}
    \label{fig:cluster_precision_prompt}
\end{figure*}

\begin{figure*}[h]
    \centering
    \begin{tcolorbox}[enhanced, width=\linewidth, boxrule=0.8mm, colback=gray!5, colframe=gray!60, 
    title=Fact Summarization, fonttitle=\bfseries\ttfamily\small, coltitle=black]

    \small \tt
The following is a list of sentences about the same event. Read them and perform the task that follows. Respond with a JSON object of key-value pairs. \\

\#\#\#\#\#\#\#\#\#\#\#\#\#\#\#\#\#\#\#\# \\

\red{\{sentence\_list\}} \\

\#\#\#\#\#\#\#\#\#\#\#\#\#\#\#\#\#\#\#\# \\

Task: Analyze them and produce a comprehensive sentence that summarizes all of them and is no longer than 25 words. \\
Key: "synthetic\_sentence" \\
Value: The comprehensive sentence. \\

Do not return anything except the JSON object of key-value pairs as output.

    \end{tcolorbox}
    \caption{\textbf{Fact summarization prompt.} To generate representative sentences that capture maximum information about a given event, we use the LLM to summarize all the similar facts in a cluster into one overarching statement that covers all of them. Model: GPT-4o-mini.}
    \label{fig:summary_fact_prompt}
\end{figure*}

\clearpage

\subsection*{Data Validation}

\begin{figure}[h]
    \centering
    \includegraphics[width=\textwidth]{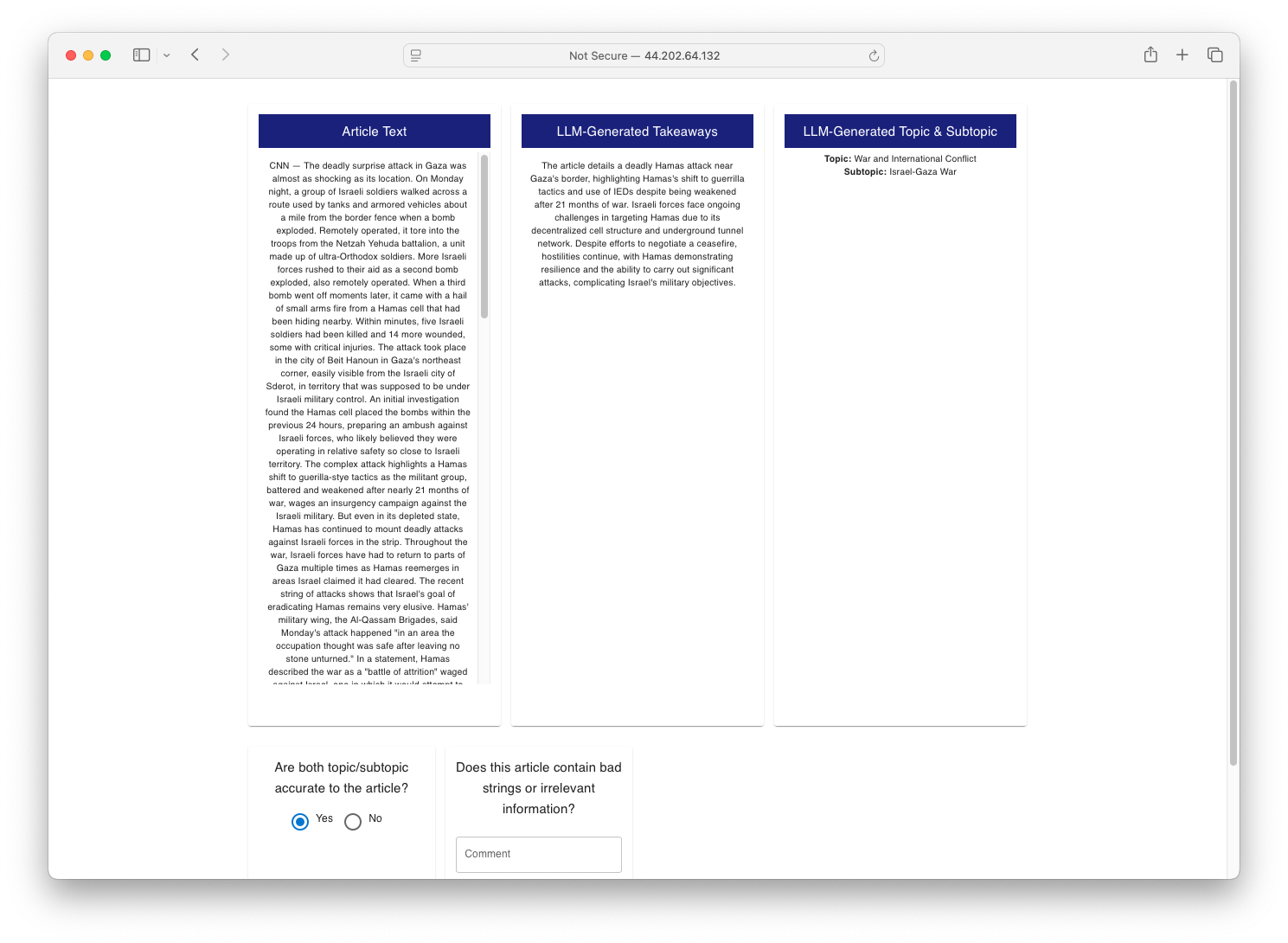}
    \caption{\textbf{Article topic validation interface.} Human annotators are shown the article text (left) along with the topic and subtopic labels assigned by the LLM (right). They are also provided a summary generated by the LLM to speed up the validation process (center) and asked to judge whether the assigned topic and subtopic are appropriate, and, if not, to suggest a better one, either from within our existing list or a new one.}
    \label{fig:topic_val}
\end{figure}

\clearpage

\begin{figure}[ht]
    \centering
    \includegraphics[width=\textwidth]{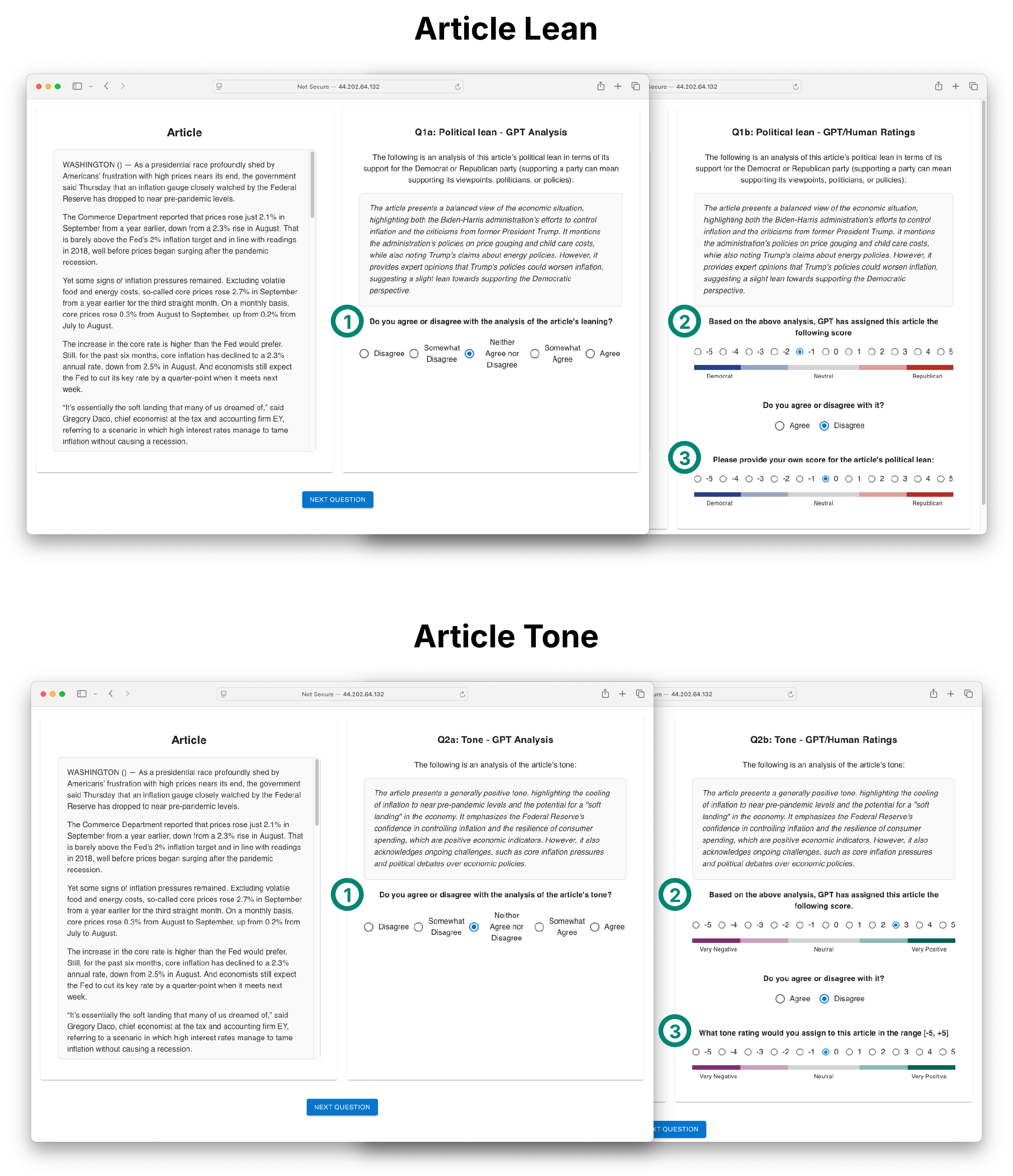}
    \caption{\textbf{Article lean and tone validation interface.} The annotators are shown the LLM's analysis of the article's lean and asked how strongly they agree or disagree with it \ding{172}. They are then shown the numerical rating assigned by LLM and asked to either confirm that it is appropriate \ding{173}, or choose a more suitable label themselves \ding{174}. The same process is then repeated for article tone.}
    \label{fig:lean_tone_val}
\end{figure}

\begin{figure}[ht]
    \centering
    \includegraphics[width=\textwidth, trim=0 0 0 50, clip]{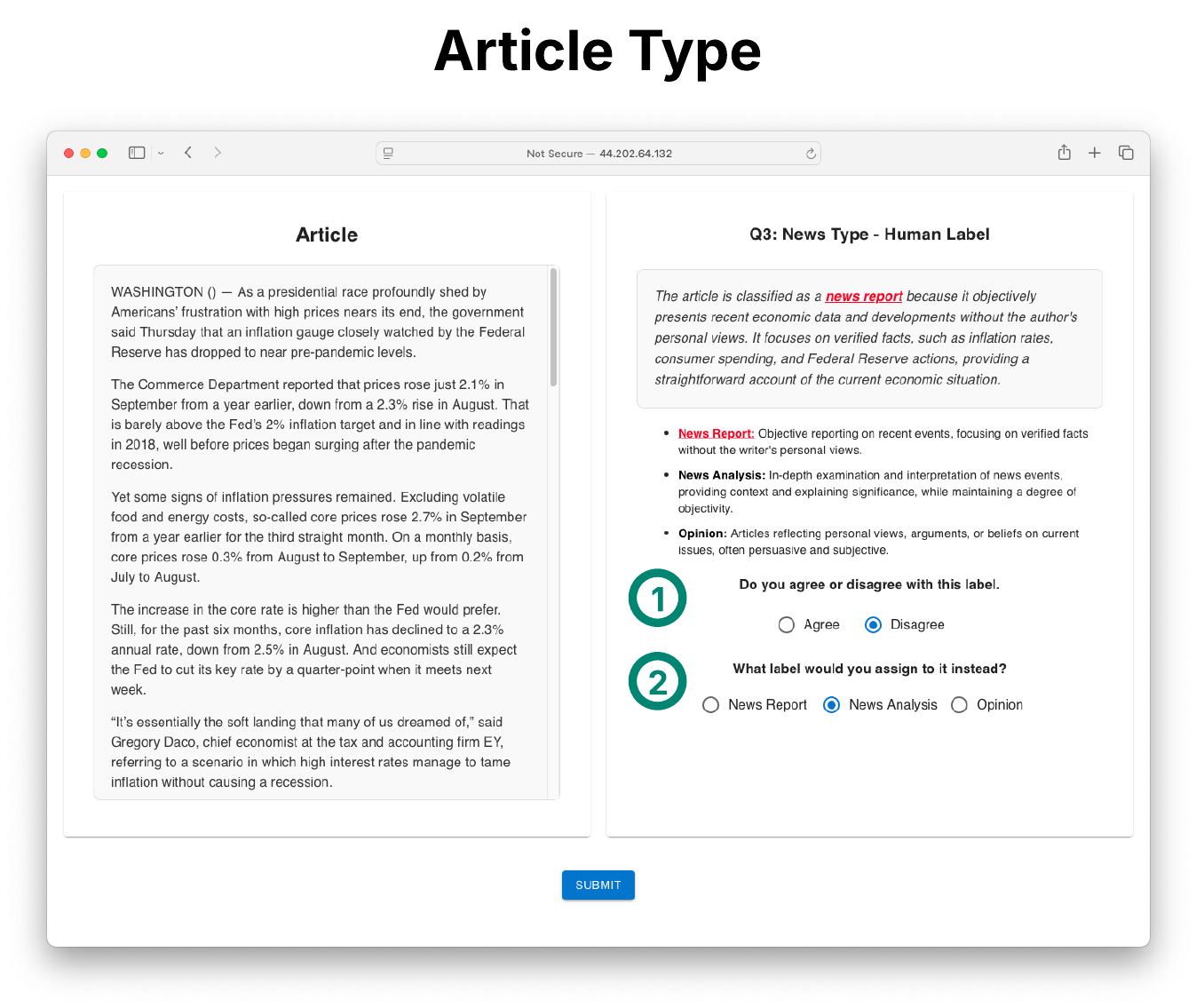}
    \caption{\textbf{Article type validation interface.} Like for tone and lean validation, the annotators are shown the LLM's analysis of the article's type and asked whether they agree or disagree with it \ding{172}. They are then shown the label assigned by LLM and asked to either confirm that it is appropriate, or choose a more suitable label themselves \ding{173}.}
    \label{fig:type_val}
\end{figure}

\begin{figure}[h!]
    \centering
    \includegraphics[width=\textwidth]{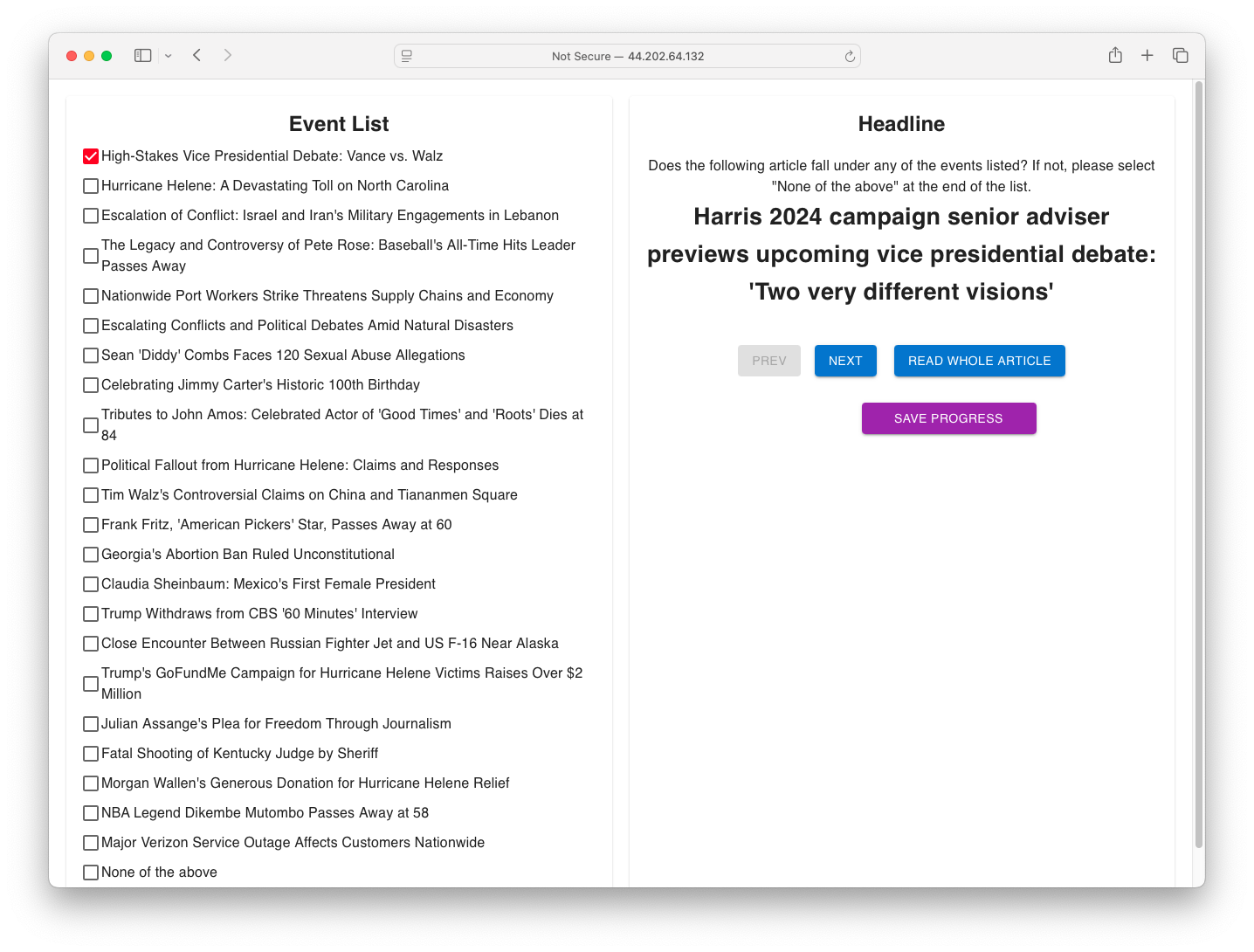}
    \caption{\textbf{Event cluster validation interface.} To validate the precision and recall of our event clustering approach, we ask the human annotator to review the headline of an article (right) and check whether it has been assigned to the appropriate event theme from that day (left). If the assignment is incorrect, they can select a more suitable event theme (or mark the article as not belonging to any of them). We repeat this review process for every single article in a day in our validation dataset.}
    \label{fig:event_val}
\end{figure}

\begin{figure}[h!]
    \centering
    \includegraphics[width=\textwidth]{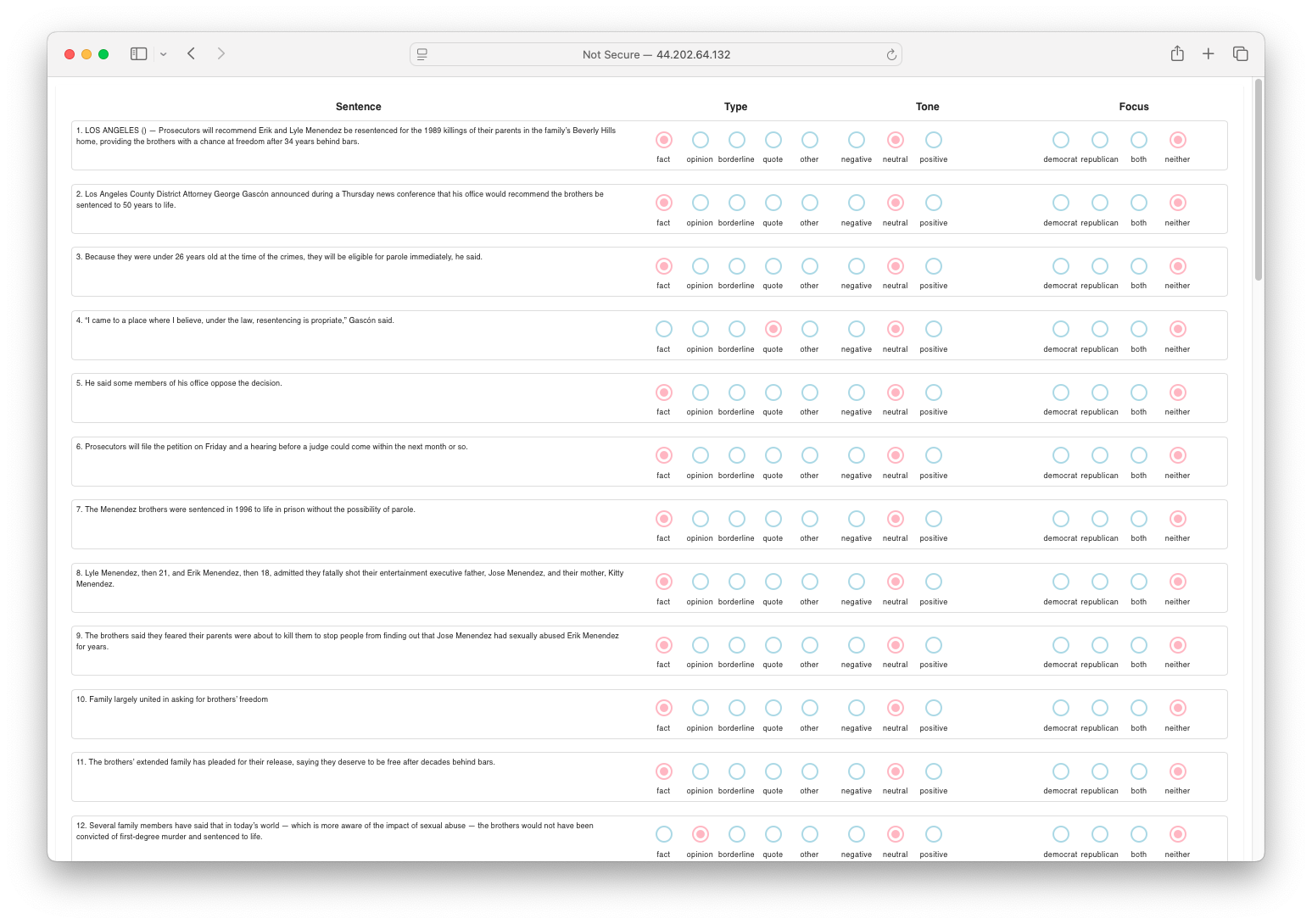}
    \caption{\textbf{Sentence label validation interface.} For validating sentence type, tone, and focus, the human annotator reads one article at a time and reviews the labels assigned by the LLM for each sentence. If they disagree, they can choose a more appropriate option.}
    \label{fig:sentence_val}
\end{figure}

\clearpage

\subsection*{Text S1: Sample Labeled Data}
\label{sec:sample_data}

The following are 10 examples of labeled articles generated by our system. You can view more examples in a spreadsheet at the following link: \url{https://tinyurl.com/media-bias-detector}

\begin{enumerate}
\item \textbf{URL}: \url{https://apnews.com/article/israel-lebanon-iran-hamas-hezbollah-haniyeh-262194a2d70273207ed1d1a5cb9ab09b} \\
\textbf{Category}: Politics \\
\textbf{Topic}: War and International Conflict \\
\textbf{Subtopic}: Israel-Lebanon Conflict \\
\textbf{Takeaways}: Recent Israeli strikes in Beirut and Tehran have targeted high-ranking figures in Hezbollah and Hamas, raising tensions and prompting discussions of retaliation. The assassination of a Hezbollah commander and Hamas leader Ismail Haniyeh has left both groups in a difficult position, as they must respond to restore deterrence without escalating the conflict further. Analysts suggest that Hezbollah may retaliate with significant military action, while Iran's response could involve its proxies or a direct strike on Israel, complicating the already volatile situation. \\
\textbf{News Type}: news analysis \\
\textbf{Justification}: The article provides an in-depth examination and interpretation of the recent strikes in Beirut and Tehran, discussing their implications for Hezbollah, Iran, and the broader regional conflict. It includes expert opinions and analysis to explain the significance and potential consequences of these events, which goes beyond mere reporting of facts. \\
\textbf{Article Lean}: 0 \\
\textbf{Reason Article Lean}: The article provides a detailed account of the strikes and their implications without showing clear support for either the Democrat or Republican party. It focuses on the geopolitical consequences rather than U.S. domestic politics. \\
\textbf{Article Tone}: -4 \\
\textbf{Reason Article Tone}: The article has a negative tone as it discusses violent strikes, assassinations, and the potential for further escalation and regional conflict. The focus is on the serious security breaches, the loss of lives, and the delicate balance required to avoid a wider war. \\
\textbf{Title Lean}: 0 \\
\textbf{Reason Title Lean}: The article title does not explicitly support either the Democrat or Republican party. It focuses on the strategic calculations of Iran and Hezbollah in response to Israeli strikes, which is a neutral observation rather than a partisan stance. \\
\textbf{Title Tone}: -3 \\
\textbf{Reason Title Tone}: The article title suggests a high-stakes situation where Iran and Hezbollah must carefully manage their responses to Israeli strikes, implying potential negative consequences if they make errors. This creates a sense of tension and urgency, contributing to a negative tone.

\item \textbf{URL}: \url{https://apnews.com/article/olympics-2024-yusuf-dikec-turkish-shooter-a7890124304080a48e7ee4294004d306} \\
\textbf{Category}: Sports \\
\textbf{Topic}: Sports - Other \\
\textbf{Subtopic}: Olympics \\
\textbf{Takeaways}: Turkish shooter Yusuf Dikec gained viral fame for his relaxed shooting style, winning a silver medal in mixed team 10-meter air pistol at the 2024 Olympics, marking Turkey's first medal in Olympic shooting. Dikec's casual attire, including a T-shirt and one hand in his pocket, contrasted sharply with his competitors' more traditional gear. He expressed hopes for a gold medal at the next Olympics in Los Angeles in 2028. Dikec embraced his newfound internet fame, sharing memes about himself on social media. \\
\textbf{News Type}: news report \\
\textbf{Justification}: The article objectively reports on the events surrounding Turkish shooter Yusuf Dikec's performance at the 2024 Olympics, focusing on verified facts such as his medal win, his viral status, and details about the competition. It does not provide in-depth analysis or personal opinions from the author. \\
\textbf{Article Lean}: 0 \\
\textbf{Reason Article Lean}: The article focuses on the achievements and viral moments of Olympic shooters without delving into any political viewpoints, politicians, or policies. It remains neutral and purely sports-centric. \\
\textbf{Article Tone}: 4 \\
\textbf{Reason Article Tone}: The article has a positive tone as it highlights the achievements of Yusuf Dikec and other shooters at the 2024 Olympics, celebrating their unique styles and historic wins. It also emphasizes the viral popularity and positive reception of these athletes on social media. \\
\textbf{Title Lean}: 0 \\
\textbf{Reason Title Lean}: The article title is focused on a sports figure and does not mention any political viewpoints, politicians, or policies. It is neutral in the context of U.S. political leanings. \\
\textbf{Title Tone}: 0 \\
\textbf{Reason Title Tone}: The article title has a neutral tone as it simply introduces Yusuf Dikec, a Turkish shooter, and mentions that he went viral at the 2024 Olympics without any positive or negative connotations.

\item \textbf{URL}: \url{https://www.breitbart.com/border/2024/08/01/new-york-supreme-court-rules-texas-can-continue-busing-migrants-to-big-apple} \\
\textbf{Category}: Politics \\
\textbf{Topic}: Immigration \\
\textbf{Subtopic}: Government Action \\
\textbf{Takeaways}: The New York Supreme Court has rejected Mayor Eric Adams' lawsuit to block Texas Governor Greg Abbott's migrant busing program, allowing Texas to continue sending migrants to New York City. Abbott has vowed to persist with the busing initiative until federal border security is improved. This ruling marks Abbott's second legal victory in two days, following a decision that allows Texas to keep buoys in the Rio Grande to secure the border. \\
\textbf{News Type}: news report \\
\textbf{Justification}: The article objectively reports on recent legal rulings involving Texas Governor Greg Abbott's migrant busing program and the installation of buoys in the Rio Grande. It focuses on verified facts, such as court decisions and statements from involved parties, without providing the writer's personal views or in-depth analysis. \\
\textbf{Article Lean}: 4 \\
\textbf{Reason Article Lean}: The article highlights legal victories for Texas Governor Greg Abbott, a Republican, and emphasizes his stance against the Biden-Harris Administration's border policies. It portrays Abbott's actions and statements in a positive light, suggesting support for Republican viewpoints and policies. \\
\textbf{Article Tone}: 4 \\
\textbf{Reason Article Tone}: The article has a positive tone towards Governor Greg Abbott, highlighting his legal victories in both the New York Supreme Court and the U.S. Court of Appeals for the Fifth Circuit. It emphasizes his success in continuing the migrant busing program and keeping the buoys in the Rio Grande, portraying these outcomes as wins against opposition. \\
\textbf{Title Lean}: 4 \\
\textbf{Reason Title Lean}: The article title suggests that Texas, a state with Republican leadership, can continue its policy of busing migrants to New York, a Democrat-led city. This policy is generally supported by Republicans as a way to address immigration issues and criticized by Democrats. \\
\textbf{Title Tone}: 0 \\
\textbf{Reason Title Tone}: The article title has a neutral tone as it simply states a factual outcome of a court ruling without any emotionally charged language or subjective commentary.

\item \textbf{URL}: \url{https://www.usatoday.com/story/news/nation/2024/08/01/paul-whelan-released-russia-prisoner-release-marine/74632493007/} \\
\textbf{Category}: Politics \\
\textbf{Topic}: Foreign Policy \\
\textbf{Subtopic}: Russia \\
\textbf{Takeaways}: Paul Whelan, a former Marine wrongfully detained in Russia on espionage charges, was released in a recent prisoner swap after being imprisoned since December 2018. Whelan, who holds multiple citizenships, was accused of espionage but maintains he was set up. His case gained attention after the release of basketball star Brittney Griner in December 2022, and both the U.S. government and his family have consistently advocated for his freedom. \\
\textbf{News Type}: news report \\
\textbf{Justification}: The article provides an objective account of Paul Whelan's background, his arrest, and the circumstances surrounding his detention and release. It focuses on verified facts and events without offering personal views or in-depth analysis, which is characteristic of a news report. \\
\textbf{Article Lean}: -1 \\
\textbf{Reason Article Lean}: The article provides a balanced view of both the Trump and Biden administrations' actions regarding Paul Whelan's detention. It includes criticisms of Trump for not addressing Whelan's case during his presidency and acknowledges Biden's efforts to secure Whelan's release, suggesting a slight lean towards supporting the Democrat party. \\
\textbf{Article Tone}: -2 \\
\textbf{Reason Article Tone}: The article has a predominantly negative tone as it details Paul Whelan's wrongful detention, his harsh imprisonment conditions, and the lack of progress in securing his release. However, it also includes some positive elements, such as the support from his family and the U.S. government's ongoing efforts to free him. \\
\textbf{Title Lean}: 0 \\
\textbf{Reason Title Lean}: The article title does not provide any clear indication of support for either the Democrat or Republican party. It is a neutral headline focused on providing information about an individual freed from Russia. \\
\textbf{Title Tone}: 4 \\
\textbf{Reason Title Tone}: The article title has a positive tone as it highlights the freeing of Paul Whelan, a Michigan man, from Russia. The use of the word 'freed' suggests a positive outcome for Whelan.

\item \textbf{URL}: \url{https://www.usatoday.com/story/news/politics/elections/2024/08/02/kamala-harris-democrat-presidential-candidate/74631136007/} \\
\textbf{Category}: Politics \\
\textbf{Topic}: Elections \\
\textbf{Subtopic}: Presidential Horse Race \\
\textbf{Takeaways}: Vice President Kamala Harris has officially secured the 2024 Democratic presidential nomination, making history as the first Black woman and first Asian American to do so. She received over 2,350 delegate votes and will accept the nomination next week. Harris is set to face Republican nominee Donald Trump in the upcoming election, amid rising support and significant fundraising. The political landscape is heating up as both candidates prepare for a competitive race. \\
\textbf{News Type}: news report \\
\textbf{Justification}: The article objectively reports on Kamala Harris securing the Democratic presidential nomination, focusing on verified facts such as the voting process, delegate counts, and reactions from political figures. It does not provide in-depth analysis or personal opinions, maintaining a neutral tone throughout. \\
\textbf{Article Lean}: -4 \\
\textbf{Reason Article Lean}: The article highlights Kamala Harris's historic nomination as the Democratic presidential candidate, emphasizing support from prominent Democratic figures and her record-breaking donations. It also criticizes Donald Trump and his allies for using racist and misogynistic rhetoric against Harris. \\
\textbf{Article Tone}: 1 \\
\textbf{Reason Article Tone}: The article has a mixed tone. It highlights the historic and positive aspects of Kamala Harris's nomination, but also addresses the negative and racist attacks she faces from her opponents. \\
\textbf{Title Lean}: -4 \\
\textbf{Reason Title Lean}: The article title highlights Kamala Harris's historic nomination as the first Black woman and Asian American presidential nominee, which is a significant achievement for the Democratic party. This positive framing supports the Democratic party by emphasizing diversity and breaking barriers. \\
\textbf{Title Tone}: 5 \\
\textbf{Reason Title Tone}: The article title has a positive tone as it highlights a historic achievement by Kamala Harris, emphasizing her groundbreaking role as the first Black woman and Asian American presidential nominee. This conveys a sense of progress and celebration.

\item \textbf{URL}: \url{https://www.breitbart.com/politics/2024/08/08/exclusive-karoline-leavitt-harris-walz-the-most-far-left-ticket-weve-ever-had} \\
\textbf{Category}: Politics \\
\textbf{Topic}: Elections \\
\textbf{Subtopic}: Presidential Horse Race \\
\textbf{Takeaways}: Vice President Kamala Harris's decision not to select Pennsylvania Gov. Josh Shiro as her running mate has sparked criticism from Trump campaign spokeswoman Karoline Leavitt, who claims it reflects the Democratic Party's shift towards radical anti-Israel sentiments. Leavitt described Shiro as an 'obvious' choice due to his pro-Israel stance, which she believes alienates key Democratic voters. She characterized the Harris-Walz ticket as the most far-left in American history, emphasizing the need to expose Walz's record. \\
\textbf{News Type}: opinion \\
\textbf{Justification}: The article reflects personal views and arguments, particularly from Trump campaign spokeswoman Karoline Leavitt, who criticizes the Democrat Party and Kamala Harris's choice of running mate. The language used is subjective and persuasive, indicating a clear bias. \\
\textbf{Article Lean}: 5 \\
\textbf{Reason Article Lean}: The article supports the Republican party by criticizing Vice President Kamala Harris and the Democrat party, labeling them as antisemitic and anti-Israel radicals. It also praises the Trump campaign's readiness to expose the Democrat vice presidential nominee's record. \\
\textbf{Article Tone}: -4 \\
\textbf{Reason Article Tone}: The article has a negative tone as it criticizes Vice President Kamala Harris's choice of running mate and portrays the Democrat Party as antisemitic and radical. It also highlights the opposition's readiness to expose negative aspects of the chosen candidate's record. \\
\textbf{Title Lean}: 4 \\
\textbf{Reason Title Lean}: The article title uses the phrase 'Most Far-Left Ticket We've Ever Had,' which is a critical characterization often used by Republicans to describe Democrats in a negative light. This suggests a Republican-leaning perspective. \\
\textbf{Title Tone}: -3 \\
\textbf{Reason Title Tone}: The article title has a negative tone as it describes the Harris-Walz ticket as the 'Most Far-Left Ticket We've Ever Had,' which implies a critical view of their political stance.

\item \textbf{URL}: \url{https://www.theguardian.com/commentisfree/article/2024/aug/08/uk-far-right-riots-racists-communities} \\
\textbf{Category}: Culture and Lifestyle \\
\textbf{Topic}: Culture and Lifestyle - Other \\
\textbf{Subtopic}: Home and Lifestyle \\
\textbf{Takeaways}: The tragic killing of three young girls in Southport has been overshadowed by a rise in far-right extremism and Islamophobia, causing fear and anxiety among minority communities. The author reflects on personal experiences with racism and the violent atmosphere that has emerged, while also highlighting the solidarity and unity shown by diverse communities in response to hate. The article calls for accountability and a change in public discourse to combat racism and protect the values of diversity and inclusion in society. \\
\textbf{News Type}: opinion \\
\textbf{Justification}: The article reflects personal views and emotions of the author, Remona Aly, on the issue of far-right extremism and its impact on society. It uses subjective language and personal anecdotes, aiming to persuade readers about the dangers of racism and the importance of solidarity. \\
\textbf{Article Lean}: -5 \\
\textbf{Reason Article Lean}: The article strongly condemns far-right extremism, racism, and Islamophobia, which aligns with the Democrat party's stance on promoting diversity, inclusion, and combating hate speech. It also calls for accountability and a change in public and political discourse to challenge divisive rhetoric, which is a common viewpoint among Democrats. \\
\textbf{Article Tone}: -2 \\
\textbf{Reason Article Tone}: The article has a predominantly negative tone as it discusses the tragic deaths of three girls and the subsequent rise in far-right extremism and Islamophobia. However, it also highlights positive aspects such as community solidarity and anti-racism protests. \\
\textbf{Title Lean}: -2 \\
\textbf{Reason Title Lean}: The article title expresses fear over riots but also highlights a show of solidarity that brings hope. This focus on community and solidarity aligns more closely with Democratic values of social cohesion and collective action. \\
\textbf{Title Tone}: 2 \\
\textbf{Reason Title Tone}: The article title starts with a negative tone by mentioning a week of riots and fear, but it shifts to a positive tone by highlighting a show of solidarity that has given the author hope.

\item \textbf{URL}: \url{https://www.washingtonpost.com/history/2024/08/08/richard-nixon-farewell-address-50-years} \\
\textbf{Category}: Politics \\
\textbf{Topic}: Politician \\
\textbf{Subtopic}: Non-US Political Official \\
\textbf{Takeaways}: Richard Nixon delivered an emotional farewell speech on August 9, 1974, just hours before resigning due to the Watergate scandal. He reflected on his life, shared personal anecdotes about his parents, and acknowledged his vulnerabilities. Nixon's speech was marked by humor and poignant moments, revealing a side of him that had rarely been seen in public. His daughter later noted the difficulty of the moment as he let down his guard, highlighting the complexity of his public persona. \\
\textbf{News Type}: news analysis \\
\textbf{Justification}: The article provides an in-depth examination of Richard Nixon's farewell speech, including context, historical references, and interpretations of his actions and emotions. It goes beyond merely reporting the event by analyzing its significance and impact, as well as including reflections from various individuals. \\
\textbf{Article Lean}: 1 \\
\textbf{Reason Article Lean}: The article provides a nuanced and somewhat sympathetic portrayal of Richard Nixon, a Republican president, highlighting his emotional farewell speech and human side. It does not overtly criticize or support any political party but offers a balanced view of Nixon's final moments in office. \\
\textbf{Article Tone}: 2 \\
\textbf{Reason Article Tone}: The article has a somewhat positive tone as it highlights Richard Nixon's emotional farewell speech, showing his vulnerable and human side. It also includes moments of humor and personal reflection, which evoke empathy and a sense of understanding from the audience. \\
\textbf{Title Lean}: 2 \\
\textbf{Reason Title Lean}: The article title suggests a more human and empathetic portrayal of Nixon, a Republican president. This could be seen as a positive depiction, which may support the Republican party by highlighting a favorable aspect of one of its prominent figures. \\
\textbf{Title Tone}: 3 \\
\textbf{Reason Title Tone}: The article title suggests a positive tone by highlighting that Nixon, in his final act as president, revealed a more human and relatable side. This implies a moment of redemption or a positive change in perception.

\item \textbf{URL}: \url{https://www.foxnews.com/sports/beach-volleyball-legend-kerri-walsh-jennings-felt-usa-patriotism-paris-something-special} \\
\textbf{Category}: Sports \\
\textbf{Topic}: Sports - Other \\
\textbf{Subtopic}: Olympics \\
\textbf{Takeaways}: Kerri Walsh Jennings, the most decorated beach volleyball Olympian, expresses deep pride in representing the United States at the Olympics. She emphasizes the importance of national pride and the emotional connection athletes feel when competing for their country. Walsh Jennings reflects on her experiences and the responsibility of being a role model for future athletes, especially with the upcoming 2028 Summer Games in Los Angeles. \\
\textbf{News Type}: news report \\
\textbf{Justification}: The article objectively reports on Kerri Walsh Jennings' reflections on her Olympic career and her feelings of patriotism, without delving into personal opinions or in-depth analysis. It focuses on verified facts and quotes from Walsh Jennings, maintaining an objective tone throughout. \\
\textbf{Article Lean}: 3 \\
\textbf{Reason Article Lean}: The article emphasizes patriotism, national pride, and the importance of representing the United States, which are themes often associated with Republican viewpoints. It also features Fox News Digital, a media outlet known for its conservative leanings. \\
\textbf{Article Tone}: 5 \\
\textbf{Reason Article Tone}: The article has a positive tone as it highlights Kerri Walsh Jennings' pride in representing the United States, her admiration for other athletes, and the sense of patriotism and honor associated with competing in the Olympics. \\
\textbf{Title Lean}: 1 \\
\textbf{Reason Title Lean}: The article title mentions USA patriotism, which is often associated with Republican values. However, it does not explicitly support any political party or viewpoint. \\
\textbf{Title Tone}: 4 \\
\textbf{Reason Title Tone}: The article title has a positive tone as it highlights Kerri Walsh Jennings, a beach volleyball legend, feeling a sense of USA patriotism in Paris, which she describes as 'something special.' This suggests a positive and uplifting experience.

\item \textbf{URL}: \url{https://www.usatoday.com/story/entertainment/celebrities/2024/08/12/vince-vaughn-children-hollywood-walk-of-fame/74775775007/} \\
\textbf{Category}: Culture and Lifestyle \\
\textbf{Topic}: Celebrity \\
\textbf{Subtopic}: Celebrity Tribute \\
\textbf{Takeaways}: Vince Vaughn was honored with a star on the Hollywood Walk of Fame, celebrating his achievements while emphasizing the importance of his family. He was joined by his wife Kyla Weber and their two children, Locklyn and Vernon, during the ceremony. Vaughn expressed gratitude for his family's support and affection, highlighting that they are the most important part of his life. He humorously acknowledged that while accolades are nice, his children hold the greatest value to him. \\
\textbf{News Type}: news report \\
\textbf{Justification}: The article objectively reports on Vince Vaughn receiving a star on the Hollywood Walk of Fame, focusing on verified facts such as the event details, Vaughn's family presence, and his speech. It does not provide in-depth analysis or personal opinions from the author. \\
\textbf{Article Lean}: 0 \\
\textbf{Reason Article Lean}: The article does not support either the Democrat or Republican party. It focuses on Vince Vaughn's personal life and his Hollywood Walk of Fame ceremony without mentioning any political viewpoints, politicians, or policies. \\
\textbf{Article Tone}: 5 \\
\textbf{Reason Article Tone}: The article has a positive tone as it highlights Vince Vaughn's achievements, his love for his family, and the joy of being honored with a star on the Hollywood Walk of Fame. The article also includes heartfelt quotes from Vaughn expressing gratitude and love for his wife and children. \\
\textbf{Title Lean}: 0 \\
\textbf{Reason Title Lean}: The article title does not provide any information that supports either the Democrat or Republican party. It simply mentions a celebrity's appearance with his children at a public event. \\
\textbf{Title Tone}: 4 \\
\textbf{Reason Title Tone}: The article title has a positive tone as it highlights a rare and special family moment involving Vince Vaughn and his children at a prestigious event, the Hollywood Walk of Fame ceremony. This suggests a celebratory and heartwarming occasion.

\end{enumerate}

\clearpage

\subsection*{Text S2: Article validation codebook}
\label{article_val_codebook}

\textbf{Introduction:} The purpose of this exercise is to evaluate the quality of GPT’s labels for news articles’ \textbf{lean}, \textbf{tone}, and \textbf{type}. We’d like to know whether its reasoning and the numerical scores it assigns to articles are reasonable and could serve as useful labels for analyzing news coverage.

You will be presented with a set of news articles along with LLM-generated labels for the following attributes:

\begin{itemize}
    \item \textbf{Tone:} Does the article have a \textit{positive}, \textit{negative}, or \textit{neutral} tone?
    
    \item \textbf{Political Lean:} Does the article support the \textit{Democrat} party or the \textit{Republican} party? Supporting a party can mean supporting its viewpoints, politicians, or policies.
    
    \item \textbf{News Type:} Is the article a \textit{news report}, \textit{news analysis}, or \textit{opinion} piece?
\end{itemize}

First, you will see an analysis of the article’s \textit{lean} and \textit{tone} by the LLM based on the above definitions. You will be asked to rate how reasonable you find this analysis.

Then, you will be shown a numerical score assigned to the article based on this analysis on a scale of $[-5, +5]$. You will be asked whether you agree or disagree with this score. If you disagree, you will be asked to suggest the score that you think is more appropriate for this article.

For \textbf{lean} and \textbf{tone}, we elicit numerical scores on an 11-point Likert scale in the range $[-5, +5]$ and then categorize them into 5 buckets each.

This results in $5 \times 5 = 25$ combinations. You will be asked to validate labels for a total of \textbf{50 articles} (2 from each bucket) published between August and October, 2024.

\textbf{Annotation Platform:} Annotation platform link.
\\
(You should enter your name in the same way in the submission ID field every time you log in to ensure your progress is saved as you go.)

You can also browse all articles in advance in this spreadsheet.

\textbf{Sample Articles:} To help you understand the range of articles in our dataset and calibrate yourself to GPT’s reasoning and numerical labels, here are some examples of news articles—one from each bucket of \textbf{lean} and \textbf{tone}:

\paragraph{1. \textit{Biden’s Student Loan Forgiveness Program Has Canceled Debt for Over One Million People} \\ \textnormal{(The New York Times)}}
\begin{itemize}
    \item \textbf{Lean:} Democrat (-4). \textit{The article highlights the Biden administration's achievements in student debt relief, emphasizing its success in fulfilling promises aligned with Democratic priorities.}
    \item \textbf{Tone:} Very Positive (+4). \textit{The tone is positive, underscoring accomplishment and progress by the administration.}
    \item \textbf{Type:} News Report. \textit{The article presents verified facts without opinionated language.}
\end{itemize}

\paragraph{2. \textit{A Utility Promised to Stop Burning Coal. Then Google and Meta Came to Town.}}
\begin{itemize}
    \item \textbf{Lean:} Neutral-Leaning-Democrat (-3). \textit{Focuses on environmental justice, clean energy, and critiques of corporate inaction—aligning with Democratic values.}
    \item \textbf{Tone:} Very Negative (-4). \textit{Highlights broken promises and public health risks.}
    \item \textbf{Type:} News Analysis. \textit{Goes beyond reporting with stakeholder perspectives and context.}
\end{itemize}

\paragraph{3. \textit{32 Things We Learned in NFL Week 6: NFC North Dominance Escalates}}
\begin{itemize}
    \item \textbf{Lean:} Neutral (0). \textit{Focuses exclusively on sports with no political content.}
    \item \textbf{Tone:} Neutral (0). \textit{Balanced reporting on both positive and negative sports events.}
    \item \textbf{Type:} News Report. \textit{Objective presentation of facts and statistics.}
\end{itemize}

\paragraph{4. \textit{An Imprecise Biography Catches Up With Tim Walz}} 
\begin{itemize}
    \item \textbf{Lean:} Neutral-Leaning-Republican (+2). \textit{Highlights misstatements by a Democratic politician more critically than similar Republican actions.}
    \item \textbf{Tone:} Negative (-3). \textit{Underscores the implications of misstatements and reputational damage.}
    \item \textbf{Type:} News Analysis. \textit{Provides interpretation and comparative political context.}
\end{itemize}

\paragraph{5. \textit{Gov. Kemp Praises Trump for Keeping 'National Focus' on Georgia in Aftermath of Hurricane Helene}}
\begin{itemize}
    \item \textbf{Lean:} Republican (+4). \textit{Praises Republican leaders and emphasizes successful collaboration.}
    \item \textbf{Tone:} Positive (+3). \textit{Highlights optimism, resilience, and mutual commendation.}
    \item \textbf{Type:} News Report. \textit{Includes factual information and direct quotes without interpretation.}
\end{itemize}

\clearpage

\subsection*{Text S3: How GPT identifies political lean}
\label{gpt_lean_summary}

To understand how GPT-4o identifies political lean, we asked it to summarize the key features from its own analyses of 50 randomly sampled articles in each of the five buckets: Pro-Democrat (rated -5 or -4), Neutral / Lean Democrat (-3, -2), Neutral (-1, 0, +1), Neutral / Lean Republican (+2, +3), Pro-Republican (+4, +5). Note that GPT was not told beforehand which category the articles belonged to or what they were originally rated as, however it was still able to extract the common themes found across articles in each bucket. Its verbatim response is provided below.

\subsubsection*{Neutral}
The articles in this group generally exhibit a neutral to slightly critical tone towards prominent Republican figures and policies, while not overtly supporting Democratic viewpoints or policies. Common aspects include:
\begin{itemize}
    \item \textbf{Neutral to Slightly Critical Tone Towards Republicans:} Focus on legal challenges, internal conflicts, and criticisms of Republican figures such as Donald Trump, George Santos, and Marjorie Taylor Greene, without explicit support or condemnation.
    \item \textbf{Balanced Reporting:} Presentation of both Democratic and Republican viewpoints on various issues.
    \item \textbf{Focus on Internal Conflicts:} Emphasis on GOP infighting regarding leadership, finances, and policies.
    \item \textbf{Humanitarian and Legal Issues:} Coverage of topics like international aid and legal proceedings in a factual manner, with no strong partisan language.
    \item \textbf{Critique of Government Inefficiency:} Criticism of outdated practices or inefficiency, without overt support for Republican alternatives.
    \item \textbf{Emphasis on Democratic Efforts:} Occasional highlighting of Democratic initiatives, such as Biden's humanitarian aid efforts.
    \item \textbf{Nuanced Perspectives:} Criticism of both parties, presenting complex, balanced narratives.
\end{itemize}

\subsubsection*{Neutral / Lean Democrat}
These articles are subtly aligned with Democratic values, characterized by:
\begin{itemize}
    \item \textbf{Critical of Republican Figures and Policies:} Emphasis on legal troubles and controversies surrounding GOP leaders.
    \item \textbf{Support for Democratic Policies and Figures:} Positive framing of Democratic actions, especially on social issues.
    \item \textbf{Emphasis on Humanitarian and Social Issues:} Focus on empathy, human rights, and international cooperation.
    \item \textbf{Criticism of Conservative Policies:} Highlighting the negative outcomes of Republican stances on immigration, criminal justice, and foreign policy.
    \item \textbf{Focus on Environmental and Climate Issues:} Coverage aligns with Democratic priorities on environmental protection.
    \item \textbf{Support for Multilateralism and Diplomacy:} Advocacy for international collaboration in foreign affairs.
    \item \textbf{Advocacy for Social and Economic Equality:} Discussion of workers' rights, education, and healthcare.
    \item \textbf{Critical of Authoritarian Regimes:} Emphasis on democracy and human rights globally.
\end{itemize}

\subsubsection*{Pro-Democrat}
These articles show a strong alignment with Democratic perspectives:
\begin{itemize}
    \item \textbf{Criticism of Republican Figures and Policies:} Consistent negative framing of Republican leaders and ideologies.
    \item \textbf{Support for Democratic Figures and Policies:} Praising Democratic accomplishments and leadership.
    \item \textbf{Alignment with Progressive Values:} Advocacy for LGBTQ+ rights, racial justice, and environmental reform.
    \item \textbf{Negative Portrayal of Conservative Policies:} Strong critique of Republican stances on abortion, gun rights, and economic policy.
    \item \textbf{Emphasis on Accountability and Ethics:} Highlighting legal or ethical shortcomings of GOP figures.
    \item \textbf{Use of Sarcasm and Satire:} Derisive tone toward Republican actions and rhetoric.
    \item \textbf{Support for Social Justice Movements:} Affirmative coverage of movements like Black Lives Matter and climate activism.
\end{itemize}

\subsubsection*{Neutral / Lean Republican}
These articles slightly lean Republican, with these common features:
\begin{itemize}
    \item \textbf{Criticism of Democratic Figures and Policies:} Focus on Biden, Hillary Clinton, and other Democrats' shortcomings.
    \item \textbf{Focus on Republican Candidates and Internal GOP Dynamics:} Coverage of GOP primaries and intra-party competition.
    \item \textbf{Support for Republican Policies and Narratives:} Favorable framing of issues like border security and military strength.
    \item \textbf{Sympathetic Portrayal of Republican Figures:} Positive or neutral coverage of GOP leaders like Nikki Haley and Trump.
    \item \textbf{Emphasis on Democratic Controversies:} Highlighting ethics concerns and conflicts involving Democrats.
    \item \textbf{Highlighting Republican Legislative Efforts:} Focus on GOP opposition and initiatives in Congress.
    \item \textbf{Coverage of Republican Base Issues:} Topics include gun rights, immigration, and criticism of "cancel culture."
\end{itemize}

\subsubsection*{Pro-Republican}
These articles consistently reflect Republican values and criticize Democratic policies:
\begin{itemize}
    \item \textbf{Positive Portrayal of Republican Figures:} Consistently favorable coverage of Donald Trump and other GOP leaders.
    \item \textbf{Criticism of Democratic Figures and Policies:} Frequent targeting of Biden, Harris, and other Democrats' actions.
    \item \textbf{Emphasis on Republican Talking Points:} Focus on national security, anti-wokeness, and free-market principles.
    \item \textbf{Highlighting Alleged Bias and Corruption:} Claims of politically motivated attacks on Republicans.
    \item \textbf{Support for Conservative Social Issues:} Opposition to progressive gender and reproductive rights movements.
    \item \textbf{Negative Framing of Left-Leaning Organizations:} Portrayal of liberal policies as harmful to conservative values.
    \item \textbf{Use of Strong Language and Rhetoric:} Emotive and loaded terminology used to frame political opposition.
\end{itemize}

\subsection*{Text S4: Sentence validation codebook}
\label{sentence_val_codebook}

We provided the human annotators the following codebook to assist with the sentence type, tone, and focus validation task:

\textbf{Introduction:} The purpose of this exercise is to evaluate the quality of GPT’s labels for sentences’ \textbf{news type}, \textbf{tone}, and \textbf{focus}. We’d like to know whether its labels are reasonable and could serve as a reliable basis for analyzing news articles.

You will be presented with a set of news articles along with LLM-generated labels for each sentence for the following attributes:

\begin{itemize}
    \item \textbf{Type:} Is the sentence a \textit{``fact,'' ``opinion,'' ``borderline,'' ``quote,''} or \textit{``other''} type of sentence?
    \begin{itemize}
        \item A \textbf{fact} is something capable of being proved or disproved by objective evidence.
        \item An \textbf{opinion} reflects the beliefs and values of whoever expressed it but should not be a quote.
        \item A \textbf{borderline} sentence is one that is not entirely a fact or opinion.
        \item A \textbf{quote} is a passage from another person or source that comes with quotation marks. If a quote is composed of multiple sentences, they should all be labeled as \textit{quote}.
    \end{itemize}

    \item \textbf{Tone:} Does the sentence have a \textit{``positive,'' ``negative,''} or \textit{``neutral''} tone?

    \item \textbf{Focus:} Does the sentence refer to \textit{Democrats}, \textit{Republicans}, \textit{both}, or \textit{neither}? Referring to a party includes referring to its policies or politicians.
\end{itemize}

If you disagree with the label for any sentence, you can select the option that you prefer.

\textbf{Examples:} Before starting this task, you may find it helpful to read this Pew Research Survey on facts and opinions in news articles. In particular, the 12 sentences they used to evaluate people’s ability to perform this task are as follows:

\paragraph{Factual statements:}
\begin{itemize}
    \item Health care costs per person in the U.S. are the highest in the developed world.
    \item President Barack Obama was born in the United States.
    \item Immigrants who are in the U.S. illegally have some rights under the Constitution.
    \item ISIS lost a significant portion of its territory in Iraq and Syria in 2017.
    \item Spending on Social Security, Medicare, and Medicaid make up the largest portion of the U.S. federal budget.
\end{itemize}

\paragraph{Opinion statements:}
\begin{itemize}
    \item Democracy is the greatest form of government.
    \item Increasing the federal minimum wage to \$15 an hour is essential for the health of the U.S. economy.
    \item Abortion should be legal in most cases.
    \item Immigrants who are in the U.S. illegally are a very big problem for the country today.
    \item Government is almost always wasteful and inefficient.
\end{itemize}

\paragraph{Borderline statements:}
\begin{itemize}
    \item Applying additional scrutiny to Muslim Americans would not reduce terrorism in the U.S.
    \item Voter fraud across the U.S. has undermined the results of our elections.
\end{itemize}

While GPT is able to classify all of these correctly, we would like to understand how well it fares when doing this classification for each sentence in a full news article. You will be asked to validate labels for a total of 50 articles (randomly sampled from August to October, 2024).

\textbf{Annotation Platform:} Annotation platform link.
\\
(You should enter your name in the same way in the submission ID field every time you log in to ensure your progress is saved as you go.)

\clearpage

\begin{figure}[h]
    \centering
    \includegraphics[width=\linewidth]{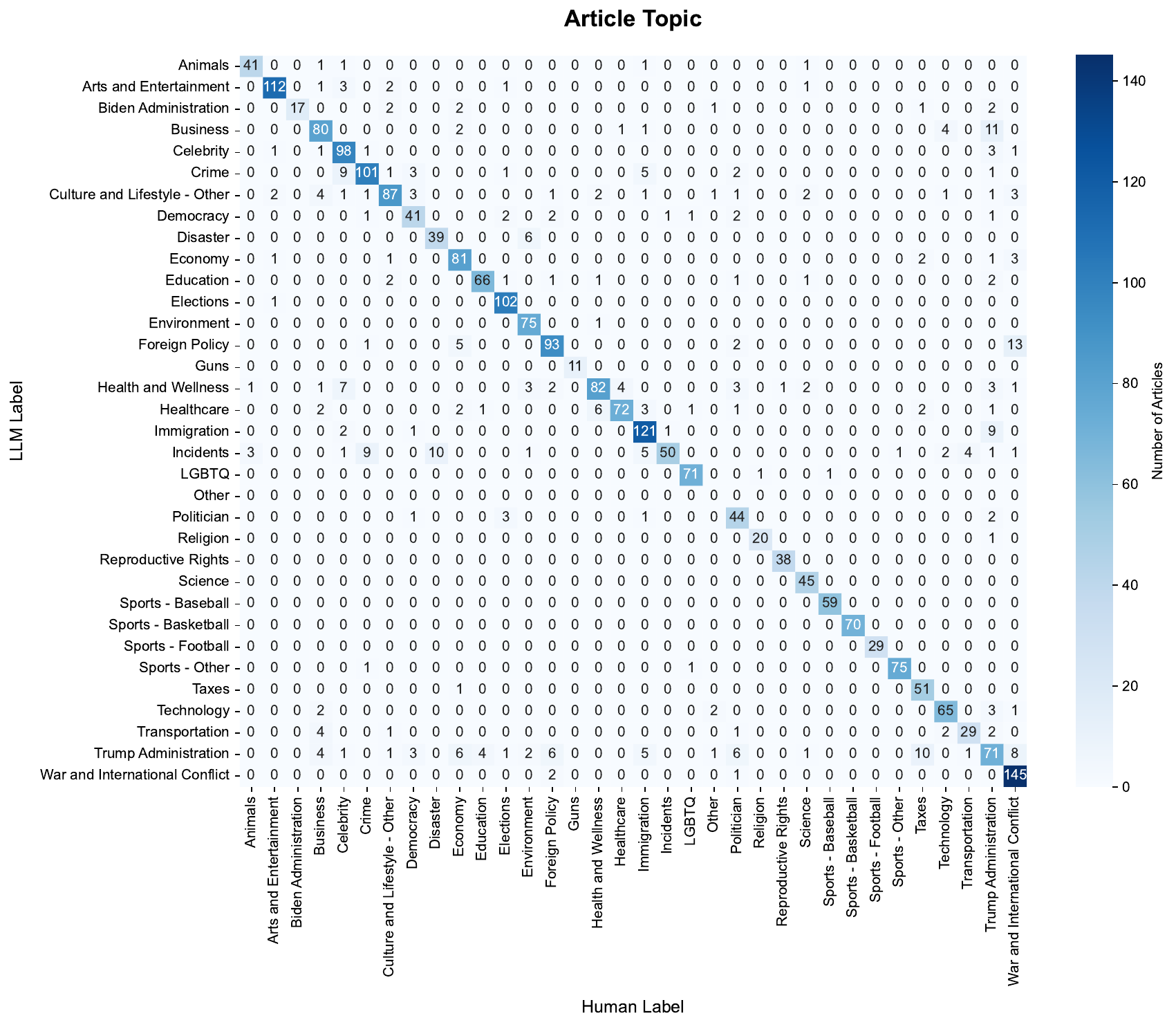}
    \caption{Confusion matrix for article topic classification between GPT-4o-mini and human annotators. The vast majority of articles are correctly classified, with confusions typically happening over articles which touch on multiple topics and can reasonably fall under either of them, e.g. the most common correction made by human annotators is from ``Foreign Policy" to ``War and International Conflict".}
    \label{fig:topic_confusion}
\end{figure}

\clearpage

\begin{figure}[h]
    \centering
    \includegraphics[width=\linewidth]{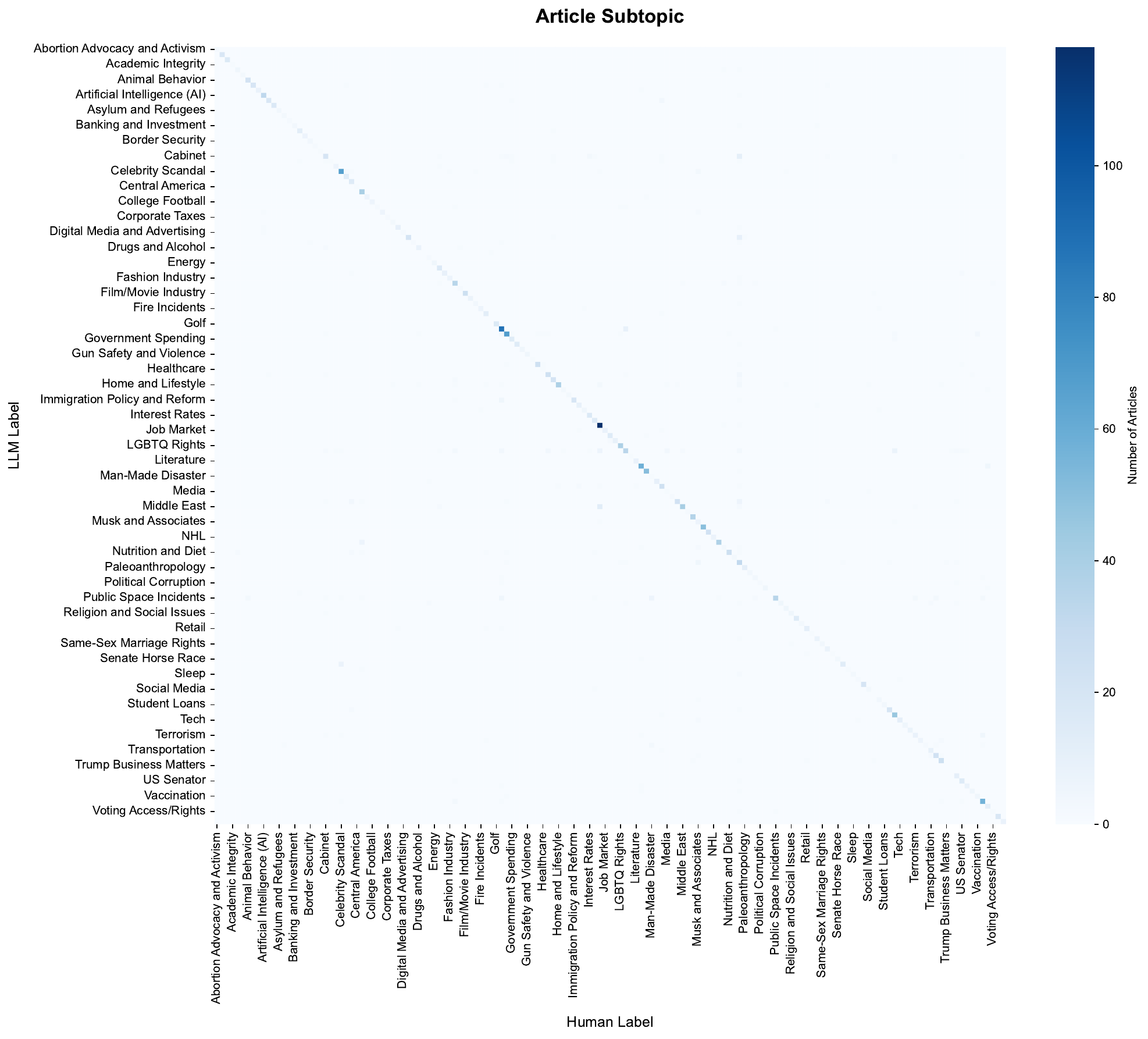}
    \caption{Confusion matrix for article subtopic classification between GPT-4o-mini and human annotators. Like in topic classification, the vast majority of articles are correctly classified, as can be seen by the density on the diagonal, with confusions typically happening over articles which touch on multiple topics and can reasonably fall under either of them.}
    \label{fig:subtopic_confusion}
\end{figure}

\clearpage

\begin{figure}[h]
    \centering
    \includegraphics[width=\linewidth]{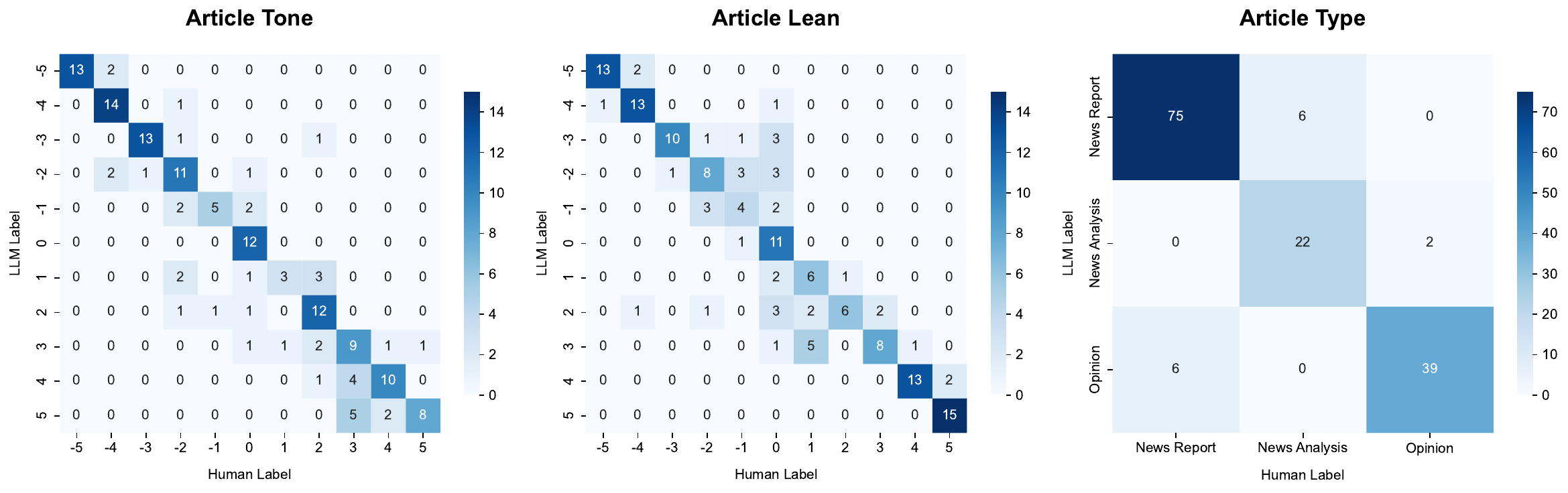}
    \caption{Confusion matrices for article lean, tone, and type labels between GPT-4o and human annotators. Note that each of these shows the combined confusion matrix for all three annotators, so they sum up to 150 articles.}
    \label{fig:lean_tone_confusion}
\end{figure}

\begin{figure}[h]
    \centering
    \includegraphics[width=\linewidth]{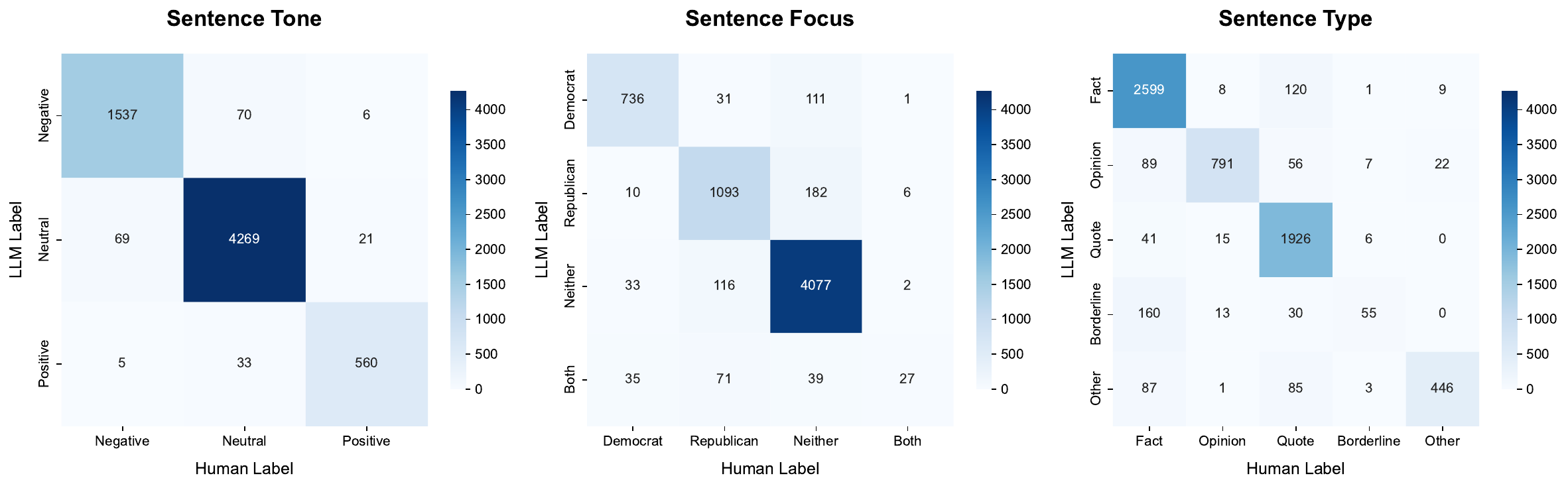}
    \caption{Confusion matrices for sentence type, tone, and focus labels between GPT-4o and human annotators. Note that each of these shows the combined confusion matrix for all three annotators, so they sum up to 6570 sentences.}
    \label{fig:sentence_confusion}
\end{figure}

\begin{figure}[ht]
    \centering
    \subfloat{
        \includegraphics[width=0.45\linewidth]{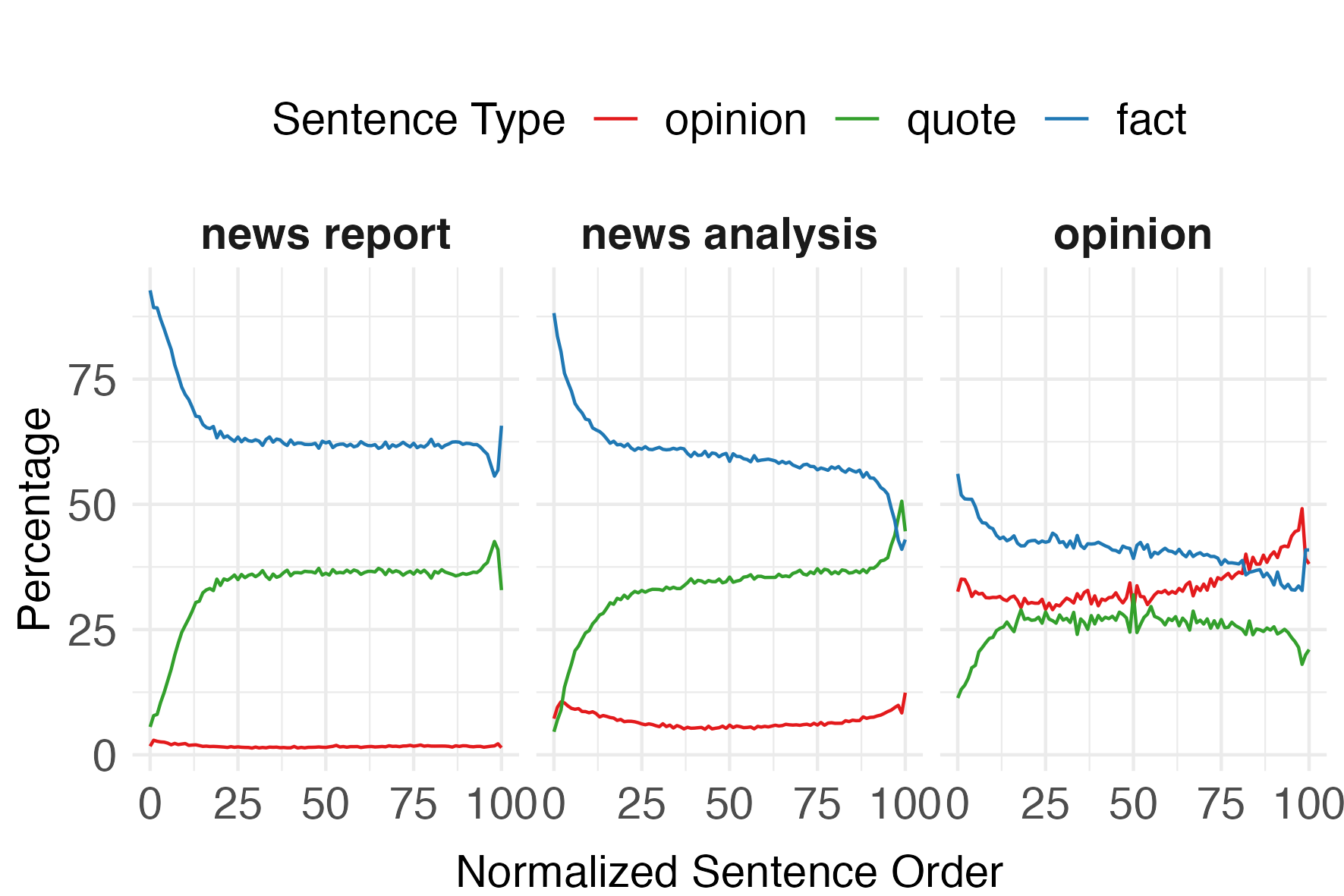}}
    \hfill
    \subfloat{
        \includegraphics[width=0.45\linewidth]{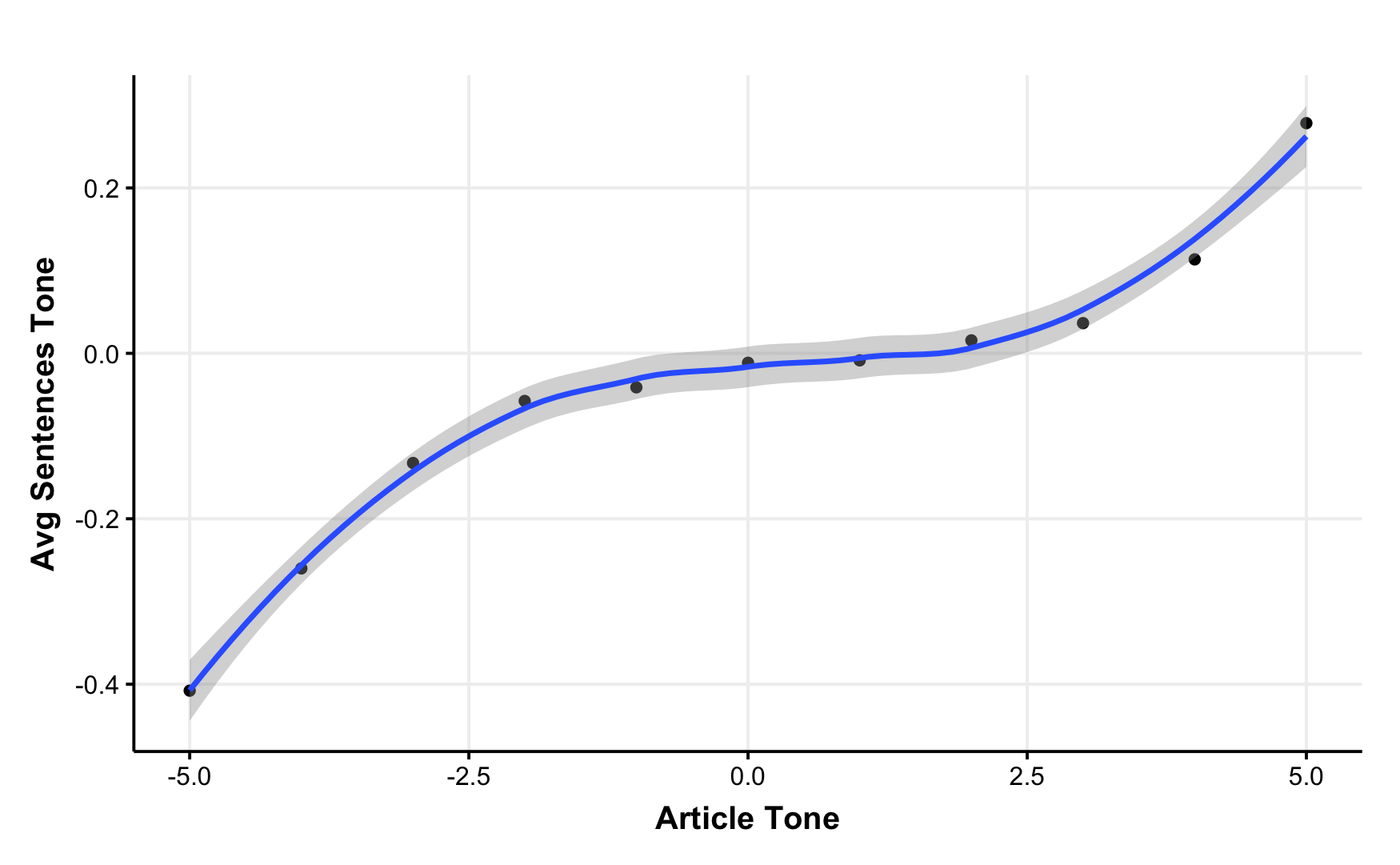}}
    \caption{A comparison of interdependently generated sentence- and article-level labels showing their agreement with each other. \textbf{Left:} We see that the distribution of fact and opinion sentences varies accordingly with the type of the article, with opinion pieces having more opinion sentences than news reports and vice versa for facts. \textbf{Right:} The average tone of sentences in an article exhibits a positive, monotonically increasing relation with the overall tone of the article itself. At the extremes, where most sentences are overwhelmingly positive or negative, the article also ends up having the same tone overall. In the middle, however, where the mix of sentences is more balanced, the tone of the article depends on its narrative and how those sentences support or contradict each other. }
    \label{fig:sent_art_comparison}
\end{figure}

\clearpage

\subsection*{Event clustering}

\begin{figure}[h!]
    \centering
    \includegraphics[width=0.7\linewidth]{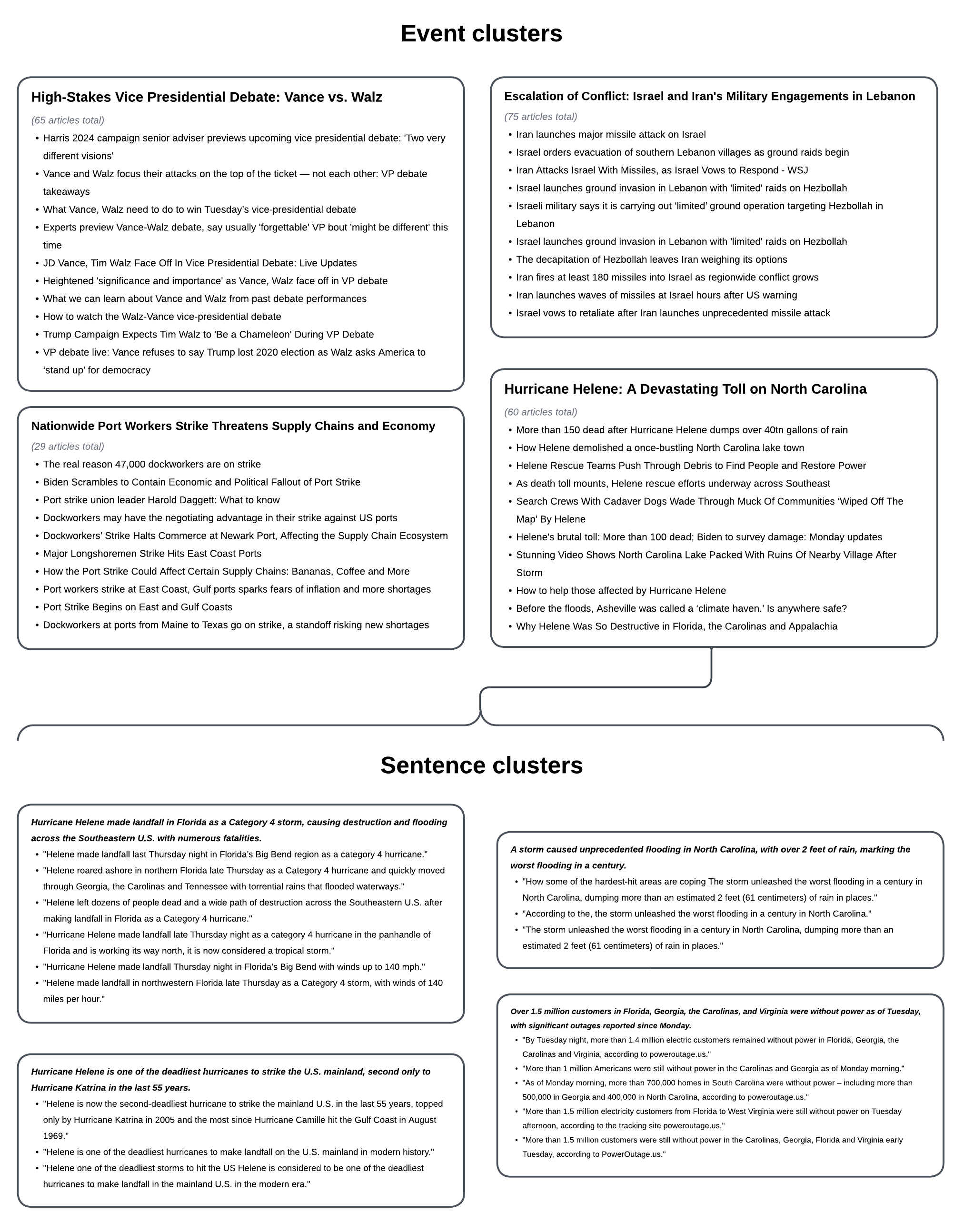}
    \caption{A sample of the data generated by our event clustering pipeline. \textbf{Top:} The major events that occured on October 1, 2024, showing the GPT-generated event titles and a sample of headlines from articles that have been clustered together. \textbf{Bottom: } The top facts about the event on Hurricane Helene on October 1, 2024. The sentence at the top of each group shows a synthetic statement generated by summarizing all the facts in the cluster while the following sentences show the original versions from those articles.}
    \label{fig:event_clustering}
\end{figure}

\clearpage

 \clearpage

\begin{table}[htbp]
\centering
\small
\begin{tabular}{p{0.8\textwidth}r}
\hline
\textbf{Event} & \textbf{Number of Articles} \\
\hline
Escalation of Conflict: Israel and Iran's Military Engagements in Lebanon & 75 \\
High-Stakes Vice Presidential Debate: Vance vs. Walz & 65 \\
Hurricane Helene: A Devastating Toll on North Carolina & 60 \\
Nationwide Port Workers Strike Threatens Supply Chains and Economy & 29 \\
The Legacy and Controversy of Pete Rose: Baseball's All-Time Hits Leader Passes Away & 20 \\
Political Fallout from Hurricane Helene: Claims and Responses & 17 \\
Celebrating Jimmy Carter's Historic 100th Birthday & 15 \\
Escalating Conflicts and Political Debates Amid Natural Disasters & 13 \\
Sean 'Diddy' Combs Faces 120 Sexual Abuse Allegations & 9 \\
Tributes to John Amos: Celebrated Actor of 'Good Times' and 'Roots' Dies at 84 & 8 \\
Tim Walz's Controversial Claims on China and Tiananmen Square & 8 \\
Claudia Sheinbaum: Mexico's First Female President & 6 \\
Georgia's Abortion Ban Ruled Unconstitutional & 6 \\
Frank Fritz, 'American Pickers' Star, Passes Away at 60 & 5 \\
Trump Withdraws from CBS '60 Minutes' Interview & 4 \\
NBA Legend Dikembe Mutombo Passes Away at 58 & 4 \\
Close Encounter Between Russian Fighter Jet and US F-16 Near Alaska & 3 \\
Julian Assange's Plea for Freedom Through Journalism & 3 \\
Fatal Shooting of Kentucky Judge by Sheriff & 3 \\
Morgan Wallen's Generous Donation for Hurricane Helene Relief & 3 \\
Major Verizon Service Outage Affects Customers Nationwide & 3 \\
Trump's GoFundMe Campaign for Hurricane Helene Victims Raises Over \$2 Million & 2 \\
\hline
\end{tabular}
\caption{The events that were covered in the news on October 1, 2024, along with the number of articles about them. }
\label{tab:event_stats}
\end{table}



\end{document}